%% file: main.tex
\documentclass[twoside]{article}

\usepackage[accepted] {aistats2026}

\usepackage{amsmath,amsfonts}
\usepackage{algorithmic}
\usepackage{algorithm}
\usepackage{array}
\usepackage{textcomp}
\usepackage{url}
\usepackage{verbatim}
\usepackage{graphicx}

\usepackage{cite}
\usepackage{caption}
\usepackage{subcaption}
\usepackage{hyperref}
\usepackage{url}
\usepackage{soul}
\usepackage{microtype}
\usepackage{amssymb}
\usepackage{tikz}
\usepackage{multirow}
\usepackage{multicol}
\usepackage{tabularx}
\usepackage{rotating,booktabs}
\usepackage{makecell}
\usepackage{graphicx}
\usepackage{algorithmic}
\usepackage{algorithm}
\usepackage{xcolor}
\usepackage{colortbl}
\usepackage{wrapfig}
\usepackage{amsmath}
\usepackage{amssymb}
\usepackage{mathtools}
\usepackage{amsthm}
\theoremstyle{plain}
\newtheorem{theorem}{Theorem}[section]
\newtheorem{proposition}[theorem]{Proposition}
\newtheorem{lemma}[theorem]{Lemma}

\theoremstyle{definition}

\theoremstyle{remark}

\usepackage{tikz}

\def\tagform@#1{\maketag@@@{\normalsize(\ignorespaces#1\unskip)}} % or \small, \large, etc.

\usepackage{enumitem}
\usepackage{siunitx} % 숫자를 정렬하고 단위를 표시하기 위한 강력한 패키지
\usepackage{booktabs}  % 더 전문적인 표를 위한 패키지

\input{math_commands.tex}

\usepackage[round]{natbib}

\begin{document}



\twocolumn[
\runningtitle{Multiple Invertible and Partial-Equivariant Function for Latent Vector Transformation to Enhance to}
\aistatstitle{Multiple Invertible and Partial-Equivariant Function for Latent Vector Transformation to Enhance Disentanglement in VAEs}
\runningauthor{Hee-Jun Jung, Jaehyoung Jeong, Kangil Kim}
\aistatsauthor{Hee-Jun Jung \And Jaehyoung Jeong \And  Kangil Kim$^\ast$}
\aistatsaddress{Gwangju Institute of \\ Science and Technology (GIST) \And  Gwangju Institute of \\ Science and Technology (GIST) \And Gwangju Institute of \\ Science and Technology (GIST)} 
]

\begin{abstract}
% Disentanglement learning is a core issue for understanding and reusing trained information in Variational AutoEncoder (VAE), and an effective inductive bias has been reported as a key factor.
 % However, the actual implementation of such bias is still vague.
% Although equivariance has been discussed as a form of inductive bias, leveraging it to improve disentanglement learning remains a challenging task.
Disentanglement learning is central to understanding and reusing learned representations in variational autoencoders (VAEs). 
Although equivariance has been explored in this context, effectively exploiting it for disentanglement remains challenging.
In this paper, we propose a novel method, called \textit{Multiple Invertible and Partial-Equivariant Transformation} (MIPE-Transformation), which integrates two main parts: 
(1) \textit{Invertible and Partial-Equivariant Transformation} (IPE-Transformation), guaranteeing an invertible latent-to–transformed-latent mapping while preserving partial input-to-latent equivariance in the transformed latent space; and (2) \textit{Exponential-Family Conversion} (EF-Conversion) to extend the standard Gaussian prior to an approximate exponential family via a learnable conversion.
In experiments on the 3D Cars, 3D Shapes, and dSprites datasets, MIPE-Transformation improves the disentanglement performance of state-of-the-art VAEs.
\end{abstract}

\section{Introduction}

Disentanglement learning aims to learn interpretable representations and has utility in tasks such as classification~\citep{classi_1}, zero-shot learning~\citep{zero-shot}, and domain adaptation~\citep{domain-1, domain-2}. 
A disentangled representation varies along one latent coordinate in a way that corresponds to a single factor of variation while keeping other factors essentially fixed. 
Among existing approaches, the Variational Autoencoder (VAE)~\citep{vae} is a primary framework; unsupervised methods encourage factorized latents and reduce inter-dimension correlation~\citep{betaVAE, beta-tcvae, factor-vae, cascadeVAE, hierarchical-vae}, but \citet{vae-inductive-bias} shows that disentanglement is not identifiable without inductive bias.

% To inject such bias, prior work follows two lines: (i) statistics-driven methods that replace the standard Gaussian prior with richer, often non-Gaussian alternatives. 
% For example, \citet{hierar_prior, isotrophic_gaussian} investigate Gaussian-mixture and Student-t priors and show the limitation of standard Gaussian prior for disentanglement learning.
% Also, \citet{Locatello2020WeaklySupervisedDW, concave} consider priors beyond the Gaussian family and improve disentanglement performance. 
% However, these approaches based on the primary defined disentangled representation~\citep{disen_definitiaon_1} rather than group-theory based definition.

To inject such bias, prior work follows two lines: 1) statistics-driven methods that replace the standard Gaussian prior with richer, often non-Gaussian alternatives. 
For example, \citet{hierar_prior, isotrophic_gaussian} investigate Gaussian-mixture and Student-\(t\) priors and demonstrate limitations of the standard Gaussian prior for disentanglement. 
Similarly, \citet{Locatello2020WeaklySupervisedDW, concave} consider priors beyond the Gaussian family and report improved disentanglement performance. 
However, these approaches are grounded in the original notion of disentangled representation~\citep{disen_definitiaon_1} rather than the group-theoretic formulation, and they do not explicitly encode equivariance.

The second line comprises group-theoretic methods that recast disentanglement in terms of the direct product of groups defined on the latent space~\citep{disen_definitiaon_2} by equivariant modeling between input and latent vector space rather than 2D planar symmetries. 
Representative works instantiate the general linear group ($GL(n)$)~\citep{commutative-vae, cfasl, group-invariant-equivariant} or cyclic groups~\citep{groupified-vae, homomorphicVAE} on latent vector space, and design equivariant VAE objectives that improve disentangled representation quality. 
However, these models typically retain a fixed standard Gaussian prior, which limits statistical flexibility.

% The second line 2) group-theoretic methods that recast disentanglement via the direct product of groups and design equivariant VAEs through defined symmetries on the latent vector space rather than 2D-planar symmetries.
% Defined a general linear group~\citep{commutative-vae, cfasl, group-invariant-equivariant}, and cyclic group~\citep{groupified-vae, homomorphicVAE} as a symmetry on the latent vector space improves disentangled representation quality.
% However, these works are typically retained with a fixed Gaussian prior, limiting statistical flexibility.

% ~\citep{disen_definitiaon_2, commutative-vae, groupified-vae, cfasl, homomorphicVAE, group-invariant-equivariant}, which are aligned factor-wise symmetries to a single latent vector dimension. 
% while the latter typically retains a fixed Gaussian prior, limiting statistical flexibility.
% However, the former does not explicitly encode equivariance—now central to robust factorization—while the latter typically retains a fixed Gaussian prior, limiting statistical flexibility.

Motivated by these observations, we target two desiderata: 1) preserving partial equivariance between the input and a transformed latent space, and 2) enabling diverse priors. 
% We introduce a learnable invertible latent-to-latent transformation that addresses both: it preserves partial equivariance and, together with an exponential-family conversion (EF-conversion), maps the transformed latent variables to flexible priors. 
We introduce a learnable, invertible latent-to-latent transformation that preserves partial equivariance between inputs and latent space, and via an \textit{exponential-family conversion} (EF-conversion), maps the transformed latent vectors to flexible (non-Gaussian) priors.
We refer to the module as the Multiple Invertible and Partial-Equivariant (MIPE) transformation, which can be inserted into existing VAEs.

Through quantitative and qualitative experiments on 3D Cars, 3D Shapes, and dSprites, MIPE yields consistent improvements in disentangled representation learning. 
Our contributions are:
\setlist{nolistsep}
\begin{enumerate}[noitemsep]
    \item A symmetric matrix exponential latent-to-latent transformation with invertibility and analyzed partial equivariance properties.
    \item A training procedure and losses that allow VAEs to learn an unknown latent distribution approximated by an exponential family.
    \item The MIPE architecture that integrates multiple IPE transformations with EF-conversion and can be plugged into state-of-the-art VAEs.
    \item Empirical analysis demonstrating the effectiveness of MIPE on standard disentanglement benchmarks.
\end{enumerate}

\section{Preliminaries}
\label{appendix: preliminaries}

% \subsection{Group Theory}
% \label{subsec: group theory}

% \hspace{4mm} \textbf{Binary operation:} Binary operation on a set $S$ is a function that $\ast: S \times S \rightarrow S$, where $\times$ is a cartesian product.

\textbf{Group:} A group is a set $G$ together with binary operation $\ast$, that combines any two elements $g_a$ and $g_b$ in $G$, such that the following properties:
\setlist{nolistsep}
\begin{itemize}[noitemsep]
    \item closure: $g_a, g_b \in G \Rightarrow g_a \ast g_b \in G$.
    \item Associativity: $\forall g_a, g_b, g_c \in G$, \textit{s.t.} $(g_a \ast g_b) \ast g_c = g_a \ast (g_b \ast g_c)$.
    \item Identity element: There exists an element $e \in G$, \textit{s.t.} $\forall g \in G, e \ast g = g \ast e = g$.
    \item Inverse element: $\forall g \in G, \exists g^{-1} \in G$: $g \ast g^{-1} = g^{-1} \ast g = e$.
\end{itemize}

\textbf{Group action:}
Let set $X$, and $(G, \circ)$ be a group,  binary operation $\cdot: G \times X \rightarrow X $, then group action $\alpha: \alpha(g, x) = g \cdot x$ following properties:
\begin{itemize}[noitemsep]
    \item Identity: $e \cdot x =x$, where $e \in G, \,\,x\in X$.
    \item Compatibility: $\forall g_1, g_2 \in G, \,\, x\in X, \,\, \alpha((g_1 \circ g_2), x) = \alpha(g_1, \alpha(g_2, x))$.
\end{itemize}

\textbf{Equivariant map:}
Given $X$ and $Y$ are $G$-set, and group action $\rho: G \times Y \rightarrow Y$. Then a function $f: X \rightarrow Y$ is equivariant if 
\begin{equation}
\label{equation: equivariant map}
f(\alpha(g, x)) = \rho(g, f(x)).
\end{equation}

% \textbf{Group action:} Let $(G, \ast)$ be a group and set $X$, binary operation $\cdot: G \times X \rightarrow X$, such that following properties:
% \begin{itemize}
%    \item Identity: $e \cdot x = x$, where $e \in G, x \in X$.
%    \item Compatibility: $\forall g_a, g_b \in G, x \in X, (g_a \ast g_b) \cdot x = g_a \cdot (g_b \cdot x)$.
%\end{itemize}

% \textbf{Equivariant map:} Let $G$ be a group and $X_1, X_2$ be two sets with corresponding group action of $G$ in each sets: $T^{X_1}_g, T^{X_2}_g$, where $g \in G$. Then a function $f: X_1 \rightarrow X_2$ is equivariant if $f(T^{X_1}_g \cdot X_1) = T^{X_2}_g \cdot f(X_1)$.

\textbf{Partial-Equivariant map:}
Let subset of $G$ be $G^\prime \subset G$, then a function $f: X \rightarrow Y$ is a partial-equivariant if 
\begin{equation}
\label{equation: partial equivariant map}
f(\alpha(g^\prime, x)) = \rho(g^\prime, f(x))\,\, \text{where}\,\, g^\prime \in G^\prime.
\end{equation}

% \textbf{Partial Equivariance}~\citep{partial_equiv}: Let subset of $G$ be $\Upsilon \subset G$, then $f$ is a partially equivariant map to $G$:
% \begin{equation}
%    f(T^{X_1}_\upsilon \cdot X_1) = T^{X_2}_\upsilon \cdot f(X_1), ~\text{where}~ \forall \upsilon \in \Upsilon.
% \end{equation}

\textbf{Homogeneous space:}
Let a group \(G\) act (on the left) on a smooth manifold \(X\) via
\(\cdot: G\times X \to X\).
The \(G\)-space \(X\) is called \emph{homogeneous} if the action is \emph{transitive}, i.e.,
for all \(x,y\in X\) there exists \(g\in G\) with \(g\cdot x = y\).

% \textbf{Homomorphsim:} Let $(G, \cdot), (H, \circ)$ be two groups. If mapping function $h: G \rightarrow H$, \textit{s.t.} $h(g_i \cdot g_j) = f(g_i) \circ f(g_j)$, then $f$ is called homomorphism.

\paragraph{Exponential Family} We introduce the conjugate prior and posterior distribution of the exponential family.
\label{appendix: background}

\textbf{Power Density Function:} 
% Power density function of the exponential family (PDF) generalized formulation:
\begin{equation}
    \begin{split}
        P{ \vx}( \vx|\boldsymbol{\theta}) 
        % & = h( \vx) \text{exp} (\boldsymbol{\theta}^\intercal T( \vx) - A(\boldsymbol{\theta})) \\ 
        & = \text{exp} (\boldsymbol{\theta}^\intercal T( \vx) - A(\boldsymbol{\theta}) + B( \vx)),
        \label{eq:pdf}
    \end{split}
\end{equation}
where \textit{sufficient statistics} $T(\cdot)$, \textit{log-normalizer} $A(\cdot)$, and \textit{carrier or base measure} $B(\cdot)$ are known functions, samples $ \vx$ from distribution, and \textit{natural parameter} $\boldsymbol{\theta}$.

If observation is a set of independent identically distributed, then Eq.~\ref{eq:pdf} is modified as
\begin{equation}
    p(\mathbf{X}|\boldsymbol{\theta}) = \prod_{n=1}^N h(\vx_n) \text{exp} (\boldsymbol{\theta}^{\intercal} \sum_{n=1}^N T(\vx_n) -A(\boldsymbol{\theta})), 
\label{eq: likelihood}
\end{equation}
where observation $\mathbf{X}=\{\vx_1, \cdots \vx_N \}$~\citep{prml}.

\textbf{Conjugate Prior:}
% As introduced in~\citet{prml, exp_family_bayesian},  
Exponential family admits a conjugate prior:
\begin{equation}
    q(\boldsymbol{\theta} | \mathcal{\xi}, \boldsymbol{\nu}) = \text{exp} (\boldsymbol{\nu} \boldsymbol{\theta}^\intercal \mathcal{\xi}- \boldsymbol{\nu} A (\boldsymbol{\theta}) + B^\prime(\mathcal{\xi}, \boldsymbol{\nu})),
    \label{eq:conju_prior}
\end{equation}
where $B^\prime(\cdot)$ is a \textit{normalize coefficient} and $\nu$ is evidence, and it is expressed by prior natural parameter $\mathcal{\xi}$.

\textbf{Posterior:} 
Let multiply the prior Eq.~\ref{eq:conju_prior} by the Eq.~\ref{eq: likelihood} to obtain the posterior distribution in the form
% The procedure represents a posterior distribution in the exponential family by adopting the following form:
\begin{equation}
\small
    p(\boldsymbol{\theta} | \mathbf{X}, \mathcal{\xi}, \boldsymbol{\nu}) \propto \text{exp} (\boldsymbol{\theta}^\intercal (\sum_{n=1}^N T(\vx_n) + \boldsymbol{\nu} \mathcal{\xi}) - A(\boldsymbol{\theta})),
    \label{eq:posteior_}
\end{equation}
where the \textit{evidence} $\boldsymbol{\nu}$, \textit{sufficient statistics} $T(\cdot)$ and \textit{log-normalizer}, $A(\cdot)$ are known functions, samples $\mathbf{X}$
% $\mathbf{X} = \{\mathbf{x}_1, \mathbf{x}_2, \ldots, \mathbf{x}_n \}$ 
from distribution, and \textit{natural parameter} of posterior $\boldsymbol{\theta}$ and of prior $\mathcal{\xi}$~\citep{prml}.

\begin{figure*}[t]
    \centering
    \begin{subfigure}{0.99\textwidth}
        \centering
        \includegraphics[width=\textwidth]{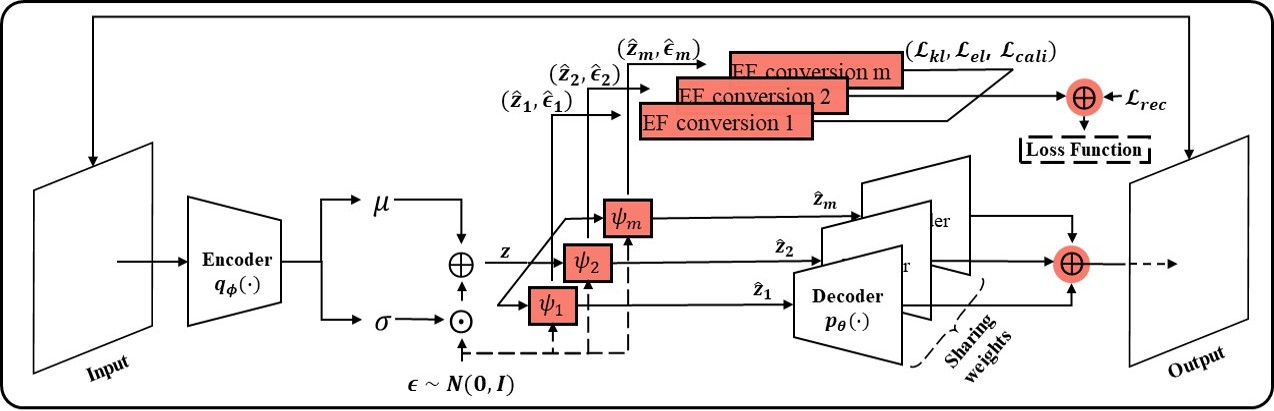}
        % \caption{Overview}
        % \label{figure: 3dshapes r2r images}
    \end{subfigure}
    % \vfill
    %     \begin{subfigure}{0.45\textwidth}
    %     \centering
    %     \includegraphics[width=\textwidth]{overview_b.jpg}
    %     \caption{b}
    %     \label{figure: 3dshapes r2r dci}
    % \end{subfigure}
    % \hfill
    %     \begin{subfigure}{0.45\textwidth}
    %     \centering
    %     \includegraphics[width=\textwidth]{overview_c.jpg}
    %     \caption{c}
    %     \label{figure: 3dshapes r2r dci}
    % \end{subfigure}
    \caption{The overall architecture of our proposed \textit{MIPET}-VAE.
    The invertible and partial-equivariant function $\psi(\cdot)$ for latent-to-latent (L2L) transformation consists of a symmetric matrix exponential to be 1) invertible and 2) partial-equivariant.
    Then 3) EF conversion module converges the distribution of unrestricted $\hat{ \vz}$ to be EF with $\mathcal{L}_{el}$ loss.
    Also, it applies KL divergence loss ($\mathcal{L}_{kl}$) between the transformed posterior and prior, which are expressed by the power density function of EF.
    In the last, EF conversion reduces the computational error ($\mathcal{L}_{cali}$) between approximated and true KL divergence.
    % 4) The reddish color represents the integration parts. The blue figures represent each property.
     % The details of the gray box are in Fig.~\ref{sub fig:suboverview}.
     }
    \label{fig:overview}
    \vspace{-10pt}
\end{figure*}

\section{Related Work}
\subsection{Statistical Approaches for Disentangled Representations}
Recently, various studies have focused on unsupervised disentanglement learning. 
% Previous works are based on the definition by~\cite{disen_definitiaon_1}.
% One of the branches is InfoGAN~\cite{info-gan} based works such as IB-GAN~\cite{ib-gan}, which implement an additional regularizer to improve informativeness~\cite{dci}.
% The other branch is based on the VAE.
$\beta$-VAE~\citep{betaVAE} penalizes the Kullback-Leibler divergence (KL divergence) using weighted hyper-parameters.
Factor VAE~\citep{factor-vae} and $\beta$-TCVAE~\citep{beta-tcvae} are trained using total correlation (TC) to encourage independent dimensions in a latent vector, employing a discriminator and decomposed components of the KL divergence term.
However, these works are based on the standard Gaussian prior.

% \subsection{Inductive Bias with Prior Distribution}
Other VAE approaches implement other prior from Gaussian distribution to transformed Gaussian distribution, Gaussian mixture distribution~\citep{gaussian-mixture} or von Mises-Fisher distribution~\citep{hyper-vae}.
~\citet{isotrophic_gaussian} shows that model regularization with a Gaussian prior is inappropriate to convey sufficient dataset information and for disentanglement learning.
Rather than using a Gaussian distribution, we employed an exponential family with high flexibility that encompasses a range of different distributions.
InteL-VAE~\citep{intel-vae} utilize the invertible function to transform a Gaussian distribution to a complex distribution as~\citet{isotrophic_gaussian}. 
Differently, we show a clearer relation of invertibility to disentanglement and improve VAEs to use its unrestricted form of prior.
% Invertible and equivariant Deep Neural Networks have been investigated with normalizing flows.
% As proven by~\cite{invertible-matrix-exponential}, utilized matrix exponential on Neural networks is invertible, but it only provides mathematical foundations of the transformation.
% Matrix exponential is utilized to implement an invertible and equivariant function to improve the generative flow compare to linear function~\cite{NEURIPS2020_d3f06eef}.
% To specify the exponential family, other works contribute uncertainty of exponential family distribution with Bayesian update~\cite{exponential_uncertainty_00, exp_family_bayesian}. In addition,~\cite{exp} hierarchically controls the natural parameter across the layers and determines the exponential family distribution with the moment of sufficient statistic.
% In our work, we show how to use it for disentanglement learning.
Differently, we consider the recent disentanglement definition based on group theory~\citep{disen_definitiaon_2}.

\subsection{Inductive Bias with Group Theory-Based Approaches}
% \textbf{Group Theory-Based Approaches for Disentangled Representations:}
Following the definitions of disentangled representation learning based on group theory, several works have emphasized equivariant and improved disentangled representation learning.
Commutative Lie Group VAE (CLG-VAE)~\citep{commutative-vae} proposed a direct mapping of the latent vector into Lie algebra to obtain a group structure (inductive bias) with constraints: commutative and hessian loss.
Furthermore, Groupified VAE~\citep{groupified-vae} utilizes the Spatial Broadcast Decoder~\citep{broadcast-decoder} to implement an equivariant function to the cyclic group while guaranteeing the commutativity and invertibility of group actions.
Topographic VAE~\citep{t-vae} combines Student's-t distributions with variational inference and enforces rotated latent vectors to be equivariant.
On the other hand, we apply an unrestricted prior and posterior for disentanglement learning.

% There are several inductive biases to learning unsupervised disentanglement, such as group theory based and sequential order.
% In this section, we briefly discuss sequential order inductive bias even though its method is considered in different domains such as text and video frames.
% To individualize the static (time-invariant) and dynamic (time-variant), \cite{sequential-autoencoder, contrastively-vae} proposed the latent variables one ($f$) is only dependent on the given times series datasets $x_{1:T}$, and the other ($\mathbf{z}_{1:T})$ is dependent on the $x_{1:T}$ and $f$.
% Moreover,~\cite{contrastively-vae} propose the novel ELBO with maximizing mutual information between the input and the latent vectors.
% These works empirically show that sequential order which includes separated latent vectors improves unsupervised disentanglement learning with diverse qualitative analysis.
% Differently in group theory based approaches, the proposed methods consider equivariant function between input and latent vector space.

\vspace{-10pt}
\section{Method}
\label{sec:method}
The overview of a VAE equipped with MIPE-transformation is shown in Fig.~\ref{fig:overview}.
We aim to explain the three core components of the MIPE-transformation.
In Section~\ref{subsec:unit mapping function}, we mathematically prove why the \textit{IPE-transformation}, which transforms latent vectors, better preserves partial-equivariance when it is an invertible and symmetric matrix.
In Section~\ref{subsec:ef based vae}, we describe \textit{EF-conversion}, which extends the Gaussian distribution to a diverse exponential family distribution.
Finally, in Section~\ref{subsec:multiple unit}, we illustrate how multiple IPE-transformations and EF-conversions can be integrated into a VAE-based model for disentangled representation. 
% to implicitly inject inductive bias.

\textbf{Notation}
We denote $g_I \in G_I, g_L \in G_L, \, \text{and} \, g_T \in G_T$.
$M_n(\mathbb{R})$ is $n \times n$ real matrix, $E_M$ is $\{ \mathbf{e}^M | M \in M_n(\mathbb{R}) \}$, 
$E_S$ is a set of $\{ \mathbf{e}^S | S \in Sym_n(\mathbb{R}) \}$, and $G_S$ is a group $G_S:(\mathbf{e}^S, \ast)$, where $\ast$ is a matrix multiplication (equal to commutative).
Further details are in Appendix~\ref{appendix: notation and groups}.
% $\mathcal{X}, \mathcal{Z}, \, \text{and} \, \hat{\mathcal{Z}}$ are homogeneous space of group $G_I, G_L, \, \text{and} \, G_T$. 

\textbf{Assumption}
% \label{paragraph: assumption}
Let's assume that 1) encoder $q_\phi(\cdot)$ is a partial-equivariant function, 2) $\mathcal{X}, \mathcal{Z}^\prime, \, \text{and} \, \hat{\mathcal{Z}}^\prime$ are homogeneous space of group $G_I, G_L, \, \text{and} \, G_T$, 3) $G_L \subset GL_n(\mathbb{R})$, where $GL_n(\mathbb{R})$ is the general linear group, and 4) matrices are restricted to a compact subset $K \subset M_n(\mathbb{R})$ 
% (e.g., a bounded SPD set such as $\{A\in \mathrm{SPD}_n:\ \alpha I \preceq A \preceq \beta I\}$ with $0<\alpha<\beta<\infty$), 
and probabilities are defined with respect to the normalized uniform probability measure $\mu$ on $K$ (a modeling assumption motivated by finite-precision implementations).

\begin{figure*}
    \centering
    \begin{subfigure}{0.73\textwidth}
    \includegraphics[width=\textwidth]{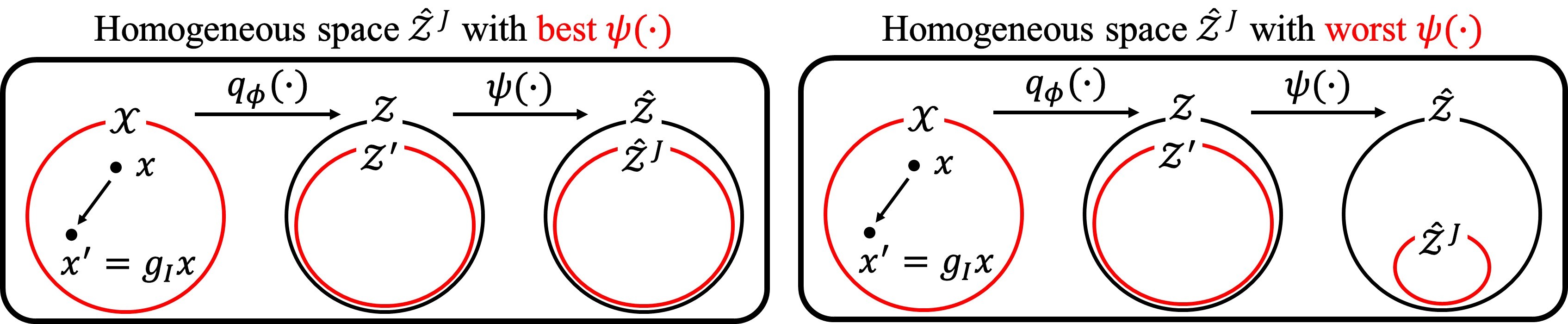}
    \caption{Visualization of homogeneous space (red) as best and worst L2L transformation $\psi(\cdot)$}
    \label{subfigure: overview b}
    \end{subfigure}
    \hfill
    \begin{subfigure}{0.25\textwidth}
    \includegraphics[width=\textwidth]{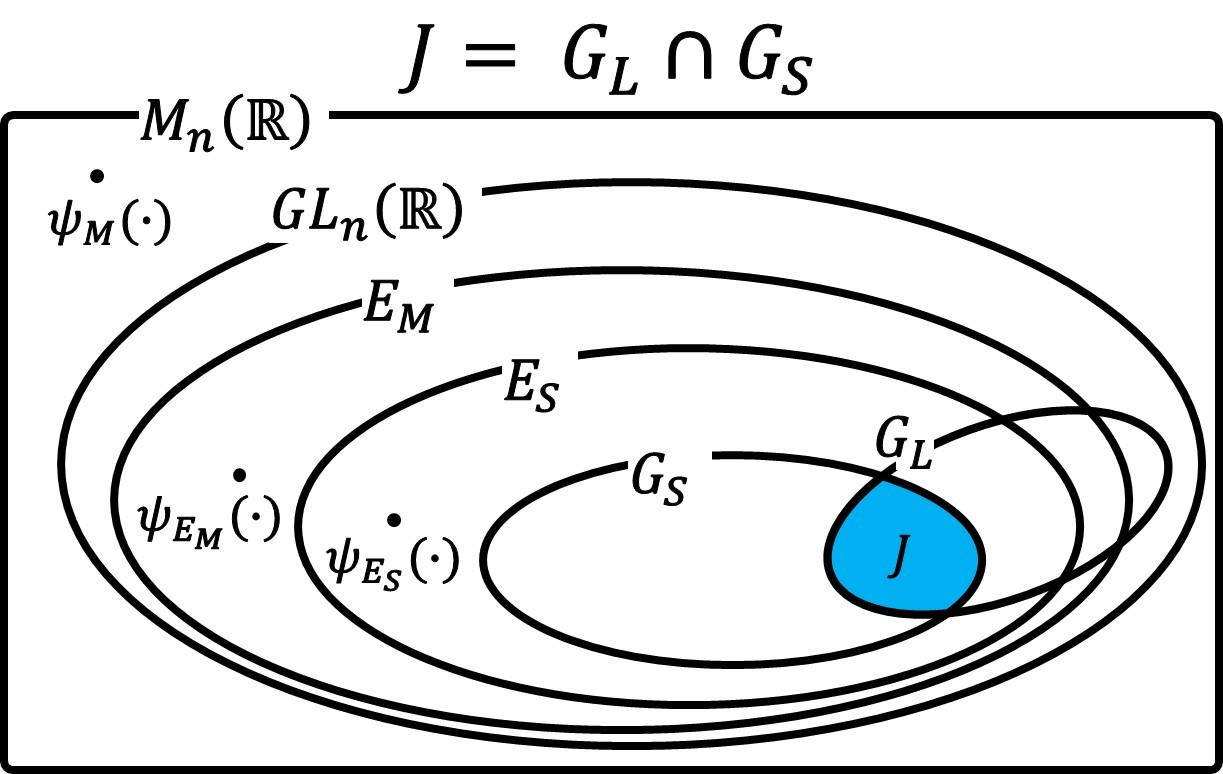}
    \caption{Venn diagram of groups}
    \label{subfigure: overview c}
    \end{subfigure}
    \caption{
    % The homogeneous space $\mathcal{Z}^J$ is determined by endoer $q_\phi(\cdot)$, and cardinality of $\hat{\mathcal{Z}}^J$ is depends on the L2L transformation $\psi^\star \;=\; \arg\max_{\psi \in M_n(\mathbb{R})}\; \big|\widehat{\mathcal{Z}}^{\,J}(\psi)\big|$.
    The homogeneous space \(\mathcal{Z}^\prime\) is induced by the encoder \(q_\phi\), and the cardinality of the \(\widehat{\mathcal{Z}}^{\,J}\) depends on the latent-to-latent (L2L) transformation. 
    % Accordingly, we choose$\psi^\star \in \arg\max_{\psi \in M_n(\mathbb{R})}\; \big|\widehat{\mathcal{Z}}^{\,J}(\psi)\big|$.
    %$\psi(\cdot) = \text{argmin}_{\psi(\cdot) \in M_n(\mathbb{R})} |\mathcal{Z}^J| - |\hat{\mathcal{Z}}^J|$.
    % The left side represents the homogeneous space of group $G_I, G_L, \, \text{and} \, G_T$ as $q_\phi(\cdot), \, \text{and} \, \psi(\cdot)$ are equivariant functions.
    % However, most cases are both $q_\phi(\cdot), \, \text{and} \, \psi(\cdot)$ are partial-equivariance, and the cardinality of the homogeneous space of the group $G_T^J$ depends on $\psi(\cdot)$.
    % We denote $g_I \in G_I, g_L \in G_L, \, \text{and} \, g_T \in G_T$, $\mathcal{X}, \mathcal{Z}, \, \text{and} \, \hat{\mathcal{Z}}$ are homogeneous space of group $G_I, G_L, \, \text{and} \, G_T$. 
    % $G_I$, and $G_L$ are obtained through encoder $q_\phi$ (encoder equivariance condition).
    % %Fig.~\ref{sub fig: partially equivaraince} 
    % The left side figure shows the relation between each space and symmetries.
    % %If $q_\phi$, and 
    % If $\psi(\cdot)$ is equivariant function over all $G_L$, and $G_T$, then %there exist $\Xi$, and 
    % there exist $\Gamma$, where $\Gamma: G_L \rightarrow G_T$, and $\Xi \circ \Gamma: G_I  \rightarrow G_T$.
    % %The same group is colored blue.
    % However, unrestricted $\psi(\cdot)$ has no guarantee to be partial- or full-equivariant.
    % The red arrows represent our method: L2L transformation guarantees $\Gamma^J: G_L^J \rightarrow G_T^J$, and $\Xi^J \circ \Gamma^J: G_I^J \rightarrow G_T^J$, given the encoder equivariance condition $\Xi: G_I \rightarrow G_L$.
    }
    \vspace{-10pt}
    \label{sub fig:suboverview}
\end{figure*}

\subsection{Partial-Equivariant Function and Invertible L2L Transformation}
\label{subsec:unit mapping function}

% \textbf{Proof Strategy and Plan}
% As shown in Fig.~\ref{subfigure: overview b}, we aim to find the best case of L2L transformation $\psi(\cdot)$ that is a symmetric exponential matrix to preserve partial-equivariance between the input and transformed latent vector space.
% \begin{figure*}[t]
% %\vskip 0.2in
%     \centering
%     \centerline{\includegraphics[width=1.0\textwidth]{figure_overview_ver28.jpg}}
%     \caption{The overall architecture of our proposed \textit{MIPET}-VAE.
%     The invertible and partial-equivariant function $\psi(\cdot)$ for L2L transformation consists of a symmetric matrix exponential to be 1) invertible and 2) partial-equivariant.
%     %$\psi(\cdot)$ transform latent vector $ \vz \sim \mathcal{N}(\mu, \Sigma)$ and transformed vector $\hat{ \vz} \sim P_T$.
%     Then 3) EF conversion module converges the distribution of unrestricted $\hat{ \vz}$ to be EF with $\mathcal{L}_{el}$ loss.
%     Also, it applies KL divergence loss ($\mathcal{L}_{kl}$) between the transformed posterior and prior, which are expressed by the power density function of EF.
%     In the last, EF conversion reduces the computational error ($\mathcal{L}_{cali}$) between approximated and true KL divergence.
%     4) The reddish color represents the integration parts. The blue figures represent each property.
%      The details of the gray box are in Fig.~\ref{sub fig:suboverview}.}
%     \label{fig:overview}
% \end{figure*}

\paragraph{Why Should L2L Transformation be Equivariant?}
% Let's assume that encoder $q_\phi(\cdot)$ is a partial-equivariant function.
% Let's consider equivariant function between the input and transformed latent vector space, directly used for a decoder in the VAE frameworks.
We transform the latent vector $\vz$ sampled from a Gaussian distribution to the transformed latent vector $\hat{\vz}$ to be sampled from a different distribution.
However, not all L2L transformations preserve the partial-equivariance of encoder condition.
% As shown in Fig.~\ref{subfigure: overview b}, finding $\psi(\cdot) = \text{argmin}_{\psi(\cdot) \in M_n(\mathbb{R})} |\mathcal{Z}^J| - |\hat{\mathcal{Z}}^J|$ get the largest cardinality of homogeneous space of group $G_T^J$.
% The homogeneous $\hat{\mathcal{Z}}^J$ implies that a single element of the latent vector $\hat{z}$ represents all elements of $\hat{\mathcal{Z}}^J$.
% Therefore, if $\hat{\vz}_0$ is a disentangled representation, then any elements of $\hat{\mathcal{Z}}^J$ are also disentangled because $\hat{\vz} = \alpha(g, \hat{\vz}_0)$, where $g \in G_1 \times G_2 \times \cdots \times G_k$, and $\hat{\vz}_0 \in \hat{\mathcal{Z}_1} \times \hat{\mathcal{Z}_2} \times \cdots \times \hat{\mathcal{Z}_k}$~\citep{disen_definitiaon_2}.
As illustrated in Fig.~\ref{subfigure: overview b}, we choose \(\psi\) to \emph{maximize} the cardinality of the homogeneous \(G_T^J\)-space induced in latent space:
\begin{equation}
\small
    \psi^\star \;=\; \arg\max_{\psi \in M_n(\mathbb{R})}\; \big|\widehat{\mathcal{Z}}^{\,J}(\psi)\big|,
\quad
\label{equation: argmax}
\end{equation}
where $G_T^J$ act on the latent space via $\alpha$ and define
$\widehat{\mathcal{Z}}^{\,J}(\psi):=\{\alpha(g,\hat{\vz})\mid g\in G_T^J\}$.
As a homogeneous $G_T^J$–space, fixing a base point $\hat{\vz}_0$ determines all elements $\alpha(g,\hat{\vz}_0)$.
If the latent space factors as $\hat{\mathcal{Z}}=\hat{\mathcal{Z}}_1\times\cdots\times\hat{\mathcal{Z}}_k$ and the action is componentwise, then every $\alpha(g,\hat{\vz}_0)$ with $g\in G_1\times\cdots\times G_k$ is also disentangled~\citep{disen_definitiaon_2}.

% Here \(\widehat{\mathcal{Z}}^{\,J}(\psi)\) denotes the \(G_T^J\)-orbit of a \(\hat{\vz}\) under the latent action, $\widehat{\mathcal{Z}}^{\,J}(\psi) \;=\; \{\, \alpha(g_T^J,\hat{\vz}) \;\mid\; g_T^J \in G_T^J \,\}$ (equal to homogeneous space).
% A homogeneous space implies that any single element of the orbit determines the entire space.
% Consequently, if \(\hat{\vz}_0\) is a disentangled representation, then every element
% \(\hat{\vz} = \alpha(g_T^J,\hat{\vz}_0)\) with \(g \in G_1 \times \cdots \times G_k\) is also disentangled, since the group action factors over the product decomposition
% \(\hat{\mathcal{Z}}_1 \times \cdots \times \hat{\mathcal{Z}}_k\) of the latent space~\citep{disen_definitiaon_2}.

% This problem is more precisely shown in Fig.~\ref{sub fig:suboverview}, which illustrates partial equivariance condition over the input space $\mathcal{X}$, latent vector space $\mathcal{Z}$, and its transformed latent vector space $\hat{\mathcal{Z}}$ with a corresponding group of symmetries $G_I$, $G_L$, and $G_T$, respectively.
% In the VAEs literature, it has not been reported to restrict L2L transformation to guarantee equivariant function between two spaces, so we propose a solution to guarantee at least a part of symmetries to be equivariant.

\paragraph{Equivariance Property with Symmetric Matrix Exponential}
% To enhance the equivariance of L2L transformation, we focus on which transformation is appropriate.
We prove that an invertible and symmetric matrix preserves partial-equivariance better than other matrices for Eq.~\ref{equation: argmax}.
We show that 1) a group with symmetric and invertible matrices guarantees equivariance of $\psi(\cdot)$ over the specific group $G_S$, 2) this $\psi(\cdot)$ being equivariant over subset of symmetries between the input space and transformed latent vector space (equal to be a partial-equivariant), and
% to show the connection between two spaces, and 
3) the invertible and symmetric matrix (symmetric matrix exponential) increases the probability of $\psi (\cdot)$ to be in the group (equal to satisfy Eq.~\ref{equation: argmax}).
% (equal to be equivariant over the subset of symmetries).
% We particularly call the transformations as~\textit{symmetries}~\citep{symmetries} to distinguish them from IPE- and I2L-transformations.

For the generality of our method, we consider an arbitrary VAE model that has no restriction on creating intersections to any set as Fig.~\ref{subfigure: overview c}.

% \begin{figure*}[t]
%    \centering
%    \includegraphics[width=0.8\textwidth]{figure_suboverview_02_1.jpg}
%    \caption{$G_I$, and $G_L$ are obtained through encoder $q_\phi$ (encoder equivariance condition).
    %Fig.~\ref{sub fig: partially equivaraince} 
%    The left side figure shows the relation between each space and symmetries.
    %If $q_\phi$, and 
%    If $\psi(\cdot)$ is equivariant function over all $G_L$, and $G_T$, then %there exist $\Xi$, and 
%    there exist $\Gamma$, where $\Gamma: G_L \rightarrow G_T$, and $\Xi \circ \Gamma: G_I  \rightarrow G_T$.
    %The same group is colored blue.
%    However, unrestricted $\psi(\cdot)$ has no guarantee to be partial- or full-equivariant.
%    The red arrows represent our method: L2L transformation guarantees $\Gamma^J: G_L^J \rightarrow G_T^J$, and $\Xi^J \circ \Gamma^J: G_I^J \rightarrow G_T^J$, given the encoder equivariance condition $\Xi: G_I \rightarrow G_L$.}
%    \label{sub fig:suboverview}
%\vspace{-10pt}
%\end{figure*}

\begin{proposition}
\label{prop1}
Any $\psi(\cdot) \in G_S$, notated as $\psi_{G_S}(\cdot)$, is equivariant to group $G_S$. 
\end{proposition}

$\textit{Proof.}$ 
The group $G_S$ is closed to matrix multiplication, and its element is always a symmetric matrix by definition. 
Then, any two elements in $G_S$ are commutative because if matrix multiplication of two symmetric matrices is symmetric then both are commutative by Lemma~\ref{lem:GS_abelian}. 
As a result, $\psi_{G_S}(\cdot)$ and group elements of $G_S$ are commutative ($G_S$ is an abelian group).
Because of the commutativity, $\psi_{G_S}(g_s\circ  \vz) = \mathbf{e}^{ \mS} g_s  \vz = g_s \mathbf{e}^{ \mS}  \vz = g_s \circ \psi_{G_S}( \vz)$ for $g_s\in G_S$ if the group action $\circ$ is set to matrix multiplication, where $\psi_{G_S} \in G_S$.
This equation satisfies the general definition of an equivariant function Eq.~\ref{equation: equivariant map} 
% that a function $f(\cdot)$ is equivariant if $f (g \circ  \vz) = g \circ f( \vz)$ for all $g$ in a group $G$ 
by matching $f$, $g$, and $G$ to $\psi_{G_S}$, $g_s$, and $G_S$, respectively. $\blacksquare$

\begin{proposition} 
\label{prop2}
If $q_\phi$ is equivariant over defined on group of symmetries $G_I^J$ and $G_L^J$, then $\psi_{G_S} (q_\phi(\cdot))$ is equivariant to symmetries in $G_I$ corresponding to $G_S \cap G_L$ ($G_I^J$) and $G_T$ corresponding to $G_S \cap G_L$ ($G_T^J$) by the equivariance of $q_\phi$.  
\end{proposition}
$\textit{Proof.}$ The function $\psi_{G_S}(\cdot)$ is an equivariant function over group elements in $G_S \cap G_L$ by Proposition~\ref{prop1}. 
Then, the composite function, $\psi_{G_S}(\cdot)$ and $q_\phi$, is an equivariant function of $G_I^J$ and $G_T^J$.
Let $g_L^J \in G_L^J$, and $g_T^J \in G_T^J$.
% of $G_I$ corresponding to $G_S \cap G_L$ and $G_T$ corresponding to $G_S \cap G_L$.
% Let $g_L^J$ be a group element in $G_S \cap G_L$, and $g_I^J$ is a group element in $G_I$ corresponding to $G_S \cap G_L$.
Then, $\psi_{G_S}(g_L^J  \vz_1) = g_L^J \psi_{G_S}( \vz_1) = g_T^J \psi_{G_S}( \vz_1)$.
$\blacksquare$

% \begin{figure}
%   \begin{center}
%     \includegraphics[width=0.35\textwidth]{figure_equivariance_ver01.jpg}
%   \end{center}
%   \caption{Equivariant map: $\mathcal{X}$, $\mathcal{Z}$, and $\hat{\mathcal{Z}}$ are input space, latent vector space, and transformed latent vector space by L2L transfomration function $\psi(\cdot): \mathbb{R}^n \rightarrow \mathbb{R}^n$. respectively. $ \vx \in \mathcal{X}$, $ \vz \in \mathcal{Z}$, and $\hat{ \vz} \in \hat{\mathcal{Z}}$.}
%   \label{figure: equivalence cycle}
% \end{figure}

Therefore, the invertible and symmetric matrix is a partial-equivariant function between the data space and transformed latent vector space.
More details are in Proposition~\ref{appendix: prop2}.

% Let $P(B)$ be the probability of $\psi(\cdot) \in B$ for a subset $B \subset M_n(\mathbb{R})$ after VAE training, and $Pr(\psi_B \in B^\prime)$ be the conditional probability of $\psi(\cdot) \in B^\prime$ given $\psi(\cdot) \in B$.
Let $P(G):= \frac{\mu(G \cap K)}{\mu(K)}$ by the assumption 4 in section~\ref{sec:method}, and $Pr(\psi_{G^\prime}(\cdot) \in G) = P(\psi(\cdot) \in G | \psi(\cdot) \in G^\prime)$ after VAE training.
\begin{proposition} 
\label{prop3}
$Pr(\psi_{E_S}(\cdot) \in G_S) > Pr(\psi_{E_M}(\cdot) \in G_S) > Pr(\psi_{M}(\cdot) \in G_S)$.
\end{proposition}

$\textit{Proof.}$ 
All $\mathbf{e}^{ \mS} \in E_S$ are in $E_M$ since $Sym_n(\mathbb{R}) \subset M_n(\mathbb{R})$. However, $E_M \not\subset E_S$ because $\mathbf{e}^{ \mS} $ is always symmetric, but $\mathbf{e}^{ \mM}$ can be an asymmetric matrix.
% ~\ref{}.
Therefore $E_M \not\subset E_S$.
Therefore, the conditional probability $Pr(\psi_{E_S}(\cdot) \in G_S) = \frac{P(G_S)}{P(E_S)}$ is greater than $Pr(\psi_{E_M}(\cdot) \in G_S) = \frac{P(G_S)}{P(E_M)}$. In the same way, $Pr(\psi_{E_M}(\cdot) \in G_S) > Pr(\psi_{M}(\cdot) \in G_S) = \frac{P(G_S)}{P(M_n(\mathbb{R}))}$ because $E_M \subset M_n(\mathbb{R})$ and non-invertible functions are only in $M_n(\mathbb{R})$. 
More details are in Proposition~\ref{appendix: prop3}. $\blacksquare$ 

% Therefore, $\psi_{E_S}$ clearly increases the probability of preserving a certain type of equivariance compared to unrestricted $\psi$ functions. 
Therefore, $\psi_{E_S}$ clearly increases the probability of Eq.~\ref{equation: argmax}, because the group $G_L$ is obtained ($|\hat{\mathcal{Z}}^J(\psi)|$ is fixed) by our assumption that $q_\phi(\cdot)$ is a partial-equivariant.
It implies that symmetric and invertible properties for a matrix are necessary to preserve partial equivariance.

The probability $Pr(\psi_{E_S}(\cdot) \in G_S)$, $Pr(\psi_{E_M}(\cdot) \in G_S)$, and $Pr(\psi_{M}(\cdot) \in G_S)$ is changed by the distribution of the observation of $\psi(\cdot)$, which depends on the model parameters.
However, the inequality $Pr(\psi_{E_S}(\cdot) \in G_S) > Pr(\psi_{E_M}(\cdot) \in G_S) > Pr(\psi_{M}(\cdot) \in G_S)$ is not changed regardless of the distribution of observation of $\psi(\cdot)$.
We empirically validate the impact of equivariance with the uncertain $P(\cdot)$ to disentanglement in Section~\ref{subsec: discussion}.

% This correspondence of decomposition is expected to transfer the independence between dimensions of $ \vz$ to the space of $\hat{ \vz}$~\citep{disen_definitiaon_2}. 

\paragraph{Invertible Property by Using Matrix Exponential}
To guarantee the invertible property of IPE-transformation, we use a function $\psi(\cdot) = \mathbf{e}^{ \mM} \ast \cdot$
 for the transformation, where $ \mM$ is in  $n\times n$ real number matrix set $M_n (\mathbb{R})$~\citep{invertible-matrix-exponential}.  
The operator $\ast$ is matrix multiplication, and $\mathbf{e}^{ \mM} = \sum_k^{\infty} \frac{ \mM^k}{k!}$.
Our motivation is to use the benefits of injecting explicit inductive bias for disentanglement~\citep{vae-inductive-bias, intel-vae}.
InteL-VAE effectively extracts hierarchical representation, which includes low-level features (affect to a specific factor) and high-level features (affect to complex factors) with an invertible transformation function~\citep{intel-vae}.

\subsection{Exponential Family Conversion for Unknown Prior}
\label{subsec:ef based vae}
In VAE frameworks, the Gaussian normal distribution is applied as a prior.
However, a prior from data is usually unknown and may not follow the Gaussian distribution~\citep{intel-vae}.
As a solution, we present a training procedure for VAEs to build an exponential family distribution from a latent variable of an arbitrary distribution.
Then, we introduce training losses obtained from the unit IPE-transformation function and EF-conversion.

\paragraph{Elements of Exponential Family Distribution Settings}
As introduced in~\citet{prml, exp_family_bayesian}, we assume exponential family always admits a conjugate prior Eq.~\ref{eq:conju_prior}.
To determine the \textit{natural parameter} of posterior and prior $\boldsymbol{\theta}_{\hat{ \vz}_m}$, and $\hat{\boldsymbol{\epsilon}}_m$, we use a natural parameter generator (NPG) designed by multi-layer perceptron~\citep{exp_family_bayesian}. 
However, generated natural parameter $\boldsymbol{\theta}_{\hat{ \vz}_m}$ is not guaranteed as the appropriate parameter of the exponential family corresponds to conjugate prior. 
To satisfy this condition, we assume observation is a set of independent identically distributed, then we follow Eq.~\ref{eq: likelihood}.
% To satisfy this condition, we assume observation is a set of independent identically distributed, then Eq.~\ref{eq:pdf} is modified: $p(\mathbf{X}|\boldsymbol{\theta}) = \prod_{n=1}^N h(\mathbf{x}_n) \text{exp} (\boldsymbol{\theta}^{\intercal} \sum_{n=1}^N T(\mathbf{x}_n) -A(\boldsymbol{\theta}))$~\citep{prml}, where observation $\mathbf{X}=\{\mathbf{x}_1, \cdots \mathbf{x}_N \}$.
% In the next, we multiply the modified formation by the prior Eq.~\ref{eq:conju_prior} to obtain the posterior distribution~\citep{prml} as Eq.~\ref{eq:posteior_}.

\paragraph{Distribution Approximation As an Exponential Family} 
By our exponential family settings, a posterior distribution in the exponential family by adopting the following Eq.~\ref{eq:posteior_}.
% The procedure represents a posterior distribution in the exponential family by adopting the following form:
% \begin{equation}
% \small
%     p(\boldsymbol{\theta} | \mathbf{X}, \mathcal{\xi}, \boldsymbol{\nu}) \propto \text{exp} (\boldsymbol{\theta}^\intercal (\sum_{n=1}^N T(\mathbf{x}_n) + \boldsymbol{\nu} \mathcal{\xi}) - A(\boldsymbol{\theta})),
%     \label{eq:posteior_}
% \end{equation}
% where \textit{sufficient statistics} $T(\cdot)$ and \textit{log-normalizer}, $A(\cdot)$ are known functions, samples $\mathbf{X} = \{\mathbf{x}_1, \mathbf{x}_2, \ldots, \mathbf{x}_n \}$ from distribution, and \textit{natural parameter} of posterior $\boldsymbol{\theta}$ and of prior $\mathcal{\xi}$~\citep{prml}. 
% The functions $T(\cdot)$, and $A(\cdot)$ are deterministic functions to maximize posterior distribution.
The \textit{evidence} is implemented as learnable parameters $\boldsymbol{\nu} \in \mathbb{R}^{n \times n}$, and natural parameter is generated by a multi-layer perceptron as~\citet{exp_family_bayesian}.
This general form approximating an exponential family distribution with learnable parameters can extend VAEs to use a wider distribution for latent variables by simply matching $\mathbf{X}$ to generated latent variables. 
After IPE-transformation, we can apply the form by using the $\hat{ \vz}_m$, $\boldsymbol{\theta}_{\hat{ \vz}_m}$, and $\boldsymbol{\theta}_{\hat{\boldsymbol{\epsilon}}_m}$ for $\mathbf{X}$, $\boldsymbol{\theta}$, and $\mathcal{\xi}$, respectively as shown in Fig.~\ref{fig:overview}.

\paragraph{EF Similarity Loss}
To maximize Eq.~\ref{eq:posteior_} to be an exponential family, we added a loss through a Lagrange multiplier as:
% We enforce an exponential-family posterior by adding a Lagrange multiplier to the objective in Eq.~\ref{eq:posteior_} as follows:
\begin{equation}
\scalebox{0.75}{$
\begin{aligned}
\text{maximize} &~\log p(\boldsymbol{\theta}_{\hat{ \vz}_m} | \hat{ \vz}_m, \boldsymbol{\theta}_{\hat{\boldsymbol{\epsilon}}_m}, \boldsymbol{\nu}_m) ~~~\text{s.t.} ~ 
\KL(f_{ \vx}( \vx|\boldsymbol{\theta}_{\hat{ \vz}_m}) || f_{ \vx}( \vx|\boldsymbol{\theta}_{\hat{\boldsymbol{\epsilon}}_m})) \geq 0
\end{aligned}
$}
\label{eq:reg_cons}
\end{equation}
\vspace{-3mm}
\begin{equation}
% \scriptsize
    \begin{split}
        \Rightarrow \mathcal{L}_{s}(\hat{ \vz}_m, \hat{\boldsymbol{\epsilon}}_m) &= \log p(\boldsymbol{\theta}_{\hat{ \vz}_m} | \hat{ \vz}_m, \boldsymbol{\theta}_{\hat{\boldsymbol{\epsilon}}_m}, \boldsymbol{\nu}_m) \\ 
        & + \boldsymbol{\lambda}_m %\textstyle\sum_{m=1}^{M}
     \KL(f_{ \vx}( \vx|\boldsymbol{\theta}_{\hat{ \vz}_m}) || f_{\mathbf{x}}( \vx|\boldsymbol{\theta}_{\hat{\boldsymbol{\epsilon}}_m}))
    \end{split}
    \label{eq:reg}
\end{equation}
\vspace{-3mm}
\begin{equation}
    \begin{split}
        \Rightarrow \mathcal{L}_{el} \coloneqq ||  \nabla_{\hat{ \vz}_m, \hat{\boldsymbol{\epsilon}}_m, \boldsymbol{\lambda}_m} \mathcal{L}_{s} ||_2^2.
        \label{eq:reg_norm}
    \end{split}
\end{equation}
The notation $\boldsymbol{\theta}_{k}$ is a generated natural parameter by a given $k \in \{ \hat{ \vz}, \hat{\boldsymbol{\theta}} \}$, and $f_{ \vx}( \vx|\boldsymbol{\theta})$ is a power density function of the exponential family.
Moreover, $\boldsymbol{\lambda}_m$ is a trainable parameter for optimizing the Lagrange multiplier, and %$\sum_{m=1}^{M} 
$ \KL(f_{ \vx}( \vx|\boldsymbol{\theta}_{\hat{ \vz}_m}) || f_{ \vx}( \vx|\boldsymbol{\theta}_{\hat{\boldsymbol{\epsilon}}_m}))$ is a KL divergence of the exponential family.

\paragraph{KL Divergence for Evidence of Lower Bound} 
% The KL divergence of Gaussian distribution~\citep{vae} is computed using mean and variance, which are the parameters of a Gaussian distribution. 
% To introduce a loss as the KL divergence of Gaussian distribution,
We compute the KL divergence of the exponential family in Eq.~\ref{eq:reg} using the learnable parameter $T(\cdot)$ and $A(\cdot)$ with given natural parameter $\boldsymbol{\theta}_{\hat{ \vz}}$ and $\boldsymbol{\theta}_{\hat{\boldsymbol{\epsilon}}}$, expressed as:
\begin{equation}
\small
    \begin{split}
         \mathcal{L}_{kl} &\coloneqq  \KL(f_{ \vx}( \vx|\boldsymbol{\theta}_{\hat{ \vz}_m}) || f_{ \vx}( \vx|\boldsymbol{\theta}_{\hat{\boldsymbol{\epsilon}}_m})) \\
        &= A(\boldsymbol{\theta}_{\hat{\boldsymbol{\epsilon}}}) - A (\boldsymbol{\theta}_{\hat{ \vz}}) + \boldsymbol{\theta}_{\hat{ \vz}}^{\intercal}  \nabla_{\boldsymbol{\theta}_{\hat{ \vz}}} A(\boldsymbol{\theta}_{\hat{ \vz}}) - \boldsymbol{\theta}_{\hat{\boldsymbol{\epsilon}}}^{\intercal}  \nabla_{\boldsymbol{\theta}_{\hat{\boldsymbol{\epsilon}}}} A(\boldsymbol{\theta}_{\hat{\boldsymbol{\epsilon}}}).
        \label{eq:kl}
    \end{split}
\end{equation}
Further details are in the Appendix~\ref{appendix: kl lower bound}.

\paragraph{KL Divergence Calibration Loss}
Although $\mathrm{KL}$ is invariant under a common invertible map $\psi$~\citep{intel-vae}, practical approximations (e.g., matrix-exponential) can break this. 
We therefore add a consistency regularizer that penalizes the gap between empirical KLs before and after applying $\psi$:
% Although $\mathrm{KL}$ is invariant to a common invertible map $\psi$~\citep{intel-vae}, implementation artifacts, and approximation of matrix exponential violate it. 
% We therefore add a small consistency regularizer that penalizes the gap between empirical KLs before and after applying $\psi$ as:
% Although the KL divergence is invariant under the same invertible map $\psi$ applied to both $q$ and $p$~\citep{intel-vae}, practical approximations, matrix exponential numerics, MIPE averaging (non-invertible), the implicit mask’s nonlinearity, and learned EF modules with mini-batch estimates—can break this invariance. 
% We therefore add a lightweight consistency regularizer that penalizes the discrepancy between empirical KLs before and after applying $\psi$:
% To reduce the error between the approximation and true matrix for the matrix exponential~\citep{matrix_exponential_approx}, we add a loss to minimize the difference of their KL divergence measured by mean squared error (MSE) as: 
% \begin{equation}
% \scriptsize
% \begin{split}
%    \mathcal{L}_{cali} =\text{MSE}(&  \KL (q_\phi( \vz|
%  \vx)||p_\theta( \vz)),  \KL (f_{ \vx}( \vx|\boldsymbol{\theta}_{\hat{ \vz}_m}) || f_{ \vx}( \vx|\boldsymbol{\theta}_{\hat{\boldsymbol{\epsilon}}_m}))),% ||_2^2,
% \label{eq:mse} 
% \end{split}
% \end{equation}
\begin{equation}
\scalebox{0.8}{$
\begin{aligned}
   \mathcal{L}_{\text{cali}} = \text{MSE}( \KL (q_\phi( \vz | \vx) \, || \, p_\theta( \vz)), \KL (f_{ \vx}( \vx | \boldsymbol{\theta}_{\hat{ \vz}_m}) \, || \, f_{ \vx}( \vx | \boldsymbol{\theta}_{\hat{\boldsymbol{\epsilon}}_m})))
\end{aligned}
$}
\label{eq:mse}
\end{equation}
which is the KL divergence calibration loss ($\mathcal{L}_{cali}$).

\begin{table*}[t]
    \caption{Performance (mean $\pm$ std) of four metrics on dSprites, 3D Shapes, and 3D Cars. 
    % CVAE settings, respectively.
    }
    \vspace{-5pt}
    \begin{center}
    {
    \resizebox{0.90\textwidth}{!}{%
    \begin{tabular}{@{~}c@{~}@{~}c@{~}@{~~}c@{~~}c@{~~}@{~~}c@{~~}c@{~~}c@{~~}c@{~~}c@{~~}c@{~~}c@{~}}
        \hline
        % \multirow{2.5}{*}{Dataset} & 
        \multirow{2.5}{*}{dSprites} & \multicolumn{2}{c}{FVM $\uparrow$} & \multicolumn{2}{c}{MIG $\uparrow$} & \multicolumn{2}{c}{SAP $\uparrow$} & \multicolumn{2}{c}{DCI $\uparrow$} \\
        \cmidrule(lr){2-3} \cmidrule(lr){4-5} \cmidrule(lr){6-7} \cmidrule(lr){8-9}
        & original & MIPET & original & MIPET & original & MIPET & original & MIPET \\
        \cmidrule(lr){1-1} \cmidrule(lr){2-3} \cmidrule(lr){4-5} \cmidrule(lr){6-7} \cmidrule(lr){8-9} 
        % \multirow{5}{*}{dSprites}
        $\beta$-VAE & 69.15($\pm$5.88) & \textbf{74.19}($\pm$5.62) & 9.49($\pm$8.30) & \textbf{19.72}($\pm$11.37) & 2.43($\pm$2.07) & \textbf{5.08}($\pm$2.90) & 18.57($\pm$12.41) & \textbf{28.81}($\pm$10.19) \\
        $\beta$-TCVAE & 78.50($\pm$7.93) & \textbf{79.87}($\pm$5.80) & 26.00($\pm$9.06) & \textbf{35.04}($\pm$4.07) & 7.31($\pm$0.61) & \textbf{7.70}($\pm$1.63) & 41.80($\pm$8.55) & \textbf{47.83}($\pm$5.01) \\
        Factor-VAE & 67.78($\pm$7.48) & \textbf{68.38}($\pm$8.55) & 14.67($\pm$10.40) & \textbf{19.31}($\pm$13.46) & 2.35($\pm$2.32) & \textbf{4.24}($\pm$3.36) & 22.58($\pm$8.50) & \textbf{30.18}($\pm$12.99) \\
        CLG-VAE & 79.06($\pm$6.83) & \textbf{81.80}($\pm$3.17) & 23.40($\pm$7.89) & \textbf{36.34}($\pm$5.55) & 7.37($\pm$0.96) & \textbf{8.03}($\pm$0.83) & 37.68($\pm$7.83) & \textbf{44.73}($\pm$5.11) \\
        Control-VAE & 62.36($\pm$8.62) & \textbf{67.71}($\pm$6.41) & 4.36($\pm$2.86) & \textbf{7.34}($\pm$4.10) & \textbf{2.11}($\pm$1.88) & 1.93($\pm$1.63) & 10.40($\pm$3.42) & \textbf{15.18}($\pm$4.61) \\
        \hline
    \end{tabular}
    }}
    \end{center}
    \vspace{-8pt}
\begin{center}
{
\resizebox{0.90\textwidth}{!}{%
\begin{tabular}{@{~}c@{~}@{~}c@{~}@{~~}c@{~~}c@{~~}@{~~}c@{~~}c@{~~}c@{~~}c@{~~}c@{~~}l@{~~}l@{~}}
    \hline
    \multirow{2.5}{*}{3D Shapes} & \multicolumn{2}{c}{FVM $\uparrow$} & \multicolumn{2}{c}{MIG $\uparrow$} & \multicolumn{2}{c}{SAP $\uparrow$} & \multicolumn{2}{c}{DCI $\uparrow$} \\
    \cmidrule(lr){2-3} \cmidrule(lr){4-5} \cmidrule(lr){6-7} \cmidrule(lr){8-9}
    & original & MIPET & original & MIPET & original & MIPET & original & MIPET \\
    \cmidrule(lr){1-1} \cmidrule(lr){2-3} \cmidrule(lr){4-5} \cmidrule(lr){6-7} \cmidrule(lr){8-9}
    $\beta$-VAE & 71.76($\pm$12.26) & \textbf{75.19}($\pm$8.16) & 37.33($\pm$22.34) & \textbf{47.37}($\pm$10.13) & 7.48($\pm$4.12) & \textbf{9.20}($\pm$2.44) & 52.07($\pm$17.92) & \textbf{54.95}($\pm$8.99) \\
    $\beta$-TCVAE & 76.62($\pm$10.23) & \textbf{80.59}($\pm$8.57) & 52.93($\pm$20.5) & \textbf{54.49}($\pm$9.44) & 10.64($\pm$5.93) & \textbf{11.58}($\pm$3.32) & 65.32($\pm$11.37) & \textbf{66.22}($\pm$7.32) \\
    Factor-VAE & 77.43($\pm$10.71) & \textbf{78.34}($\pm$8.35) & 24.23($\pm$26.13) & \textbf{48.59}($\pm$10.58) & 4.99($\pm$4.46) & \textbf{9.84}($\pm$2.73) & \textbf{68.68}($\pm$12.92) & 60.23($\pm$9.90) \\
    CLG-VAE & 77.04($\pm$8.22) & \textbf{80.17}($\pm$8.43) & 49.74($\pm$8.18) & \textbf{53.87}($\pm$7.41) & 9.20($\pm$2.44) & \textbf{12.83}($\pm$3.01) & 57.70($\pm$8.60) & \textbf{60.74}($\pm$7.77) \\
    Control-VAE & 71.05($\pm$14.35) & \textbf{71.89}($\pm$8.33) & 24.88($\pm$13.68) & \textbf{32.28}($\pm$10.74) & 6.60($\pm$3.59) & \textbf{7.14}($\pm$2.09) & 40.08($\pm$13.45) & \textbf{43.06}($\pm$8.68) \\
    \hline
\end{tabular}
}}
\end{center}
    \vspace{-8pt}
    \begin{center}
{
\resizebox{0.90\textwidth}{!}{%
\begin{tabular}{@{~}c@{~}@{~}c@{~}@{~~}c@{~~}c@{~~}@{~~}c@{~~}c@{~~}c@{~~}c@{~~}c@{~~}c@{~~}c@{~}}
    \hline
    \multirow{2.5}{*}{3D Cars} & \multicolumn{2}{c}{FVM $\uparrow$} & \multicolumn{2}{c}{MIG $\uparrow$} & \multicolumn{2}{c}{SAP $\uparrow$} & \multicolumn{2}{c}{DCI $\uparrow$} \\
    \cmidrule(lr){2-3} \cmidrule(lr){4-5} \cmidrule(lr){6-7} \cmidrule(lr){8-9}
    & original & MIPET & original & MIPET & original & MIPET & original & MIPET \\
    \cmidrule(lr){1-1} \cmidrule(lr){2-3} \cmidrule(lr){4-5} \cmidrule(lr){6-7} \cmidrule(lr){8-9}
    $\beta$-VAE & \textbf{89.48}($\pm$5.22) & 88.95($\pm$5.94) & 6.90($\pm$2.70) & \textbf{7.27}($\pm$1.99) & 1.30($\pm$0.48) & \textbf{1.88}($\pm$1.12) & \textbf{19.85}($\pm$4.87) & 18.90($\pm$4.49) \\
    $\beta$-TCVAE & 95.84($\pm$3.40) & \textbf{96.43}($\pm$2.42) & \textbf{11.87}($\pm$2.90) & 10.80($\pm$1.22) & 1.55($\pm$0.38) & \textbf{1.88}($\pm$1.12) & \textbf{27.91}($\pm$4.31) & 26.08($\pm$2.47) \\
    Factor-VAE & 89.64($\pm$2.62) & \textbf{93.66}($\pm$3.83) & \textbf{10.63}($\pm$1.30) & 9.43($\pm$1.12) & 2.51($\pm$0.37) & \textbf{3.28}($\pm$0.83) & \textbf{26.27}($\pm$3.17) & 23.13($\pm$0.72) \\
    CLG-VAE & 86.11($\pm$7.12) & \textbf{91.06}($\pm$5.09) & 6.19($\pm$2.42) & \textbf{8.51}($\pm$2.11) & \textbf{2.06}($\pm$0.60) & 1.99($\pm$0.93) & 16.91($\pm$4.01) & \textbf{18.31}($\pm$2.83) \\
    Control-VAE & 88.76($\pm$7.66) & \textbf{89.10}($\pm$6.90) & 4.68($\pm$2.67) & \textbf{5.08}($\pm$2.68) & 1.16($\pm$0.74) & \textbf{1.45}($\pm$0.86) & 14.70($\pm$3.84) & \textbf{15.22}($\pm$4.15) \\
    \hline
\end{tabular}
}}
\end{center}
    \label{tab:quantitative analysis}
    \vspace{-10pt}
\end{table*}

\paragraph{Implicit Semantic Mask}
We propose an implicit semantic mask to improve disentanglement learning.
We apply mask matrix $\mathcal{M}$ which consists of 0 or 1 element to log-normalizer to prevent less effective weight flow as:
\begin{equation}
    %\begin{split}
    \mathcal{M}_{ij} = 
    \begin{cases}
        1 \text{     if } |\mathcal{W}_{ij}| \geq \mathbf{\mu}_{|\mathcal{W}_{ij}|} - \lambda \mathbf{\sigma}_{|\mathcal{W}_{ij}|} \\
        0 \text{     otherwise}
    \end{cases},
    %\end{split}
    \label{eq:mask}
\end{equation}
where $\mathcal{W}$ is the weight of log-normalizer, $\lambda$ is a hyper-parameter, $\mathbf{\mu}_{|\mathcal{W}_{ij}|}$, and $\mathbf{\sigma}_{|\mathcal{W}_{ij}|}$ are the mean, and standard deviation of weight respectively.
Previous work~\citep{Yang_2020_CVPR} utilizes a semantic mask in input space directly, but we inject the semantic mask implicitly on the latent space.

\subsection{Integration for Multiple IPE-Transformation and EF-Conversion}
\label{subsec:multiple unit}
We mathematically extend IPE-transformation to MIPE-transformation, which is the equivalent process of $\beta$-VAE to enhance disentanglement. 
Each IPE-transformation function operates independently, then the reconstruction error for objective function is defined as:
% \begin{equation}
% \scriptsize
%     \begin{split}
%        \mathcal{L}_{rec} &\coloneqq \frac{1}{k} \sum_{i=1}^k \Bigg [ \int q_i(\hat{ \vz}_i| \vx) \log p_\theta( \vx|\hat{ \vz}_i) \mathsf{d}\hat{ \vz}_i \prod_{j=1, j\neq i}^k \int q_j(\hat{ \vz}_j| \vx) %q_3(\mathbf{z}_3|\mathbf{x}) \cdots q_k(\mathbf{z}_k|\mathbf{x})
%        \text{d}\hat{ \vz}_j \Bigg ] \\
%        & = \frac{1}{k} \sum_{i=1}^k  E_{q_{\phi,\psi_i} ( \vz| \vx)} \log p_\theta( \vx|\psi_i( \vz)),
%     \end{split}
%     \label{eq: reconstruction error}
% \end{equation}
\begin{equation}
\scalebox{0.85}{$
\begin{aligned}
   \mathcal{L}_{\text{rec}} 
   % &\coloneqq \frac{1}{k} \sum_{i=1}^k \Bigg[ \int q_i(\hat{ \vz}_i| \vx) \log p_\theta( \vx|\hat{ \vz}_i) \, \mathrm{d}\hat{ \vz}_i 
   % \prod_{\substack{j=1 \\ j\neq i}}^k \int q_j(\hat{ \vz}_j| \vx) \, \mathrm{d}\hat{ \vz}_j \Bigg] \\
   & = \frac{1}{k} \sum_{i=1}^k \mathbb{E}_{q_{\phi,\psi_i} ( \vz| \vx)} \log p_\theta( \vx|\psi_i( \vz))
\end{aligned}
$}
\label{eq: reconstruction error}
\end{equation}
where $\hat{ \vz}_i = \psi_i( \vz)$.
Therefore, we define ELBO as:
\begin{equation}
\small
% \scriptsize
    \begin{split}
        \mathcal{L}^\prime(\phi, \theta, \psi_{i \in [1, k]};  \vx) & = \frac{1}{k} \sum_{i=1}^k  \E_{q_{\phi,\psi_i} ( \vz_i| \vx)}  \log p_\theta( \vx|\psi_i( \vz)) \\
        & -\sum_{i=1}^k  \KL (q_{\phi, \psi_i}( \vz| \vx) || p_{\psi_i}( \vz)).
    \end{split}
    \label{eq: elbo}
\end{equation}
However, following Eq.~\ref{eq: elbo}, k samples are generated, and each sample is disentangled for different factors.
We implement the output as the average of the sum of the k samples to obtain a single sample with a superposition effect from k samples.
Moreover, the KL divergence term in Eq.~\ref{eq: elbo} represents that increasing number of MIPE-transformation is equal to an increasing $\beta$ hyper-parameter in $\beta$-VAE~\citep{betaVAE} and more details are in Appendix~\ref{appendix: multiple unit}.

The VAEs equipped with MIPE-transformation (MIPET-VAEs) can be trained with the following loss:
\begin{equation}
%\footnotesize
\begin{split}
    %\scriptsize
    &\mathcal{L}(\phi, \theta, \psi_{i \in [1, k ]};  \vx) = \mathcal{L}_{rec} - \mathcal{L}_{kl}
    - \mathcal{L}_{el} -\mathcal{L}_{cali}.
\end{split}
\label{eq:objective_function}
\end{equation}

\begin{table*}[t]
\small
    \caption{Impact of the number of MIPE-transformation function on the $\beta$-TCVAE and $\beta$-VAE with dSprites, 3D Shapes, and 3D Cars datasets in terms of the four metrics.
    The blue and red box plots represent each model's single and multiple IPE-transformation cases, respectively (A-$n$: MIPET-$\beta$-TCVAE (4), B-$n$: MIPET-$\beta$-TCVAE (6), C-$n$: MIPET-$\beta$-VAE, $n$: the number of MIPE-transformation). Further details are in Table~\ref{appendix table :main impact of IE}.}
    \centering
    \begin{tabular}{>{\centering\arraybackslash}m{17mm}|>{\centering\arraybackslash}m{17mm}|>{\centering\arraybackslash}m{17mm}|>{\centering\arraybackslash}m{17mm}||>{\centering\arraybackslash}m{17mm}|>{\centering\arraybackslash}m{17mm}|>{\centering\arraybackslash}m{17mm}|>{\centering\arraybackslash}m{17mm}}
        \hline
         \multicolumn{4}{c||}{dSprites} & \multicolumn{4}{c}{3D Shapes} \\
         % \cline{2-5}
         \hline
         FVM & MIG & SAP & DCI & FVM & MIG & SAP & DCI \\
         \hline
         \includegraphics[trim=0 0 0 -5, width=0.09\textwidth]{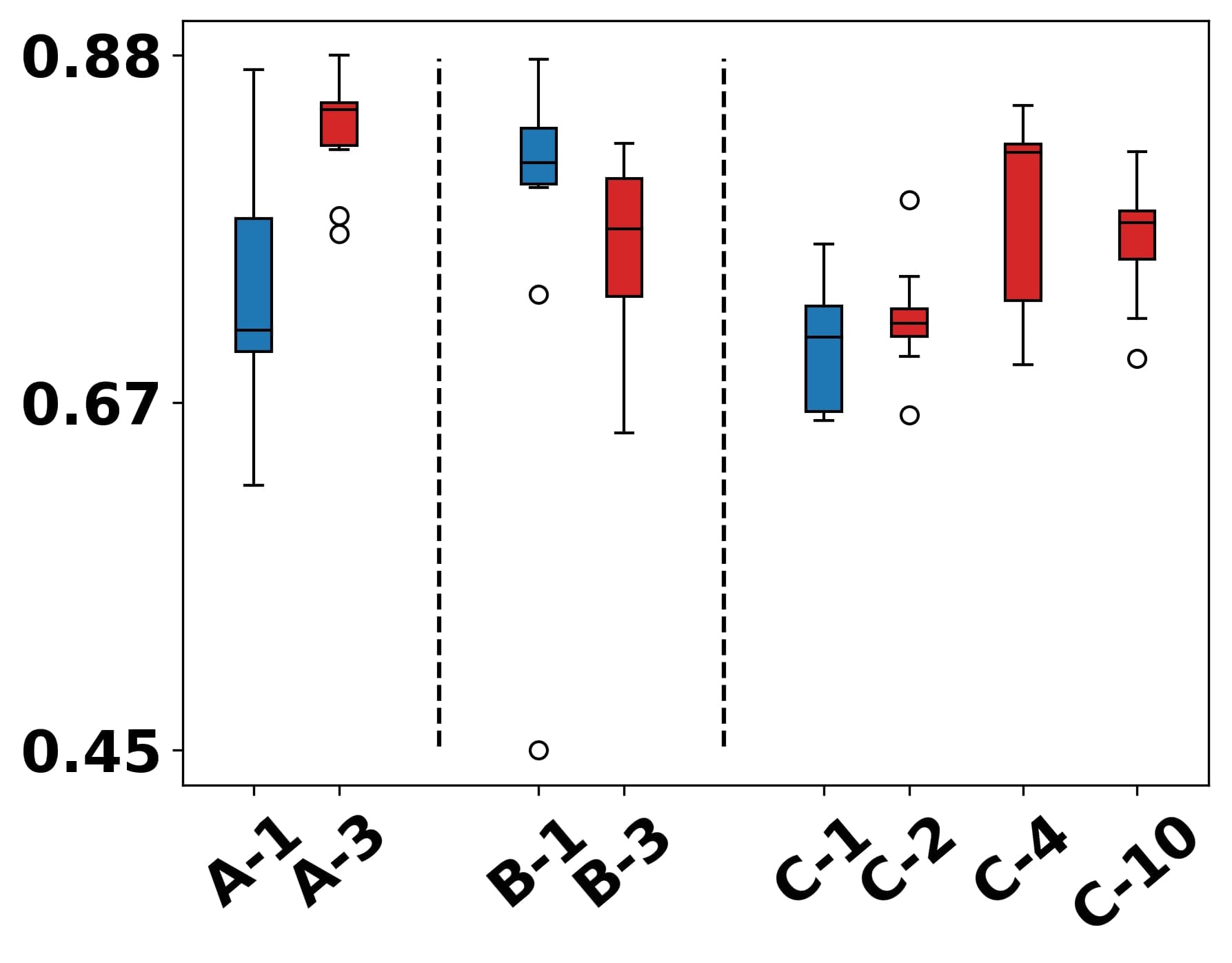} & \includegraphics[width=0.09\textwidth]{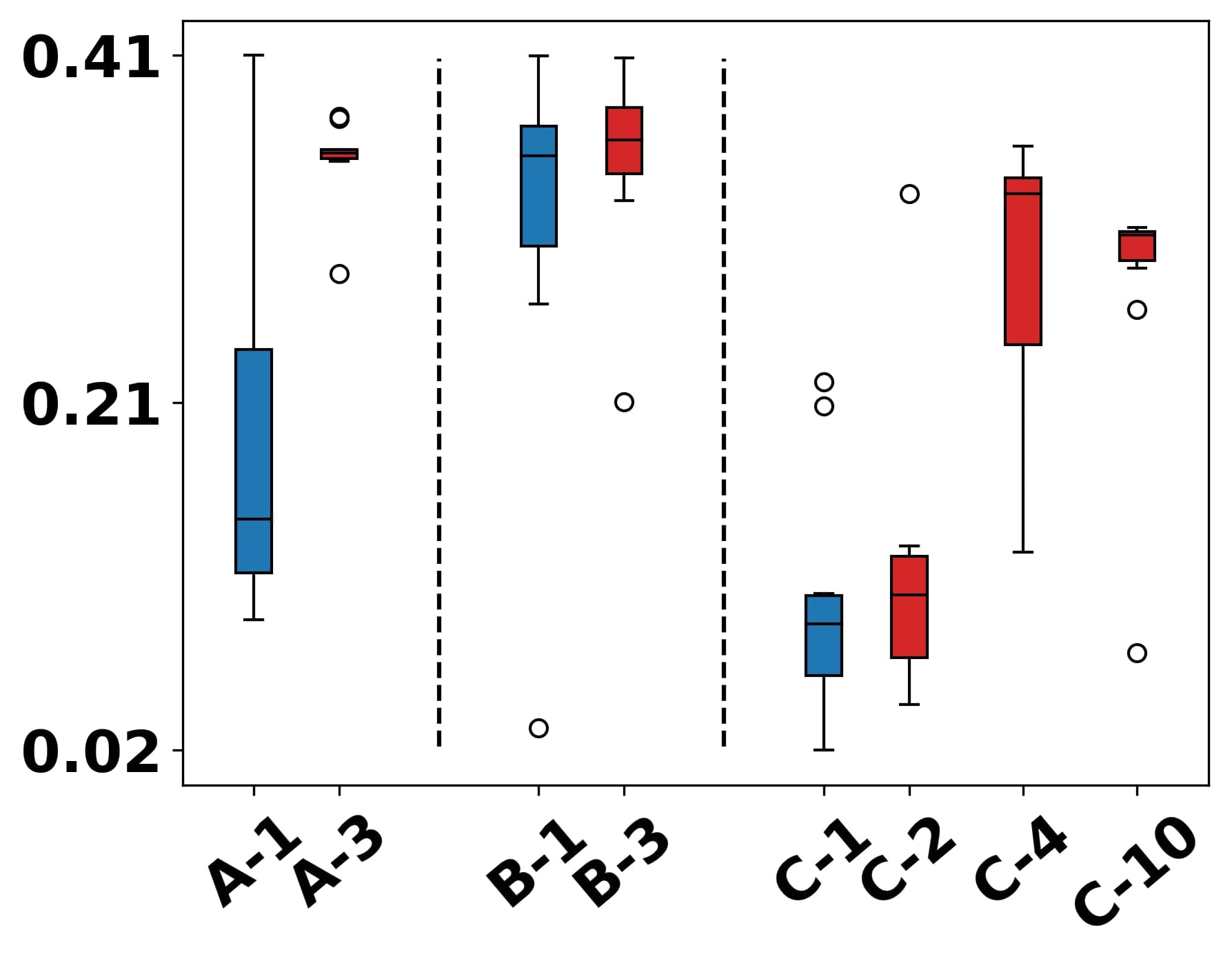} & \includegraphics[width=0.09\textwidth]{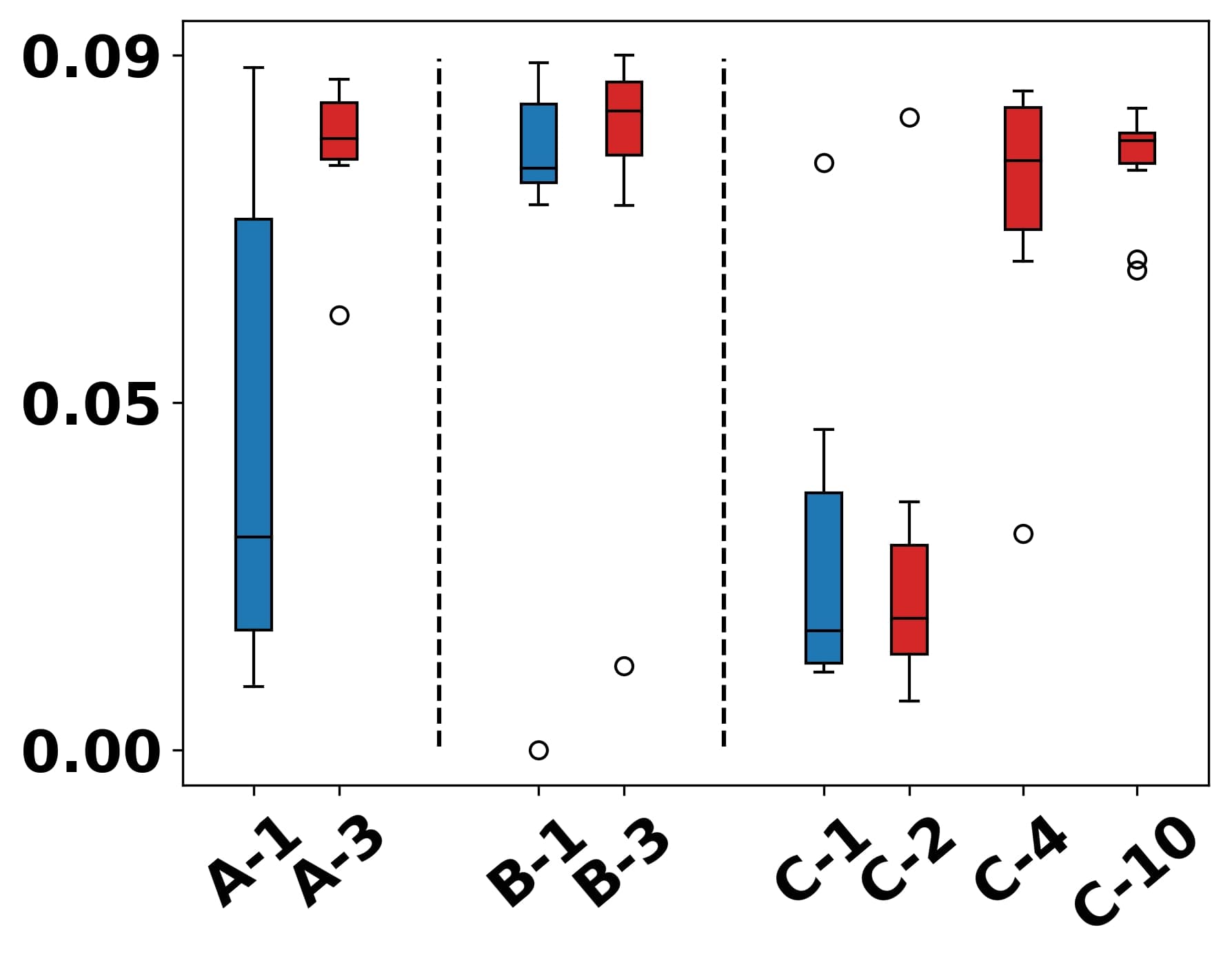} & \includegraphics[width=0.09\textwidth]{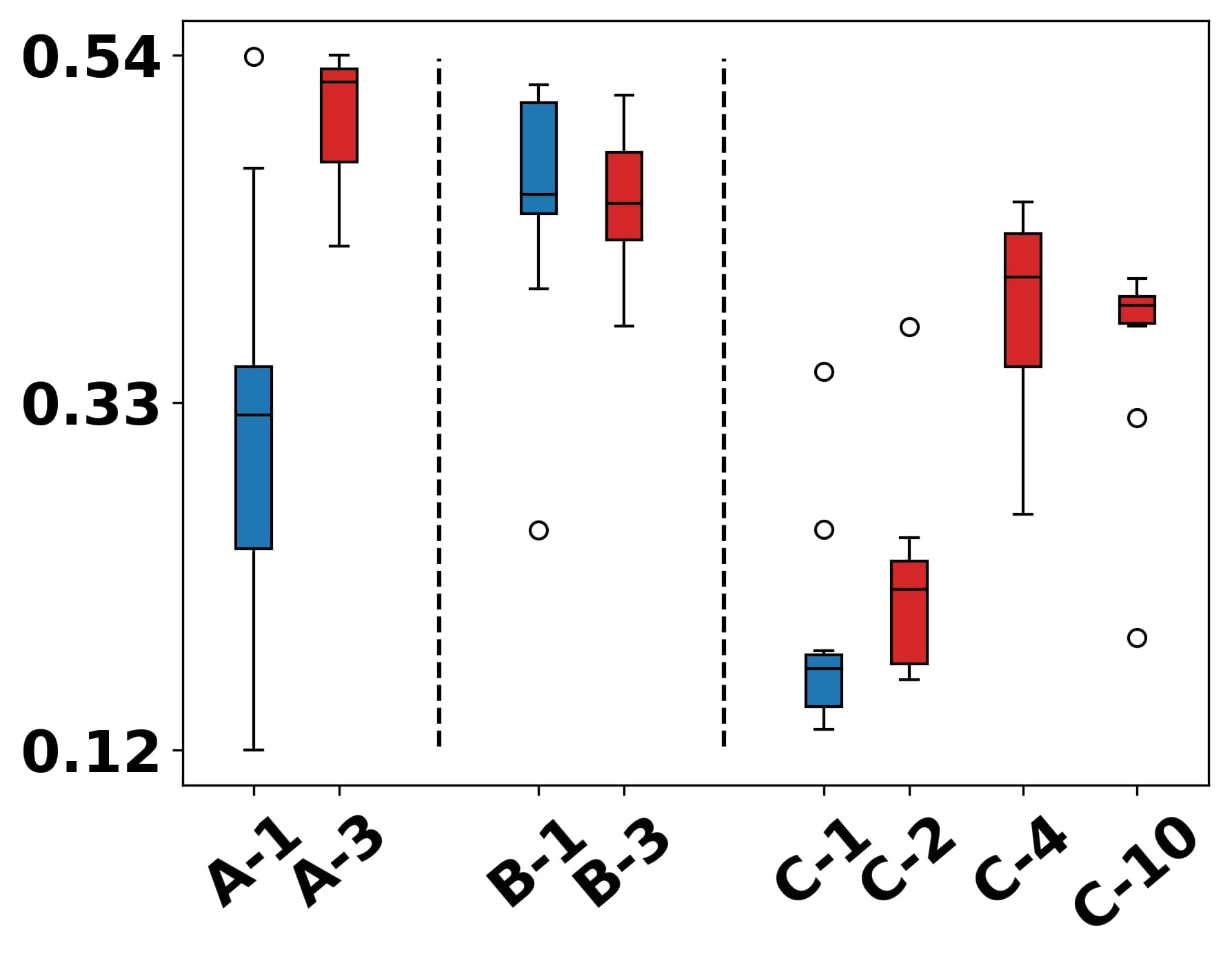} &
         \includegraphics[trim=0 0 0 -5, width=0.09\textwidth]{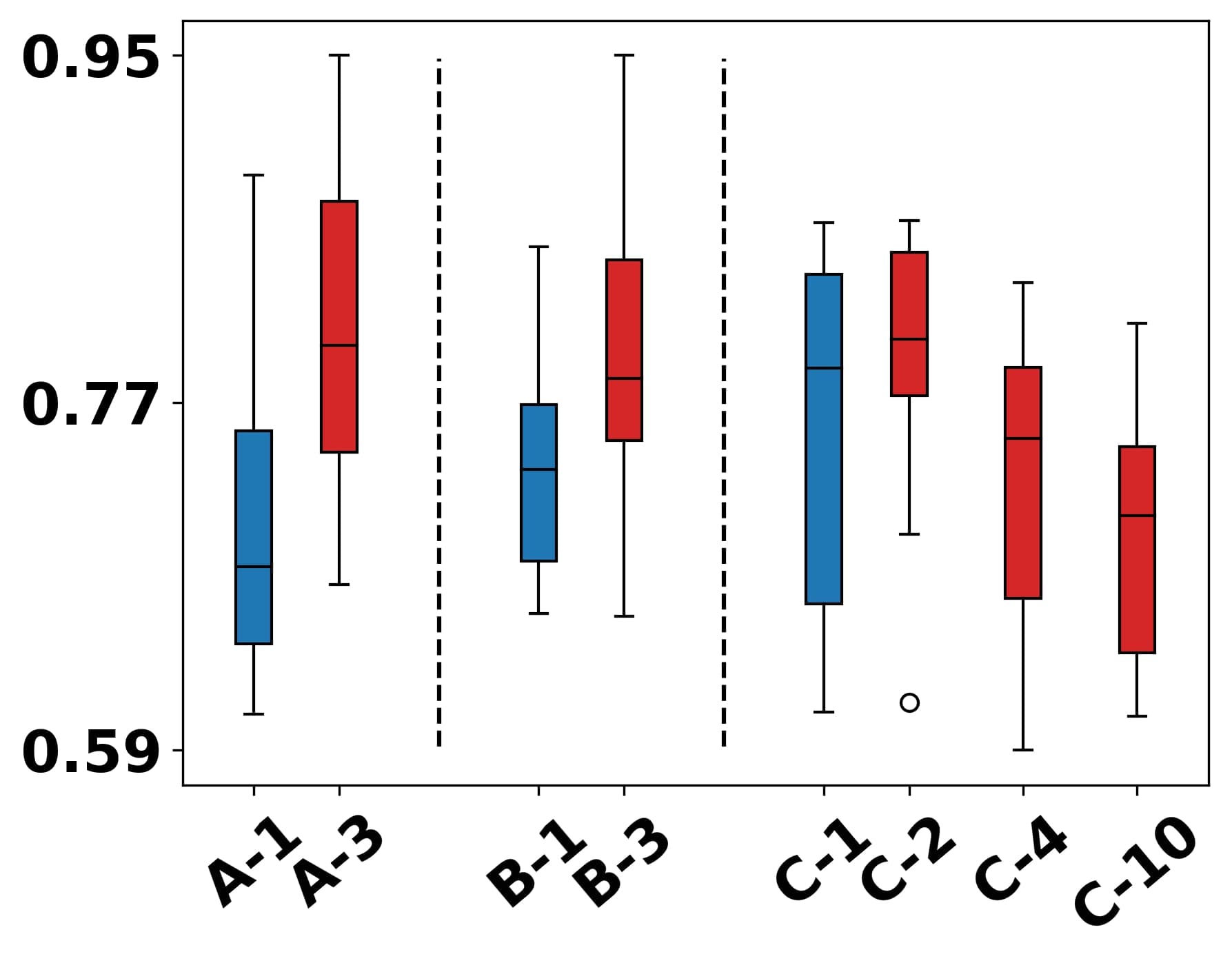} &
         \includegraphics[width=0.09\textwidth]{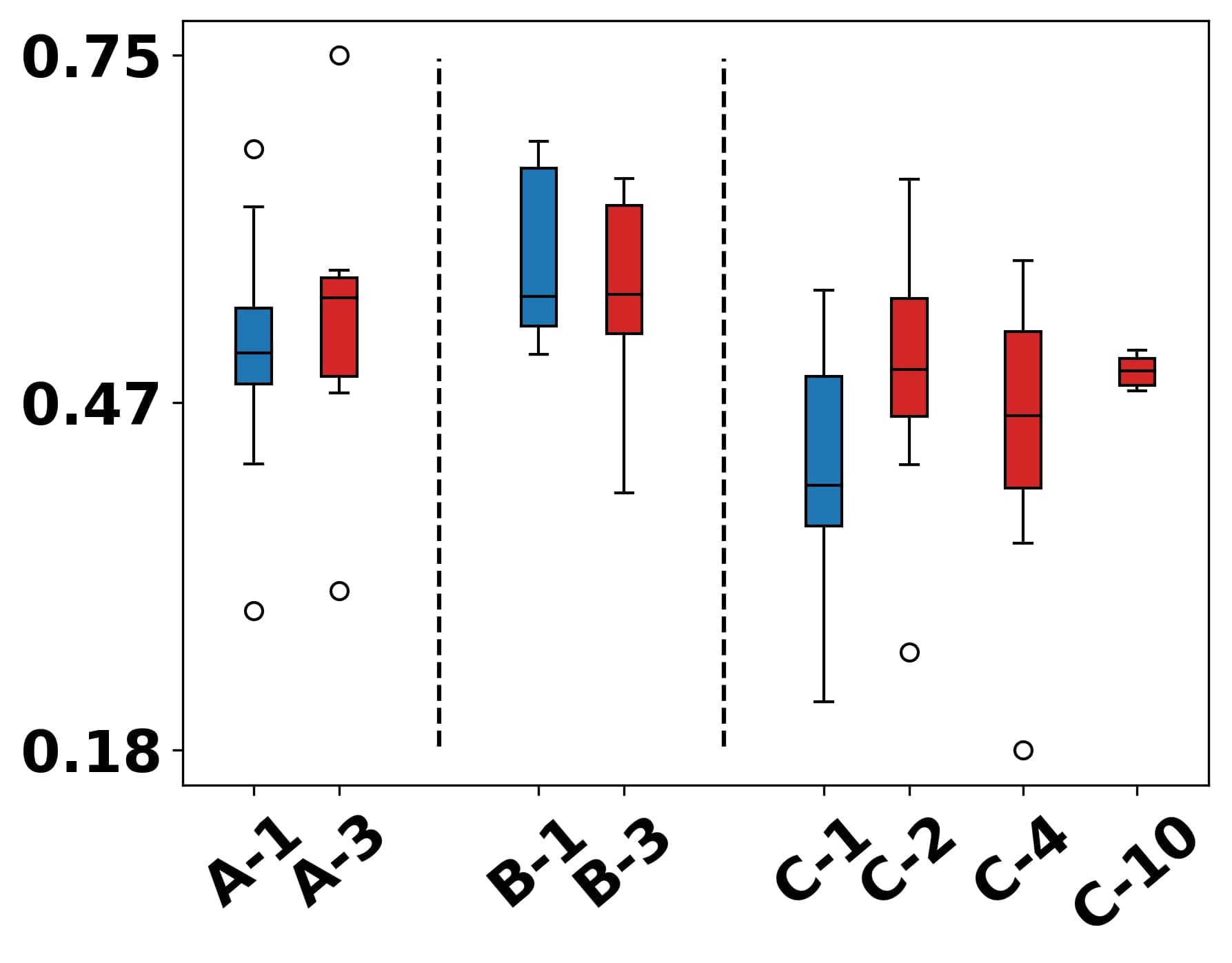} &
         \includegraphics[width=0.09\textwidth]{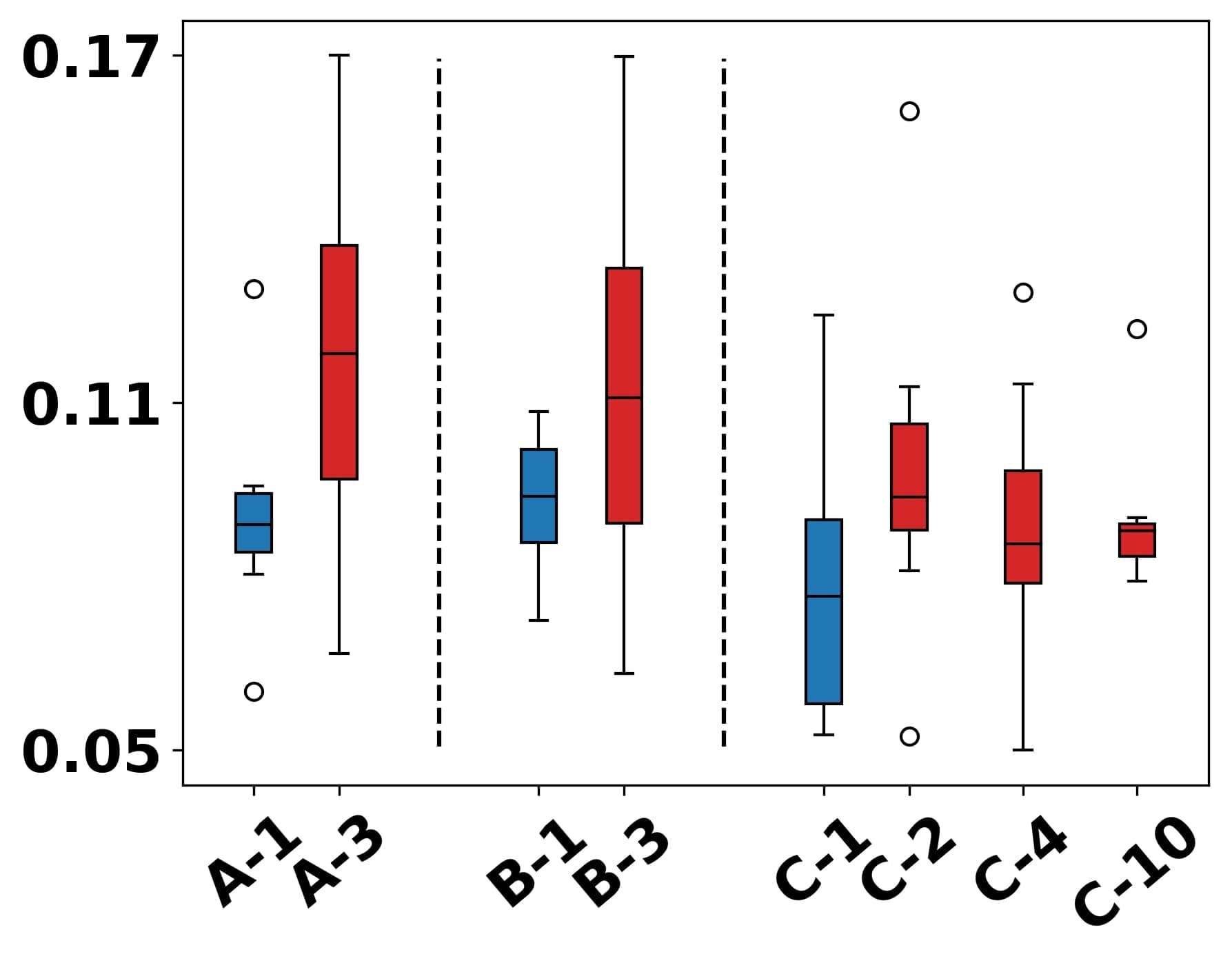} &
         \includegraphics[width=0.09\textwidth]{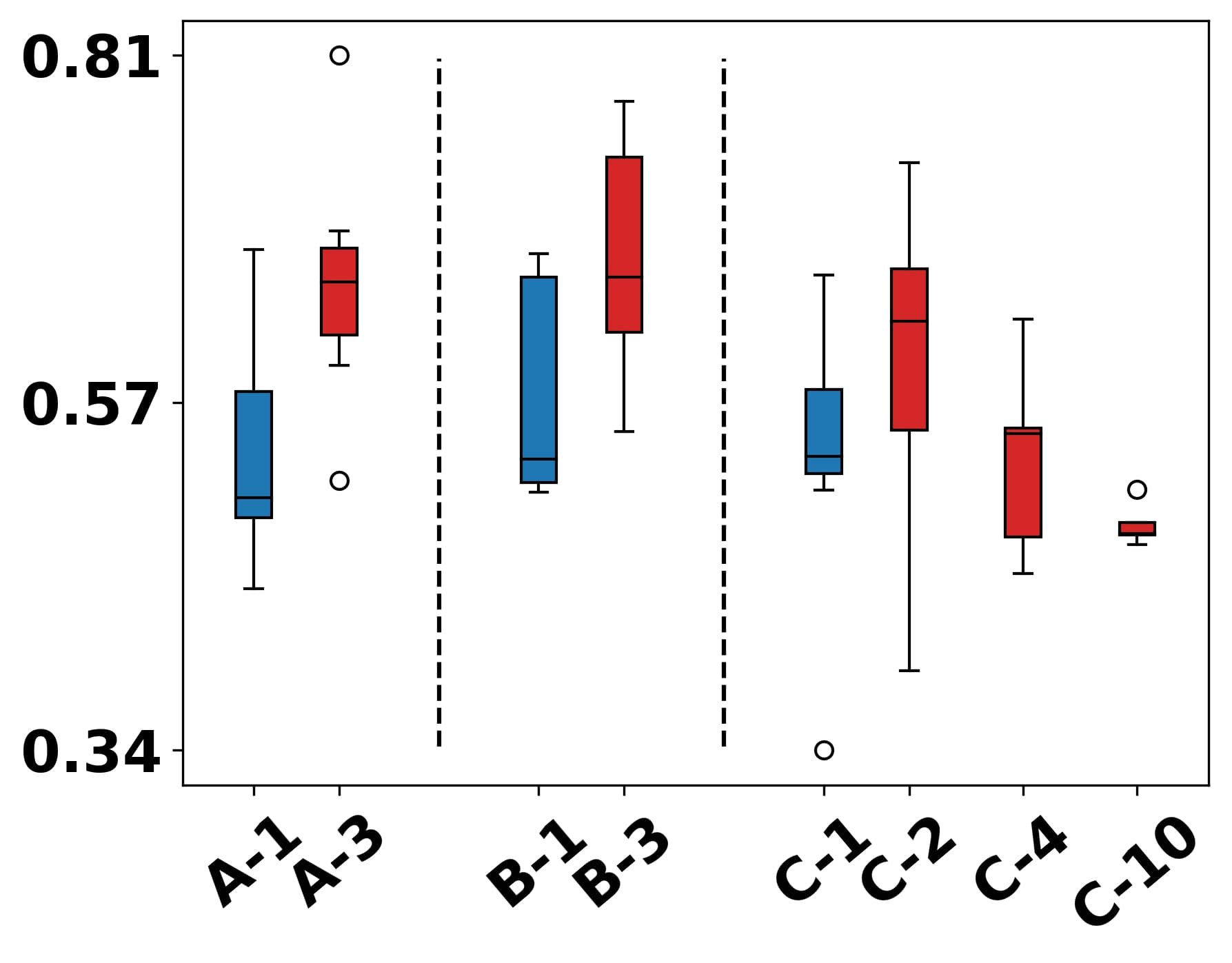}\\
         \hline
         % 3D Cars & 
         % \includegraphics[trim=0 0 0 -5, width=0.18\textwidth]{figure_3Dcar_FVM.jpg} &
         % \includegraphics[width=0.18\textwidth]{figure_3Dcar_MIG.jpg} &
         % \includegraphics[width=0.18\textwidth]{figure_3Dcar_SAP.jpg} &
         % \includegraphics[width=0.18\textwidth]{figure_3Dcar_DCI.jpg}\\
         % \hline 
    \end{tabular}
    \label{tab:main impact of IE}
% \vspace{-10pt}
\end{table*}

\begin{table*}[t]
    \caption{Impact of the mask (mean$\pm$std.) and its ratio $\lambda$ in Eq.~\ref{eq:mask} on 3D Cars.
    ($\infty$: no masking case, 
    %We set $\lambda$ from $\{0.0, 0.5, \ldots, 2.0, \infty \}$ for hyper-parameter tuning, and $\infty$ represents no masking case.
    gray box: the best setting over all metrics, bold text: the best in each metric.)
%    We represent best case for all four metrics as gray, and bold represents the highest value.
    Each model runs with ten random seeds.}
    \tiny
    \centering
    \begin{tabular}{c|c|c|c|c||c|c|c|c}
        \hline
        \multirow{1}{*}{ratio} & \multicolumn{4}{c||}{$\beta$-VAE (1)} & \multicolumn{4}{c}{CLG-VAE (0.5)}\\
        \cline{2-9}
        $\lambda$ & FVM $\uparrow$ & MIG $\uparrow$ & SAP $\uparrow$ & DCI $\uparrow$ & FVM $\uparrow$ & MIG $\uparrow$ & SAP $\uparrow$ & DCI $\uparrow$ \\
        \hline
        0.0 & 90.46($\pm$6.50) & 4.84($\pm$2.32) & 1.29($\pm$0.81) & 16.76($\pm$4.68) & \cellcolor{gray} \textbf{90.06}($\pm$4.44) & \cellcolor{gray}\textbf{9.28}($\pm$2.09) & \cellcolor{gray}1.82($\pm$0.82) & \cellcolor{gray}\textbf{19.12}($\pm$3.41) \\
        0.5 & 91.35($\pm$5.52) & 5.37($\pm$2.74) & 1.17($\pm$0.67) & 16.65($\pm$3.76) & 88.69($\pm$4.78) & 6.90($\pm$1.96) & 1.85($\pm$0.67) & 17.52($\pm$3.16) \\
        1.0 & \textbf{91.78}($\pm$6.20) & 4.99($\pm$2.27) & 1.36($\pm$0.81) & 16.50($\pm$2.53) & 83.60($\pm$11.48) & 8.12($\pm$3.66) & \textbf{2.37}($\pm$1.50) & 17.07($\pm$3.89) \\
        1.5 & \cellcolor{gray} 90.04($\pm$5.88) & \cellcolor{gray}\textbf{7.22}($\pm$2.87) & \cellcolor{gray} \textbf{1.36}($\pm$0.48) & \cellcolor{gray}\textbf{18.23}($\pm$2.84) & 84.76($\pm$6.86) & 7.70($\pm$2.11) & 2.05($\pm$0.73) & 17.06($\pm$2.77)\\
        2.0 & 87.79($\pm$8.88) & 4.75($\pm$2.49) & 1.01($\pm$0.99) & 16.64($\pm$3.75) & 85.78($\pm$4.18) & 7.83($\pm$1.79) & 1.91($\pm$0.96) & 17.26($\pm$2.07)\\ 
        \hline
        $\infty$ & 89.43($\pm$11.72) & 3.74($\pm$2.32) & 0.77($\pm$0.39) & 15.45($\pm$4.59) & 82.96($\pm$11.84) & 8.07($\pm$2.52) & 2.32($\pm$1.02) & 17.46($\pm$4.07)\\
        \hline
    \end{tabular}
    \label{tab:mask}
% \vspace{-10pt}
\end{table*}

% \begin{table*}[t]
%     \caption{
%     $\textit{p}$-value of t-test for original vs MIPET results of Table~\ref{tab:quantitative analysis}, which are averaged over models (bold: positive and significant, italic: positive but insignificant, normal: lower performance).}
%     \scriptsize
%     \centering
%     \begin{tabular}{c|c|c|c|c||c|c|c|c||c|c|c|c}
%     \hline
%          \multirow{2}{*}{$\textit{p}$-value}& \multicolumn{4}{c||}{VAEs} & \multicolumn{4}{c||}{CLG-VAE} & \multicolumn{4}{c}{$\beta$-TCVAEs}\\
%          \cline{2-13}
%          & FVM & MIG &SAP & DCI & FVM & MIG &SAP & DCI & FVM & MIG &SAP & DCI \\
%          \hline
         
%          dSprites & \textbf{0.000} & \textbf{0.000} & \textbf{0.000} & \textbf{0.000} & \textbf{0.030} & \textbf{0.000} & \textbf{0.005} & \textbf{0.000} & \textit{0.281} & \textbf{0.000} & \textit{0.170} & \textbf{0.009}\\
         
%          3D Shapes & \textbf{0.080} & \textbf{0.007} & \textbf{0.016} & \textit{0.191} & \textit{0.085} & \textbf{0.029} & \textbf{0.000} & \textit{0.088} & \textit{0.111} & \textit{0.383} & \textit{0.277} & \textit{0.390} \\
         
%          3D Cars & 0.659 & \textit{0.250} & \textbf{0.003} & 0.583 & \textbf{0.003} & \textbf{0.000} & 0.630 & \textit{0.071} & \textit{0.278} & 0.923 & \textit{0.119} & 0.933 \\
%          \hline
%     \end{tabular}
%     \label{tab:main statistical significant}
% \end{table*}
% \vspace{-10pt}
\section{Experiment Settings}
\label{sec:experiment}
\subsection{Models}
As baseline models, we select VAE~\citep{vae}, $\beta$-VAE~\citep{betaVAE}, $\beta$-TCVAE~\citep{beta-tcvae}, Factor-VAE~\citep{factor-vae}, Control-VAE~\citep{control-vae},and CLG-VAE~\citep{commutative-vae}. These models are compared to their extension to adopt MIPET, abbreviated by adding the MIPET prefix.
% We apply the proposed method to $\beta$-TCVAE only with the EF similarity loss term because $\beta$-TCVAE penalizes the divided KL divergence terms.
% We set the same encoder and decoder architecture in each model to exclude the overlapped effects.
Also, we follow the same model architecture which are introduced in previous works~\citep{factor-vae} and model details are in Table~\ref{tab:dsprites architecure}-\ref{tab:others architecure}.
%\hl{To check the impact of MIE-transformation, we do not consider the Groupified VAE because the latter is implemented with an extended decoder (different capacity).}
%Model architectures are in Table~\ref{tab:dsprites architecure}-\ref{tab:others architecure}, and Appendix~\ref{appendix: model details}.
%Groupified VAE is excluded from method comparison
%Groupified VAE is excluded from method comparison because of its decoder architecture incompatible to the common decoder. 
%the other decoders.  incompatible decoder architecture to  
%two factors affect to disentanglement learning, groupified method and 2) different architecture of decoder.
%because it is improper to evaluate the impact of intermediate transformation of the latent vector space by large changes of its decoder architecture,

\subsection{Datasets}
% As previous works~\cite{betaVAE, beta-tcvae, factor-vae, commutative-vae, groupified-vae, t-vae} have shown in synthetic datasets~\cite{dsprites17, 3d-car-dataset, 3dshapes18}, we also mainly demonstrate with synthetic datasets because disentangled representation is still a challenge in these datasets. 
% We compare well-known VAEs to MIPET-VAEs on the following data sets with 
As in prior works~\citep{betaVAE, beta-tcvae, factor-vae, commutative-vae, groupified-vae, t-vae}, we primarily evaluate our models on synthetic datasets, where learning disentangled representations remains a well-known challenge~\citep{dsprites17, 3d-car-dataset, 3dshapes18}. 
We benchmark MIPET-VAEs against several standard VAEs on the dSprites~\citep{dsprites17}, 3D Shapes~\citep{3dshapes18}, and 3D Cars~\citep{3d-car-dataset} datasets.
Further details are in the Appendix~\ref{appendix: datasets}
% We benchmark MIPET-VAEs against several standard VAEs on the following datasets: 1) dSprites~\cite{dsprites17} which consists of 737,280 binary $64 \times 64$ images of dSprites with five independent ground truth factors(number of values), $i.e.$ shape(3), orientation(40), scale(6), x-position(32), and y-position(32). 
% 2) 3D Shapes~\cite{3dshapes18} which consists of 480,000 RGB $64 \times 64 \times 3$ images of 3D Shapes with six independent ground truth factors: shape(4) orientation(15), scale(8), wall color(10), floor color(10), and object color(10).
% 3) 3D Cars~\cite{3d-car-dataset} which consists of 17,568 RGB $64 \times 64 \times 3$ images of 3D Shapes with three independent ground truth factors: car models(183), azimuth directions(24), and elevations(4).

\subsection{Training}
We set 256 mini-batch size in the datasets (dSprites, 3D Shapes, and 3D Cars), Adam optimizer with learning rate $4 \times 10^{-4}$, $\beta_1=0.9$, $\beta_2=0.999$, and epochs from $\{30, 67, 200\}$ as a common setting for all the comparative methods.
% For the comparison, we follow training and inference on the whole dataset.
% We train each model for 30, 67, and 200 epochs on the dSprites, 3D Shapes, and 3D Cars, respectively, as introduced in~\cite{factor-vae, 3d-car-setting}. 
% We tune $\beta$ from $\{1, 2, 4, 10\}$ and $\{4, 6 \}$ for $\beta$-VAE and $\beta$-TCVAE, respectively.
% The $\alpha=1$ and $\gamma=1$ of $\beta$-TCVAE as~\cite{beta-tcvae}.
% We set the dimension size of the latent vectors from $\{6, 10\}$ for 10 on dSprites and 3D Cars datasets and 6 for 3D Shapes, but we set 10 for CLG-VAE because it sets 10 dimensions size on 3D Shapes in~\cite{commutative-vae}.
% Regarding the CLG-VAE, we fix $\lambda_{\text{decomp}}$, $\lambda_{\text{hessian}}$, and forward group features as 40, 20, and 0.2, respectively.
% Because the hyper-parameters showed the best result in~\cite{commutative-vae}.
% We set group reconstruction from $ \{0.2, 0.5, 0.7\}$.
% For Control-VAE, we set the maximum KL divergence value from $\{10, 12, \ldots, 20 \}$.
In addition, we set masking ratio $\lambda$ from $\{0.0, 0.5, \ldots 2.0, \infty \}$.
% To check the impact of MIPE-transformation, we do not consider the Groupified VAE because the latter is implemented with an extended decoder (different capacity).
We then evaluate familiar benchmarks, FVM score~\citep{factor-vae}, MIG~\citep{beta-tcvae}, SAP~\citep{sap}, and DCI~\citep{dci}.
Further details are in the Appendix~\ref{appendix: hyper-tuning}.

% \subsection{Evaluation} 
% We conduct experiments on  NVIDIA A100, RTX 2080 Ti, and RTX 3090. We set 100 samples to evaluate global empirical variance in each dimension and run it a total of 800 times to estimate the FVM score introduced in~\cite{factor-vae}.
% For the MIG~\cite{beta-tcvae}, SAP~\cite{sap}, and DCI~\cite{dci}, we follow default values introduced in~\cite{Michlo2021Disent}, training and evaluation 100 and 50 times with 100 mini-batches, respectively.
% We evaluate four disentanglement metrics for a less biased understanding of the actual states of disentanglement.

\begin{table*}[ht]
    \caption{Ablation study for the equivariant property (w/o E) and EF-conversion (w/o EF).
    Each metric is averaged over 40 and 20 settings of $\beta$-VAE and $\beta$-TCVAE, respectively.}
    \vspace{-5pt}
    \begin{center}
    {
    \resizebox{0.99\textwidth}{!}{%
    \begin{tabular}{@{~}c@{~}@{~~}c@{~~}c@{~~}c@{~~}c@{~~}c@{~~}c@{~~}c@{~~}c@{~~}c@{~~}c@{~~}c@{~~}c@{~}}
        \hline
        \multirow{2.5}{*}{3D Shapes}
        & \multicolumn{3}{c}{$\beta$-VAE}
        & \multicolumn{3}{c}{$\beta$-TCVAE}
        & \multicolumn{3}{c}{Factor-VAE}
        & \multicolumn{3}{c}{CLG-VAE} \\
        \cmidrule(lr){2-4} \cmidrule(lr){5-7} \cmidrule(lr){8-10} \cmidrule(lr){11-13}
        & MIPET & w/o E & w/o EF
        & MIPET & w/o E & w/o EF
        & MIPET & w/o E & w/o EF
        & MIPET & w/o E & w/o EF \\
        \cmidrule(lr){1-1} \cmidrule(lr){2-4} \cmidrule(lr){5-7} \cmidrule(lr){8-10} \cmidrule(lr){11-13}
        FVM &
        \textcolor{red}{\textbf{75.19}}($\pm$8.16) & 74.91($\pm$10.46) & 22.27($\pm$1.29) &
        \textcolor{red}{\textbf{80.59}}($\pm$8.57) & 77.90($\pm$8.66) & 66.38($\pm$7.57) &
        \textcolor{red}{\textbf{78.34}}($\pm$8.35) & 71.50($\pm$6.78) & 59.28($\pm$5.15) &
        \textcolor{red}{\textbf{80.17}}($\pm$8.43) & 69.75($\pm$6.91) & 61.55($\pm$9.05)\\
        MIG &
        47.37($\pm$10.13) & \textcolor{red}{\textbf{47.45}}($\pm$8.98) & 0.28($\pm$0.09) &
        \textcolor{red}{\textbf{54.49}}($\pm$9.44) & 51.37($\pm$11.54) & 36.08($\pm$17.42) &
        \textcolor{red}{\textbf{48.59}}($\pm$8.35) & 42.25($\pm$11.28) & 32.67($\pm$10.49) & 
        \textcolor{red}{\textbf{53.87}}($\pm$7.41) & 45.92($\pm$4.77) & 33.89($\pm$3.46) \\
        SAP &
        9.20($\pm$2.44) & \textcolor{red}{\textbf{9.43}}($\pm$2.59) & 0.26($\pm$0.07) &
        \textcolor{red}{\textbf{11.58}}($\pm$3.32) & 10.23($\pm$3.13) & 7.13($\pm$3.09) &
        \textcolor{red}{\textbf{9.84}}($\pm$2.73) & 8.19($\pm$0.69) & 7.04($\pm$0.87) & 
        \textcolor{red}{\textbf{12.83}}($\pm$3.01) & 50.14($\pm$5.43) & 46.66($\pm$4.81) \\
        DCI &
        \textcolor{red}{\textbf{54.95}}($\pm$8.99) & 54.23($\pm$9.05) & 0.10($\pm$0.02) &
        \textcolor{red}{\textbf{66.22}}($\pm$7.32) & 61.18($\pm$8.87) & 56.85($\pm$11.72) &
        \textcolor{red}{\textbf{60.23}}($\pm$9.90) & 50.65($\pm$6.85) & 48.83($\pm$5.37) & 
        \textcolor{red}{\textbf{60.74}}($\pm$7.77) & 50.14($\pm$5.43) & 46.66($\pm$4.81) \\
        \hline
    \end{tabular}
    }}
    \end{center}
    \vspace{-10pt}
    \label{tab:main ablations}
\end{table*}

\begin{figure}[ht]
    \centering
    % \begin{subfigure}{0.49\textwidth}
    % \includegraphics[width=\textwidth]{figure_shapes3d_dci_matrix.jpg}
    % \caption{3D Shapes dataset DCI matrix.}
    % \label{figure: dci_matrix_3dshapes}
    % \end{subfigure}
    % \hfill
    % \begin{subfigure}{0.49\textwidth}
    % \includegraphics[width=\textwidth]{figure_car_dcimatrix}
    % \caption{3D Cars dataset DCI matrix.}
    % \label{figure: dci_matrix_3dcars}
    % \end{subfigure}
    \includegraphics[width=0.48\textwidth]{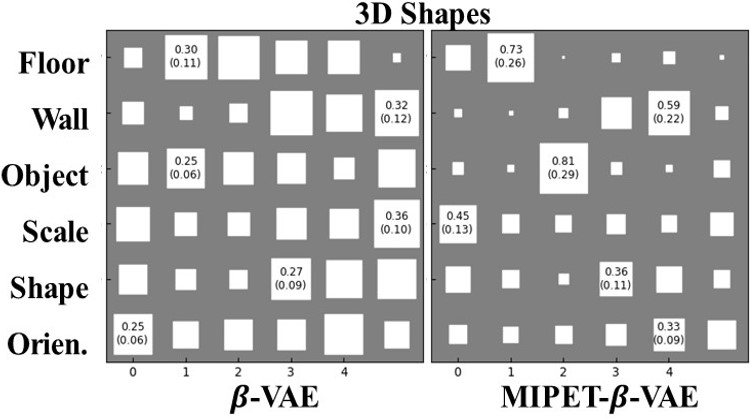}
    \caption{
    Each square represents a value in the DCI matrix, which describes the relationship between the $i^{th}$ latent dimension and each factor.
    The size of each square is relative to the values within each row.
    The ideal case resembles a sparse matrix.
    The y-axis represents the factors of each dataset, while the x-axis corresponds to the latent vector dimensions.
    The number shown in each row of the matrix indicates the maximum value and standard deviation of that row.
    Higher maximum and standard deviation values suggest greater sparsity, indicating closer alignment with the ideal case.
    % More details in Appendix~\ref{appendix: dci matrix}.
    }
    % \label{figure: dci_matrix}
    \label{figure: dci_matrix_3dshapes}
\vspace{-15pt}
\end{figure}

% \begin{table*}[t]
%     \caption{
%     Ablation study for the equivariant property (w/o E), and EF-conversion (w/o EF).
%     Each metric is averaged over 40 and 20 settings of $\beta$-VAE and $\beta$-TCVAE, respectively.} 
%     \centering
%     \scriptsize
%     \begin{tabular}{>{\centering\arraybackslash}m{0.07\textwidth}|>{\centering\arraybackslash}m{0.09\textwidth}*{2}{|>{\centering\arraybackslash}m{0.09\textwidth}}||>{\centering\arraybackslash}m{0.09\textwidth}*{2}{|>{\centering\arraybackslash}m{0.09\textwidth}}}
%         \hline
%         \multirow{3}{*}{3D Shapes} & \multicolumn{3}{c||}{$\beta$-VAE} & \multicolumn{3}{c}{$\beta$-TCVAE}\\
%         \cline{2-7}
%         &  MIPET & MIPET (w/o E) & MIPET (w/o EF) & MIPET & MIPET (w/o E) & MIPET (w/o EF)  \\
%         \hline
%         FVM & 
%         \textbf{75.19}($\pm$8.16) & 74.91($\pm$10.46) & 22.27($\pm$1.29) & \textbf{80.59}($\pm$8.57) & 
%         77.90($\pm$8.66) & 66.38($\pm$7.57) \\
%         MIG & 47.37($\pm$10.13) & \textbf{47.45}($\pm$8.98) & 0.28($\pm$0.09)  & \textbf{54.49}($\pm$9.44) & 51.37($\pm$11.54) & 36.08($\pm$17.42)  \\
%         SAP & 9.20($\pm$2.44) & \textbf{9.43}($\pm$2.59) & 0.26($\pm$0.07) & \textbf{11.58}($\pm$3.32) & 10.23($\pm$3.13) & 7.13($\pm$3.09) \\
%         DCI & \textbf{54.95}($\pm$8.99) & 54.23($\pm$9.05) & 0.10($\pm$0.02) & \textbf{66.22}($\pm$7.32) & 61.18($\pm$8.87) & 56.85($\pm$11.72) \\
%         \hline
%     \end{tabular}
%     \label{tab:main ablations}
% \end{table*}

\begin{figure}[ht]
    \centering
    \begin{subfigure}{0.49\textwidth}
    \includegraphics[width=\textwidth]{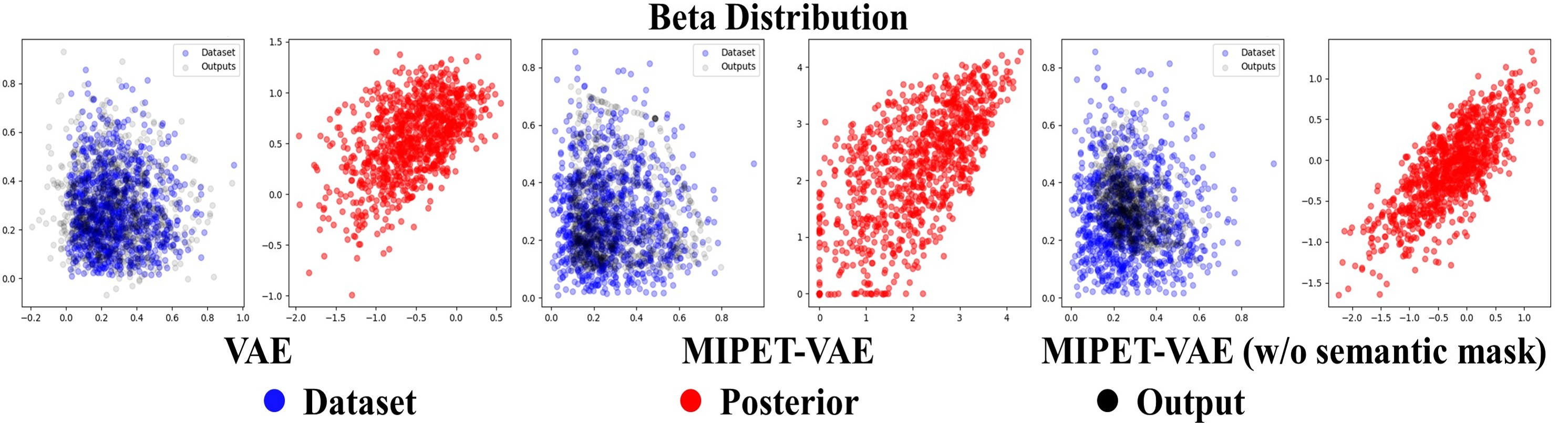}
    \caption{Posterior visualization with Beta distribution}
    \label{figure: exponential family prior}
    \end{subfigure}
    % \hfill
    % \begin{subfigure}{0.49\textwidth}
    % \includegraphics[width=\textwidth]{figure_2d_gamma.jpg}
    % \caption{Posterior visualization with Gamma distribution}
    % \label{figure: exponential family prior_gamma}
    % \end{subfigure}
    % \vfill
    % \begin{subfigure}{0.49\textwidth}
    % \includegraphics[width=\textwidth]{figure_2d_dirichlet.jpg}
    % \caption{Posterior visualization with Dirichlet distribution}
    % \label{figure: exponential family prior_dirichlet}
    % \end{subfigure}
    \vfill
    \begin{subfigure}{0.49\textwidth}
    \includegraphics[width=\textwidth]{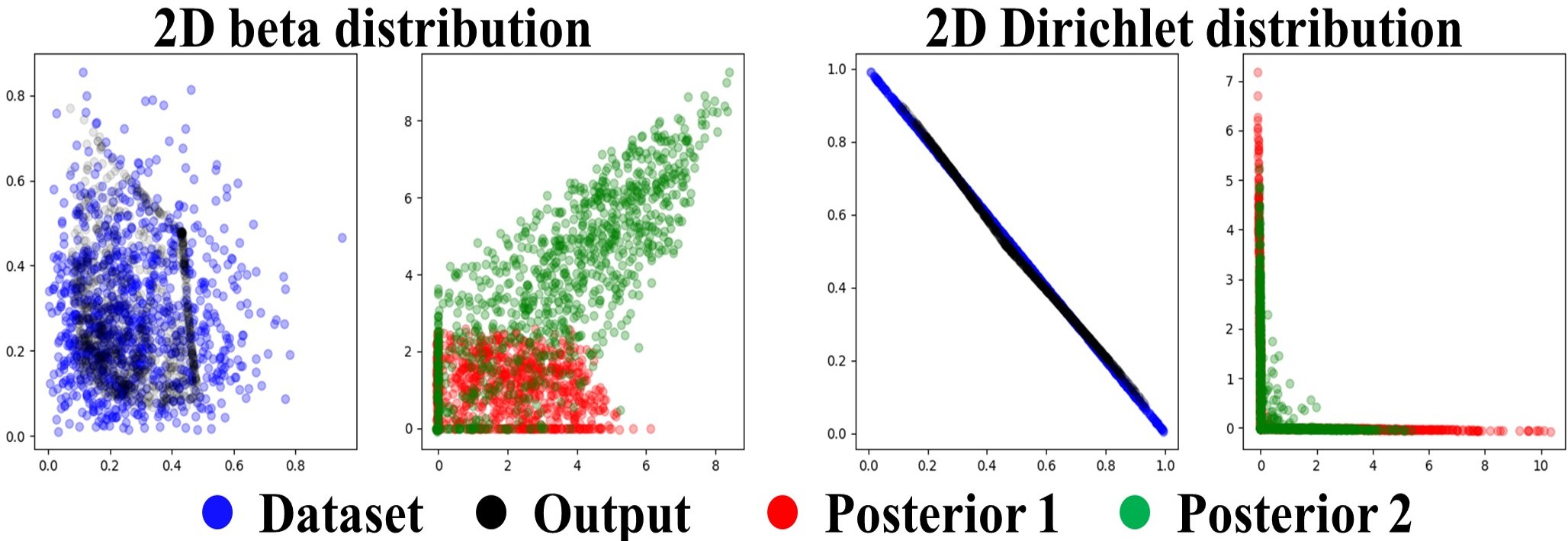}
    \caption{Posterior visualization of MIPET layer}
    \label{figure: multiple-ipe transformation}
    \end{subfigure}
    \caption{
    Non-Gaussian posterior learned by an IPE module without intended guidance to a specific distribution in a toy setting.
    We compare VAE, MIPET-VAE, and MIPET-VAE without a semantic mask to assess how well each model captures the underlying distribution.
    % A toy test compares VAE, MIPET-VAE, and MIPET-VAE without a semantic mask to investigate whether the model approximates the beta distribution.
    We construct the VAE with a 4-layer Multi-Layer Perceptron (MLP) as the encoder and a single linear layer as the decoder.
    Blue plots are randomly sampled from a two-dimensional beta distribution, red plots are the posterior, and black plots are the output results.}
    \label{figure: xxx}
\vspace{-10pt}
\end{figure}

% \begin{figure}[ht]
%     \centering
%     \includegraphics[width=0.48\textwidth]{figure_shapes3d_dci_matrix.jpg}
%     \caption{
%     Each square represents a value in the DCI matrix, which describes the relationship between the $i^{th}$ latent dimension and each factor.
%     The size of each square is relative to the values within each row.
%     The ideal case resembles a sparse matrix.
%     The y-axis represents the factors of each dataset, while the x-axis corresponds to the latent vector dimensions.
%     The number shown in each row of the matrix indicates the maximum value and standard deviation of that row.
%     Higher maximum and standard deviation values suggest greater sparsity, indicating closer alignment with the ideal case.
%     More details in Appendix~\ref{appendix: dci matrix}.}
%     \label{figure: dci_matrix}
% \vspace{-20pt}
% \end{figure}

\section{Results and Discussion}
\label{sec:results}
\subsection{Quantitative Analysis}

\paragraph{Disentanglement Metrics}
We set the number of IPE-transformation functions to be equal to balancing hyper-parameter $\beta$ on $\beta$-VAE because of Eq.~\ref{eq:objective_function}.
The number of IPE-transform functions of $\beta$-TCVAE is 3.
However, in the case of CLG-VAE, we set it to 1 because its approach is based on the group theory, not directly controlling a KL divergence term such as $\beta$-VAE.
We average each model performance value with 40, 20, 60, 10, and 30 cases in VAEs, $\beta$-TCVAEs, Control-VAE, Factor-VAE and 
CLG-VAEs, respectively.

As shown in Table~\ref{tab:quantitative analysis}, MIPET-VAEs disentanglement performance is broadly improved with four metrics on each dataset.
In particular, most FVM results significantly affect the model performance and stability on all datasets.
Therefore, our proposed method obtains a specific dimension that corresponds to a specific single factor.
These results imply that applied to MIPE-transformation functions on VAEs elaborate disentangled representation learning.

\paragraph{Sensitivity to the Number of IPE-transformation and EF-conversion}
\label{sensitivity to the number of ie}
We analyze the impact of the MIPE-transformation function.
As presented in Table~\ref{tab:main impact of IE}, MIPE-transformation outperforms IPE-transformation in disentanglement learning across all datasets.
Indeed, the results of MIPET-$\beta$-VAEs more generally and clearly demonstrate the impact of the MIPE-transfomation function.
Our derivation in Section~\ref{appendix: multiple unit} clearly explains the impact of MIPE-transformation.
This result shows the impact of the multiple uses of IPE-transformation and EF-conversion.
\vspace{-10pt}

\paragraph{Impact of Implicit Semantic Mask}
We set masking hyper-parameter $\lambda$ from $\{0.0, 0.5, \cdots, 2.0, \infty\}$, and each model has different $\lambda$ for best case.
In Table~\ref{tab:mask}, VAE and CLG-VAE with masked log-normalizer show better and well-balanced results than the models without masking, which implies improvement of disentanglement.

% \begin{figure*}[t]
%     \centering
%     \includegraphics[width=0.8\textwidth]{figure_2d_mipet.jpg}
%     \caption{
%     Demonstration of distribution separation using IPE-transformation on two-dimensional exponential family distributions (Beta and Dirichlet).
%     % A toy test with two-dimensional exponential family distribution (beta and Dirichlet distribution).
%     We set the number of IPE-transformation as 2.
%     Green and red plots refer to latent vectors (posterior) from each IPE-transformation.}
%     \label{figure: multiple-ipe transformation}
% \end{figure*}

% \begin{figure*}[t]
%     \centering
%     \subfloat[dSprites]{\includegraphics[width=0.33\textwidth]{figure_dsprites_quali_ver13.jpg}\label{fig:dsprites}}%
%     \hfill
%     \subfloat[3D Shapes]{\includegraphics[width=0.33\textwidth]{figure_dsprites_quali_ver13.jpg}\label{fig:3dshape}}%
%     \hfill
%     \subfloat[3D Car]{\includegraphics[width=0.33\textwidth]{figure_dsprites_quali_ver13.jpg}\label{fig:3dcars}}%
%     % \hfill
%     % \subfloat[CelebA]{\includegraphics[width=0.33\textwidth]{figure_dsprites_quali_ver13.jpg}\label{fig:celeba}}%
%     \caption{Qualitative results on various datasets. The left-side grids are input images and their variants by changing the activations of each dimension of the latent vectors. 
%     The first column shows input images.
%     A number of each dataset refers to the latent factors of each dataset.
%     % The right-side table shows matching pre-defined factors of the dataset (red: MIPET, blue: no MIPET).
%     }
%     \label{fig:qualitative_results}
% \end{figure*}

\begin{figure*}[t]
    \centering
    % \subfloat[dSprites]{\includegraphics[width=0.24\textwidth]{figure_dsprites_quali_ver13.jpg}\label{fig:dsprites}}%
    % \hfill
    % \subfloat[3D Shapes]{\includegraphics[width=0.24\textwidth]{figure_3dshape_quali_commut_ver04.jpg}\label{fig:3dshape}}%
    % \vfill
    \subfloat[dSprites]{\includegraphics[width=0.48\textwidth]{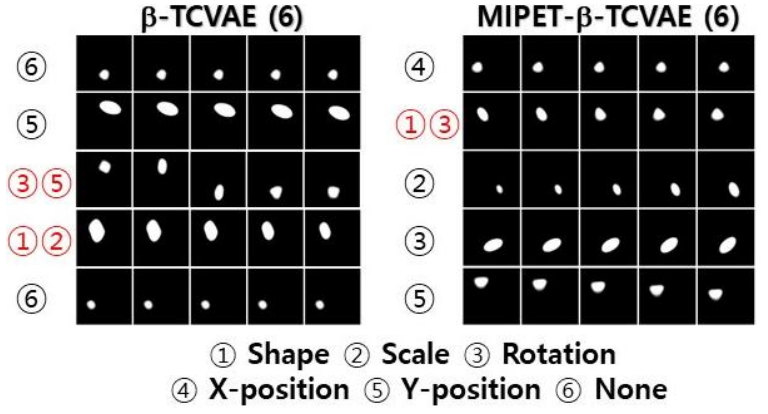}\label{fig:dsprites}}%
    \hfill
    \subfloat[3D Shapes]{\includegraphics[width=0.48\textwidth]{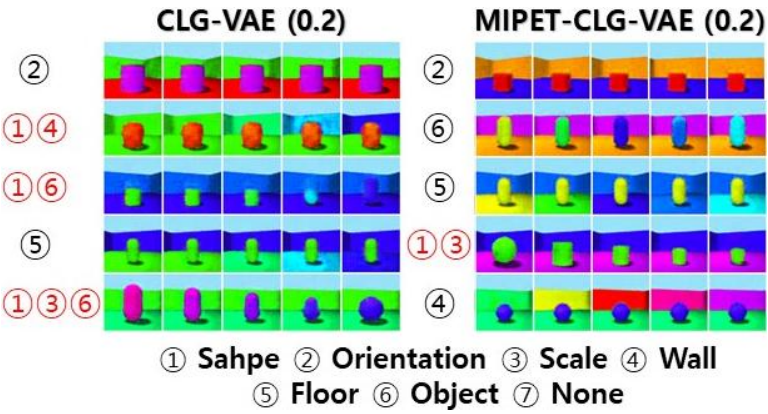}\label{fig:3dshapes}}%
    \vfill
    \subfloat[3D Cars]{\includegraphics[width=0.48\textwidth]{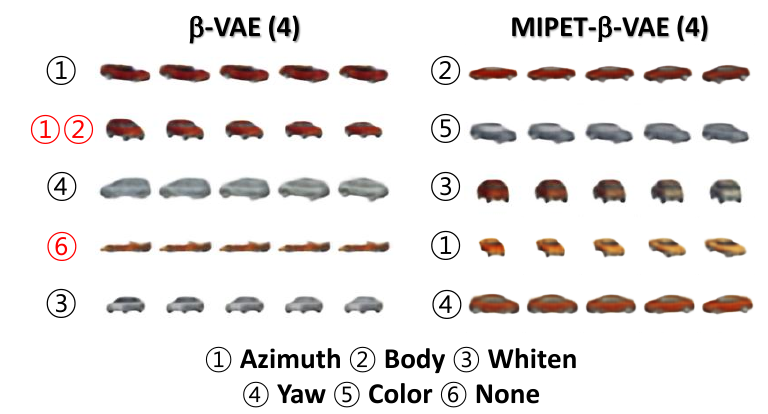}\label{fig:3dcars}}%
    \hfill
    \subfloat[CelebA]{\includegraphics[width=0.48\textwidth]{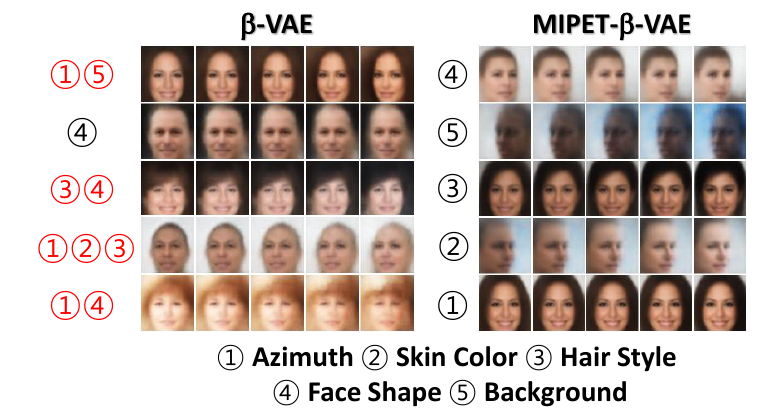}\label{fig:celeba}}%
    \caption{Qualitative results on various datasets, which show the factors learned for each dimension of $z$. 
    % Further details are in the Appendix~\ref{appendix: details of traversal}.
    % The left-side grids are input images and their variants by changing the activations of each dimension of the latent vectors. 
    % The first column shows input images.
    % A number of each dataset refers to the latent factors of each dataset.
    % The right-side table shows matching pre-defined factors of the dataset (red: MIPET, blue: no MIPET).
    }
    \label{fig:qualitative_results}
    \vspace{-10pt}
\end{figure*}

\paragraph{Ablation Study}
\label{sec:ablation}
We conduct an ablation study to evaluate the separate impact of equivariant property and the EF-conversion.
We have already presented the impact of the multiple uses of IPE-transform and EF-conversion in the previous paragraph.
We evaluate the impact of the other properties by setting MIPE-transformation 1) without equivariant (w/o E), which is implemented as an asymmetric matrix, and 2) without EF-conversion (w/o EF).
To exclude group theory interference with other methods, we select $\beta$-VAE and $\beta$-TCVAE.
As the results are shown in Table~\ref{tab:main ablations}, most of the results show that MIPET-VAEs performance is better than other cases.
In particular, MIPET (w/o EF) results are lower than MIPET (w/o E) results and are clearly shown in all cases.

% \vspace{-10pt}
\subsection{Qualitative Analysis}

\paragraph{Relation Between Factors and Latent Vector Dimensions}
\label{appendix: dci matrix}

% \begin{figure}[ht]
%     \centering
%     \includegraphics[width=0.48\textwidth]{figure_car_dcimatrix.jpg}
%     \caption{3D Car Dataset: Azim. refers to the azimuth factor, and Elev. refers to the elevation factor.}
%     \label{fig:enter-label}
% \end{figure}

% \paragraph{Relation Between Factors and Latent Vector Dimensions.}
As shown in Fig.~\ref{figure: dci_matrix_3dshapes}, 
% the MIPET model shows the larger maximum value and standard deviation of each row except for the scale factor in the dSprites dataset.
our model shows a close to sparse matrix compared to the $\beta$-VAE result in 3D Shapes, also the maximum value and standard deviation of each row are larger than $\beta$-VAE.
This indicates that our model achieves better alignment of individual factors with specific latent dimensions compared to other methods.
More details are in Appendix~\ref{appendix: dci matrix}.
% In the 3D Car dataset, a trade-off was observed. While the maximum value across models and evaluation factors showed a slight decrease, the azimuth factor exhibited a notable improvement compared to the declines seen in other factors.

% \paragraph{Does IPE-transformation map the Gaussian Distribution to Diverse Exponential Family?}
\paragraph{The Impact of IPE-transformation}
The IPE-transformation $\psi(\cdot)$ is designed to achieve two objectives: 1) to represent a non-Gaussian exponential family distribution as the posterior, and 2) to preserve the inherent structure of the dataset. 
To assess the first objective, we construct a toy dataset sampled from a beta distribution (additional cases are provided in Appendix~\ref{appendix: ef}) and configure the decoder as a single linear layer, noting that the linear transformation of a Beta distribution remains within the same family. 
As shown in Fig.~\ref{figure: exponential family prior}, the posterior learned by MIPET-VAE (red dots) closely approximates the beta distribution, whereas the posteriors of VAE and MIPET (without the semantic mask) retain Gaussian distribution. 
Regarding the second objective, MIPET-VAE exhibits improved preservation of the dataset structure relative to VAE (black dots).
Additional experimental details are reported in Appendix~\ref{appendix: ef}.

\paragraph{Does Each IPE-transformation Independently Operate?}
We demonstrate how each IPE-transformation represents the exponential family.
As shown in Fig.~\ref{figure: multiple-ipe transformation}, each IPE-transformation represents a different distribution because the green and red plots are distributed in different posteriors, even though all IPE-transformations share the same decoder weights.
% \vspace{-5pt}

\paragraph{Disentanglement of Factors Across Latent Dimensions}
We randomly sample an image for each dimension of the latent vector space and creates 5 variants of its generated latent vector by selecting values from $\{$-2, 2$\}$ with 5 intervals for the dimension, then generate their corresponding output images.
For the generation, we select $\beta$-TCVAE (6), which shows the best FVM scores in dSprites dataset.
Thereafter, we evaluate the semantic roles of each dimension before and after applying MIPE-transformation function.

In Fig.~\ref{fig:dsprites}, $\beta$-TCVAE struggles with rotation and y-position, as shown on the 3$^{rd}$ row, and with scale and shape represented on the 4$^{th}$ row.
On the contrary, MIPET-$\beta$-TCVAE separates rotation and y-position factor (4$^{th}$, and 5$^{th}$ rows).
Applied our method on $\beta$-TCVAE shows better disentangled representation on dSprites dataset.
These results also show that our proposed method improves disentangled representation learning.
As shown in the Fig.~\ref{fig:3dshapes}, CLG-VAE struggles with shape and wall color factors in 2$^{nd}$ dimension, and shape and object color factors in 3$^{rd}$ dimension.
In particular, it struggles with three factors in 5$^{th}$ dimension.
On the other hand, MIPET-CLG-VAE separates shape, wall, and object color factors.
The qualitative analysis with 3D Cars dataset, as shown in Fig.~\ref{fig:3dcars}, the left side is the $\beta$-TCVAE result, and it struggles with azimuth and body factors shown in the 2$^{nd}$ row.
However, MIPET-$\beta$-TCVAE separates azimuth (4$^{th}$ row) and body (1$^{st}$ row).
In particular, MIPET-$\beta$-TCVAE learns \textit{color} factor (2$^{rd}$ row) which does not exist on $\beta$-TCVAE.
Furthermore, our model shows improvement on the CelebA~\cite{celebA(64)} dataset, as shown in Fig.~\ref{fig:celeba}.
$\beta$-VAE struggles with most factors of the dataset, but ours separates all factors.
% \vspace{-5pt}

\section{Conclusion}
\label{sec: conclusion}
In this paper, we address the problem of injecting inductive bias for learning unsupervised disentangled representations.
To build the bias in VAE frameworks, we propose MIPE-transformation composed of 1) IPE-transformation for the benefits of invertibility and partial-equivariance for disentanglement, 2) a training loss and module to adapt unrestricted prior and posterior to an approximated exponential family, and 3) integration of multiple units of IPE-transformation function and EF-conversion for more expressive bias. %(MIE-transformation).
The method is easily equipped on state-of-the-art VAEs for disentanglement learning and shows significant improvement on dSprites, 3D Shapes, and 3D Cars datasets. 
We expect that our method can be applied to more VAEs, and extended to downstream applications.

\section*{Acknowledgments}
This work was supported by the National Research Foundation of Korea (NRF) grant funded by the Korea government (MSIT) (No.2022R1A2C2012054, Development of AI for Canonicalized Expression of Trained Hypotheses by Resolving Ambiguity in Various Relation Levels of Representation Learning).

\bibliography{main}

% \paragraph*{References}

% References follow the acknowledgements.  Use an unnumbered third level
% heading for the references section.  Please use the same font
% size for references as for the body of the paper---remember that
% references do not count against your page length total.

% \begin{thebibliography}{}
% \setlength{\itemindent}{-\leftmargin}
% \makeatletter\renewcommand{\@biblabel}[1]{}\makeatother
% \bibitem{} J.~Alspector, B.~Gupta, and R.~B.~Allen (1989).
%     \newblock Performance of a stochastic learning microchip.
%     \newblock In D. S. Touretzky (ed.),
%     \textit{Advances in Neural Information Processing Systems 1}, 748--760.
%     San Mateo, Calif.: Morgan Kaufmann.

% \bibitem{} F.~Rosenblatt (1962).
%     \newblock \textit{Principles of Neurodynamics.}
%     \newblock Washington, D.C.: Spartan Books.

% \bibitem{} G.~Tesauro (1989).
%     \newblock Neurogammon wins computer Olympiad.
%     \newblock \textit{Neural Computation} \textbf{1}(3):321--323.
% \end{thebibliography}

%%%%%%%%%%%%%%%%%%%%%%%%%%%%%%%%%%%%%%%%%%%%%%%%%%%%%%%%%%%%
\section*{Checklist}

% %%% BEGIN INSTRUCTIONS %%%
The checklist follows the references. For each question, choose your answer from the three possible options: Yes, No, Not Applicable.  You are encouraged to include a justification to your answer, either by referencing the appropriate section of your paper or providing a brief inline description (1-2 sentences). 
Please do not modify the questions.  Note that the Checklist section does not count towards the page limit. Not including the checklist in the first submission won't result in desk rejection, although in such case we will ask you to upload it during the author response period and include it in camera ready (if accepted).

\textbf{In your paper, please delete this instructions block and only keep the Checklist section heading above along with the questions/answers below.}
% %%% END INSTRUCTIONS %%%

\begin{enumerate}

  \item For all models and algorithms presented, check if you include:
  \begin{enumerate}
    \item A clear description of the mathematical setting, assumptions, algorithm, and/or model. [Yes - We present in Section~\ref{sec:method}, and Appendix~\ref{appendix: proof details}.]
    \item An analysis of the properties and complexity (time, space, sample size) of any algorithm. [Yes -  We present in the Appendix~\ref{appendix: complexity}.]
    \item (Optional) Anonymized source code, with specification of all dependencies, including external libraries. [Yes - We submit our code as supplement materials.]
  \end{enumerate}

  \item For any theoretical claim, check if you include:
  \begin{enumerate}
    \item Statements of the full set of assumptions of all theoretical results. [Yes - We present in Section~\ref{sec:method}, and Appendix~\ref{appendix: proof details}.]
    \item Complete proofs of all theoretical results. [Yes - We present in Section~\ref{sec:method}, and Appendix~\ref{appendix: proof details}.]
    \item Clear explanations of any assumptions. [Yes - We present in Section~\ref{sec:method}, and Appendix~\ref{appendix: proof details}.]     
  \end{enumerate}

  \item For all figures and tables that present empirical results, check if you include:
  \begin{enumerate}
    \item The code, data, and instructions needed to reproduce the main experimental results (either in the supplemental material or as a URL). [Yes - We submit our code as supplement materials.]
    \item All the training details (e.g., data splits, hyperparameters, how they were chosen). [Yes - We present in Section~\ref{sec:experiment} and Appendix~\ref{appendix: details of experimentla environment}.]
    \item A clear definition of the specific measure or statistics and error bars (e.g., with respect to the random seed after running experiments multiple times). [Yes - We present in Section~\ref{sec:results}.]
    \item A description of the computing infrastructure used. (e.g., type of GPUs, internal cluster, or cloud provider). [Yes - We present in Appendix~\ref{appendix: details of experimentla environment}.]
  \end{enumerate}

  \item If you are using existing assets (e.g., code, data, models) or curating/releasing new assets, check if you include:
  \begin{enumerate}
    \item Citations of the creator If your work uses existing assets. [Yes - We submit our code as supplement materials.]
    \item The license information of the assets, if applicable. [Yes - We submit our code as supplement materials.]
    \item New assets either in the supplemental material or as a URL, if applicable. [Yes - We submit our code as supplement materials.]
    \item Information about consent from data providers/curators. [Yes - We present in Appendix~\ref{appendix: details of experimentla environment}.]
    \item Discussion of sensible content if applicable, e.g., personally identifiable information or offensive content. [Not Applicable]
  \end{enumerate}

  \item If you used crowdsourcing or conducted research with human subjects, check if you include:
  \begin{enumerate}
    \item The full text of instructions given to participants and screenshots. [Not Applicable]
    \item Descriptions of potential participant risks, with links to Institutional Review Board (IRB) approvals if applicable. [Not Applicable]
    \item The estimated hourly wage paid to participants and the total amount spent on participant compensation. [Not Applicable]
  \end{enumerate}

\end{enumerate}

\clearpage
\appendix
\onecolumn
\aistatstitle{MIPET Supplementary Materials}

\section{Notation and Groups}
\label{appendix: notation and groups}
\begin{table}[h]
\caption{Terms and Notations}
    \centering
    \begin{tabular}{ll}
    %\multicolumn{2}{c}{Terms and Notations}\\
    \hline
$\psi(\cdot)$ & Invertible function \\
$ \vz$ & Latent vector from encoder \\
$\hat{ \vz}_m$ & Transformed latent vector by $\psi_m(\cdot)$\\
$\boldsymbol{\theta}_{\hat{ \vz}_m}$ & Natural Parameter of posterior \\
$\hat{\boldsymbol{\epsilon}}_m$ & Transformed prior samples by $\psi_m(\cdot)$ \\
$\boldsymbol{\theta}_{\hat{\boldsymbol{\epsilon}}_m}$ & Natural Parameter of prior \\
$A$ & Log-Normalizer \\
$T$ & Sufficient Statistics \\
$\boldsymbol{\nu}$ & Evidence\\
$f_{ \vx}(\cdot)$ & Power Density Function \\
$GL_n(\mathbb{R})$ & General Linear Group \\
$M_n(\mathbb{R})$ & A set of $n \times n$ real matrix \\
$Sym_n(\mathbb{R})$ & A set of $n \times n$ symmetric real matrix \\
$E_M$ & $\{ \mathbf{e}^M | M \in M_n(\mathbb{R}) \}$  \\
$E_S$ &  $\{ \mathbf{e}^S | S \in Sym_n(\mathbb{R}) \}$ \\
$\psi_{M}(\cdot)$ & $\psi_{M}(\cdot) \in M_n(\mathbb{R})$\\ 
$\psi_{E_S}(\cdot)$ & $\psi_{E_S}(\cdot) \in E_S$ \\
$\psi_{E_M}(\cdot)$ & $\psi_{E_M}(\cdot) \in E_M$ \\ 
$G_S$ & $G_S:(\mathbf{e}^S, \ast)$ \\
$G_I$ & Group of input space for symmetries \\
$G_L$ & Group of latent space for symmetries \\
$G_T$ & Group of transformed latent space for symmetries \\
$J$ & $G_S \cap G_L$ \\
% $G_S^J$ & $G_S:(\mathbf{e}^S, \ast)$ \\
$G_I^J$ & Group of input space for symmetries corresponds to $J$ \\
$G_L^J$ & Group of latent space for symmetries corresponds to $J$ \\
$G_L^J$ & Group of transformed latent space for symmetries corresponds to $J$\\
% $\Gamma$ & $G_L \times G_T \rightarrow G_T$ \\
% $\Gamma^J$ & $G_L^J \times G_T^J \rightarrow G_T^J$ \\
% $\Xi$ & $G_I \times G_L \rightarrow G_L$  \\
% $\Xi^J$ & $G_I^J \times G_L^J \rightarrow G_L^J$  \\
$\mathcal{X}$ & Input space \\ 
$\mathcal{Z}$ & Latent vector space  \\
$\hat{\mathcal{Z}}$ & Transformed latent vector space \\
$\mathcal{Z}^\prime$ & Homogeneous space of group $G_L$ (homogeneous $G_L$-space), $\mathcal{Z}^\prime \subseteq \mathcal{Z}$ \\
$\hat{\mathcal{Z}}^\prime$ & Homogeneous space of group $G_T$ (homogeneous $G_T$-space), $\hat{\mathcal{Z}}^\prime \subseteq \hat{\mathcal{Z}}$  \\
$\hat{\mathcal{Z}}^J$ & Homogeneous space of group $G_T^J$ (homogeneous $G_T^J$-space), $\hat{\mathcal{Z}}^J \subseteq \hat{\mathcal{Z}}$  \\
$ \KL(\cdot||\cdot)$ & Kullback-Leibler divergence \\
$\mathbf{0}$ & zero vector \\
$\mathbf{0}_{n,n}$ & n by n zero matrix\\ 
%$\mathcal{X}$ $\mathcal{Z}$ $\mathcal{\hat{Z}}$ & $\mathbf{x}_1$ $\mathbf{x}_2$ $\mathbf{z}_1$ $\mathbf{z}_2$ $\mathbf{\hat{z}}_1$ $\mathbf{\hat{z}}_2$
    \hline
    \label{term_notation}
    \end{tabular}
% \vspace{-20pt}
\end{table}

\paragraph{Notation.}
We denote $g_I \in G_I$, $g_L \in G_L$, and $g_T \in G_T$.
Let $M_n(\mathbb{R})$ be the set of $n \times n$ real matrices and $\mathrm{Sym}_n(\mathbb{R})$ the set of real symmetric matrices.
For any set $\mathcal{A} \subseteq \mathrm{Sym}_n(\mathbb{R})$, define $\exp(\mathcal{A}) := \{\exp(S) \mid S \in \mathcal{A}\}$.
We write $E_M := \{\exp(M) \mid M \in M_n(\mathbb{R})\}$ and $E_S := \exp(\mathrm{Sym}_n(\mathbb{R}))$.

\medskip
\noindent\textbf{Standing assumption on $G_S$.}
We assume a \emph{pairwise-commuting} symmetric set $\mathcal{S} \subseteq \mathrm{Sym}_n(\mathbb{R})$, i.e., $[S_1,S_2]=0$ for all $S_1,S_2 \in \mathcal{S}$.
Then we define
\[
  G_S := \exp(\mathcal{S}) = \{\exp(S) \mid S \in \mathcal{S}\} \subset \mathrm{GL}_n(\mathbb{R}),
\]
with the group operation given by the usual matrix multiplication ``$\ast$''.

\section{Proof}
\label{appendix: proof details}

\begin{lemma}[Abelian Lie subgroup]\label{lem:GS_abelian}
If $\mathcal{S} \subseteq \mathrm{Sym}_n(\mathbb{R})$ is pairwise-commuting, then $(G_S,\ast)$ is an abelian Lie subgroup of $\mathrm{GL}_n(\mathbb{R})$.
In particular, $G_S$ is closed under multiplication and inversion, has identity $I=\exp(0)$, and is commutative.
\end{lemma}

\begin{proof}[Proof sketch]
Since all $S \in \mathcal{S}$ commute and are symmetric, they are simultaneously diagonalizable by some orthogonal matrix $Q$:
$S_i = Q \Lambda_i Q^\top$ with diagonal $\Lambda_i$.
Hence $\exp(S_i) = Q \exp(\Lambda_i) Q^\top$ are symmetric positive definite and commute.
Moreover, for any $S_1,S_2 \in \mathcal{S}$,
the Baker–Campbell–Hausdorff formula reduces to $\exp(S_1)\exp(S_2) = \exp(S_1+S_2)$ because $[S_1,S_2]=0$.
Since $S_1+S_2 \in \mathcal{S}$, we have closure.
The inverse is $\exp(S)^{-1}=\exp(-S)$ and $-S \in \mathcal{S}$.
Commutativity follows from $[S_1,S_2]=0$.
Smoothness of $\exp$ implies $G_S$ is a (abelian) Lie subgroup.
\end{proof}

\begin{proposition} 
\label{appendix: prop2}
By the Lemma~\ref{lem:GS_abelian}, if $q_\phi$ is equivariant over defined on group of symmetries $G_I^J$ and $G_L^J$, then $\psi_{G_S} (q_\phi(\cdot))$ is equivariant to symmetries in $G_I$ corresponding to $G_S \cap G_L$ and $G_T$ corresponding to $G_S \cap G_L$ by the equivariance of $q_\phi$.  
\end{proposition}
$\textit{Proof.}$ The function $\psi_{G_S}(\cdot)$ is an equivariant function over group elements in $G_S \cap G_L$ by Proposition~\ref{prop1}. 
Then, the composite function, $\psi_{G_S}(\cdot)$ and $q_\phi$, is an equivariant function 
of $G_I$ corresponding to $G_S \cap G_L$ and $G_T$ corresponding to $G_S \cap G_L$.
Let $g_L^J$ be a group element in $G_S \cap G_L$, and $g_I^J$ is a group element in $G_I$ corresponding to $G_S \cap G_L$, and $g_T^J$ is a group element where corresponding to $G_S \cap G_L$ on the latent vector space transformed from the original latent vector space.
Then, group element $g_T^J$ is equal to $g_L^J$:
\begin{align}
    %\begin{split}
        & \hat{ \vz}_1 = \psi_{G_S}( \vz_1), \text{and} \\
        & \hat{ \vz}_2 = \psi_{G_S}( \vz_2) = \psi_{G_S}(g_L^J  \vz_1) = g_L^J \psi_{G_S}( \vz_1) ~(\because \text{Prop.~\ref{prop1}}), \\
    \begin{split}
        & \text{then} ~g_L^J \psi_{G_S}( \vz_1) = g_T^J \psi_{G_S}( \vz_1) ~(\because \hat{ \vz}_2 = g_T^J \hat{ \vz}_1) 
        \label{equation: appendix equivariance}\\
        & \Rightarrow (g_L^J - g_T^J) \psi_{G_S}( \vz_1) = \mathbf{0},
    \end{split}
\end{align}
where $\mathbf{0}$ is a zero vector.
Eq.~\ref{equation: appendix equivariance} is defined when $\forall  \vz \in \mathcal{Z}$ by the equivariance definition.
In other words, Eq.~\ref{equation: appendix equivariance} is satisfied only if the kernel (linear algebra) of $g_L^J - g_T^J$, notated as \textit{ker}$(g_L^J - g_T^J)$, includes the basis of $\mathbb{R}^n$ vector space.
If the standard basis of $\mathbb{R}^n$ vector space is in \textit{ker}$(g_L^J - g_T^J)$, then $(g_L^J - g_T^J)=\mathbf{0}_{n,n}$, where $\mathbf{0}_{n,n}$ is an n by n zero matrix.
Other bases of $\mathbb{R}^n$ vector space are expressed by the standard basis. Therefore $g_L^J- g_T^J = \mathbf{0}_{n,n}$. 
%If $\psi_{G_S}(\mathbf{z}_1)$ is an eigenvector of $g_L^J-g_T^J$, then $g_L^J-g_T^J \neq \mathbf{0}_{d,d}$.
%However Eq.~\ref{equation: appendix equivariance} is satisfied with $\forall \mathbf{z} \in \mathcal{Z}$ by the equivariance definition.
%Therefore $g_L^J- g_T^J = \mathbf{0}_{d,d}$.

%$\hat{\mathbf{z}}_1 = \psi_{G_S}(\mathbf{z}_1)$ and $\hat{\mathbf{z}}_2 = \psi_{G_S}(\mathbf{z}_2) = \psi_{G_S}(g_L^J \mathbf{z}_1) = g_L^J \psi_{G_S}(\mathbf{z}_1)$ because $\psi_{G_S}$ is equivariant by Lemma~\ref{lem1}.
%$\hat{\mathbf{z}}_2 = g_T^J \hat{\mathbf{z}}_1$ then $g_L^J \psi_{G_S}(\mathbf{z}_1) = g_T^J \psi_{G_S}(\mathbf{z}_1)$.
%$ = \psi_{G_S}(g_L^J \mathbf{z}_1)$.

%because $\psi_{G_S}$ is a $n\timesn$ matrix for linear transformation function whose domain and range are defined on the same vector space.
%and $z$ and $\hat{z}$ are in the same space).

Then, $\psi_{G_S}(g_L^J  \vz_1) = g_L^J \psi_{G_S}( \vz_1) = g_T^J \psi_{G_S}( \vz_1)$.
The encoder is an equivariant function over input space $\mathcal{X}$ as $q_{\phi} (g_I^J  \vx_1) = g_L^J q_{\phi}( \vx_1)$.
%The latent vector $\mathbf{z}$ holds equivariance property as $q_{\phi} (g_I^J \mathbf{x}_1) = g_L^J q_{\phi}(\mathbf{x}_1)$. 
Mixing two equivarience property, we can derive another equivariance relation $g_T^J \psi_{G_S}(q_\phi( \vx_1)) = \psi_{G_S}(q_\phi(g_I^J  \vx_1))$
%$\psi_{G_S}(g_L^J q_{\phi} (g_I^J x)) = g_T^J \psi(g_L^J q_{\phi}(x))$.
This result implies that the equivariance between input space and a latent space is preserved for $G_S \cap G_L$ if the latent vector $ \vz$ is transformed by $\psi_{G_S}.
$%if the transformation from the latent vector space to the same space is equivarient. 
$\blacksquare$

We show that $\psi_{G_S}$ preserves equivariance between $G_L^J$ and $G_I^J$.
%group elements of latent space in $G_L^J$ and input space.
If there exists equivariant function between input and latent vector space, there should be a group $G_L$ for a latent space and its corresponding group $G_I$ in an input space by definition of equivariance ($q_{\phi} (g_I x) = g_L q_{\phi}(x)$). 
%We assumed input space has group $G_I$ for symmetries, and latent space also has group $G_L$ which corresponds to $G_I$.
%In other words, encoder $q_\phi$ is equivariant between input and latent space as Fig.~\ref{figure: equivalence cycle} ($q_{\phi} (g_I x) = g_L q_{\phi}(x)$).
%Then, the composite function of equivariant, $\psi_{G_S}(\cdot)$ and $q_\phi$, to the shared $\circled{a}$ is also equivariant to $G_I$ corresponding to $\circled{a}$ (see Appendix~\ref{appendix: proof equivariance}).$\blacksquare$
In other words, $\psi_{G_S}(\cdot)$ guarantees to preserve the equivariance of I2L-transformation to certain symmetries in $G_S \cap G_L$ after IPE-transformation as shown in Fig.~\ref{sub fig:suboverview}.

% Let $P(B)$ be the probability of $\psi(\cdot) \in B$ for a subset $B \subset M_n(\mathbb{R})$ after VAE training, and $Pr(\psi_B \in B^\prime)$ be the conditional probability of $\psi(\cdot) \in B^\prime$ given $\psi(\cdot) \in B$.

% \begin{proposition}
% \label{prop:prob-compare}
% Under Assumptions~1--4, the following probability inequality holds:
% \[
%     \Pr(\psi(\cdot) \in G_S \mid \psi(\cdot)\in E_S) \;>\;
%     \Pr(\psi(\cdot) \in G_S \mid \psi(\cdot)\in E_M) \;>\;
%     \Pr(\psi(\cdot) \in G_S \mid \psi(\cdot)\in M_n(\mathbb{R})),
% \]
% where probabilities are defined with respect to the normalized uniform measure on a compact subset 
% $K \subset M_n(\mathbb{R})$.
% \end{proposition}

% \begin{proof}
% Since $G_S \subseteq E_S \subseteq E_M \subseteq M_n(\mathbb{R})$, 
% we have
% \[
% \Pr(\psi(\cdot) \in G_S \mid \psi(\cdot)\in E_S) 
% = \frac{\mu(G_S \cap K)}{\mu(E_S \cap K)},
% \]
% \[
% \Pr(\psi(\cdot) \in G_S \mid \psi(\cdot)\in E_M) 
% = \frac{\mu(G_S \cap K)}{\mu(E_M \cap K)},
% \]
% \[
% \Pr(\psi(\cdot) \in G_S \mid \psi(\cdot)\in M_n(\mathbb{R})) 
% = \frac{\mu(G_S \cap K)}{\mu(M_n(\mathbb{R}) \cap K)},
% \]
% where $\mu$ is the normalized uniform measure on $K$.
% Because $G_S \subseteq E_S \subseteq E_M \subseteq M_n(\mathbb{R})$ and all denominators are finite by compactness of $K$, 
% the ratios satisfy
% \[
% \frac{\mu(G_S \cap K)}{\mu(E_S \cap K)} 
% > \frac{\mu(G_S \cap K)}{\mu(E_M \cap K)} 
% > \frac{\mu(G_S \cap K)}{\mu(M_n(\mathbb{R}) \cap K)}.
% \]
% Thus the stated inequality follows. $\blacksquare$
% \end{proof}

Let $P(G):= \frac{\mu(G \cap K)}{\mu(K)}$ by the assumption 4 in section~\ref{sec:method}, and $Pr(\psi_{G^\prime}(\cdot) \in G) = P(\psi(\cdot) \in G | \psi(\cdot) \in G^\prime) = \frac{P(G\cap G^\prime)}{P(G^\prime)}$ after VAE training.
Then,  
\begin{proposition} 
\label{appendix: prop3}
Under Assumption 4, the following probability inequality holds: $Pr(\psi_{E_S}(\cdot) \in G_S) > Pr(\psi_{E_M}(\cdot) \in G_S) > Pr(\psi_{M}(\cdot) \in G_S)$, where probabilities are defined with respect to the normalized uniform measure on a compact subset 
$K \subset M_n(\mathbb{R})$.
\end{proposition}

$\textit{Proof.}$ 
Since $G_S \subset E_S \subset E_M \subset M_n(\mathbb{R})$, 
\[
\Pr(\psi_{E_S}(\cdot) \in G_S) = \frac{P(G_S \cap E_S)}{P(E_S)} = \frac{P(G_S)}{P(E_S)}
= \frac{\mu(G_S \cap K)}{\mu(E_S \cap K)}, \,\, \big( \because G_S \subset E_S \big)
\]
\[
\Pr(\psi_{E_M}(\cdot) \in G_S)  = \frac{P(G_S \cap E_M)}{P(E_M)} = \frac{P(G_S)}{P(E_M)}
= \frac{\mu(G_S \cap K)}{\mu(E_M \cap K)}, \,\, \big( \because G_S \subset E_M \big)
\]
\[
\Pr(\psi_{M}(\cdot) \in G_S) = \frac{P(G_S \cap M_n(\mathbb{{R}}))}{P(M_n(\mathbb{{R}}))}  = \frac{P(G_S)}{P(M_n(\mathbb{R}))}
= \frac{\mu(G_S \cap K)}{\mu(M_n(\mathbb{R}) \cap K)}, \,\, \big( \because G_S \subset M_n(\mathbb{R}) \big)
\]
where $\mu$ is the normalized uniform measure on $K$.
Because $G_S \subseteq E_S \subseteq E_M \subseteq M_n(\mathbb{R})$ and all denominators are finite by compactness of $K$, 
the ratios satisfy
\[
\frac{\mu(G_S \cap K)}{\mu(E_S \cap K)} 
> \frac{\mu(G_S \cap K)}{\mu(E_M \cap K)} 
> \frac{\mu(G_S \cap K)}{\mu(M_n(\mathbb{R}) \cap K)}.
\]
Thus the stated inequality follows. $\blacksquare$

\subsection{KL Divergence for Evidence of Lower Bound} 
\label{appendix: kl lower bound}
The KL divergence of Gaussian distribution~\citep{vae} is computed using mean and variance, which are the parameters of a Gaussian distribution. 
To introduce a loss as the KL divergence of Gaussian distribution, we compute KL divergence of the exponential family in Eq.~\ref{eq:posteior_} using the learnable parameter %$f_{\mathbf{x}}(\mathbf{x}|\boldsymbol{\theta}_{\hat{\mathbf{z}}})$ and $f_{\mathbf{x}}(\mathbf{x}|\boldsymbol{\theta}_{\hat{\boldsymbol{\epsilon}}})$ 
$T(\cdot)$ and $A(\cdot)$ with given natural parameter $\boldsymbol{\theta}_{\hat{ \vz}}$ and $\boldsymbol{\theta}_{\hat{\boldsymbol{\epsilon}}}$, expressed as:
\begin{equation}
    \begin{split}
         \mathcal{L}_{kl} &\coloneqq  \KL(f_{ \vx}( \vx|\boldsymbol{\theta}_{\hat{ \vz}_m}) || f_{ \vx}( \vx|\boldsymbol{\theta}_{\hat{\boldsymbol{\epsilon}}_m})) \\
        &= A(\boldsymbol{\theta}_{\hat{\boldsymbol{\epsilon}}}) - A (\boldsymbol{\theta}_{\hat{ \vz}}) + \boldsymbol{\theta}_{\hat{ \vz}}^{\intercal}  \nabla_{\boldsymbol{\theta}_{\hat{ \vz}}} A(\boldsymbol{\theta}_{\hat{ \vz}}) - \boldsymbol{\theta}_{\hat{\boldsymbol{\epsilon}}}^{\intercal}  \nabla_{\boldsymbol{\theta}_{\hat{\boldsymbol{\epsilon}}}} A(\boldsymbol{\theta}_{\hat{\boldsymbol{\epsilon}}}).
        \label{appendix eq:kl}
    \end{split}
\end{equation}
Because $ \KL(f_{ \vx}( \vx|\boldsymbol{\theta}_{\hat{ \vz}}) || f_{ \vx}( \vx|\boldsymbol{\theta}_{\hat{\boldsymbol{\epsilon}}}))$ is followed as:
%where $\cdot$ is an element-wise product.
%More details of Eq.~\ref{eq:kl} are in Appendix~\ref{appendix: kl}.

%The second term of Eq.~\ref{appendix: eq: elbo} is equal to $ \KL(f_{ \vx}( \vx|\boldsymbol{\theta}_{\hat{ \vz}}) || f_{ \vx}( \vx|\boldsymbol{\theta}_{\hat{\boldsymbol{\epsilon}}}))$ because power density function of posterior and prior are $f_{ \vx}( \vx|\boldsymbol{\theta}_{\hat{ \vz}}) $ and $f_{ \vx}( \vx|\boldsymbol{\theta}_{\hat{\boldsymbol{\epsilon}}})$, respectively.
\begin{equation}
% \small
    \begin{split}
     \KL(f_{ \vx}( \vx|\boldsymbol{\theta}_{\hat{ \vz}}) || f_{ \vx}( \vx|\boldsymbol{\theta}_{\hat{\boldsymbol{\epsilon}}})) & = \int^{\infty}_{-\infty} f_{ \vx} ( \vx | \boldsymbol{\theta}_{\hat{ \vz}}) \log f_{ \vx} ( \vx | \boldsymbol{\theta}_{\hat{ \vz}}) \mathsf{d}  \vx \\
    & - \int^{\infty}_{-\infty} f_{ \vx} ( \vx | \boldsymbol{\theta}_{\hat{ \vz}}) \log f_{ \vx} ( \vx | \boldsymbol{\theta}_{\hat{\boldsymbol{\epsilon}}}) \mathsf{d}  \vx.
    \end{split}
\label{appe: kl_divergence}
\end{equation}
We designed sufficient statistics as matrix multiplication (multi-layer perceptron).
%Log-normalizer is a function with MLP and a sigmoid function, and sigmoid is activated after last layer.%and by the exponential family's power density function,
Then, 
% \scriptsize
% \begin{equation}
% \begin{split}
%     \int^{\infty}_{-\infty} f_{ \vx} ( \vx | \boldsymbol{\theta}_{\hat{ \vz}}) \log f_{ \vx} ( \vx | \boldsymbol{\theta}_{\hat{ \vz}}) \mathsf{d}  \vx & = \int^{\infty}_{-\infty} f_{ \vx} ( \vx| \boldsymbol{\theta}_{\hat{ \vz}}) \cdot \\ &~~~~~~~~~~~~~~[\boldsymbol{\theta}_{\hat{ \vz}}^\intercal \mathbf{T}( \vx) - A(\boldsymbol{\theta}_{\hat{ \vz}}) + B( \vx)] \text{d} \vx \\
%     & = -A(\boldsymbol{\theta}_{\hat{ \vz}}) \int^{\infty}_{-\infty} f_{ \vx} ( \vx | \boldsymbol{\theta}_{\hat{ \vz}}) \text{d} \vx \\ 
%     & + \int^{\infty}_{-\infty} f_{ \vx} ( \vx | \boldsymbol{\theta}_{\hat{ \vz}}) [\boldsymbol{\theta}_{\hat{ \vz}}^\intercal \mathbf{T}( \vx) + B( \vx)]\text{d} \vx\\
%     & = -A(\boldsymbol{\theta}_{\hat{ \vz}}) + \boldsymbol{\theta}_{\hat{ \vz}}^\intercal  \int^{\infty}_{-\infty} T( \vx) f_{ \vx} ( \vx| \boldsymbol{\theta}_{\hat{ \vz}})  \text{d} \vx \\
%     & + \int^{\infty}_{-\infty} B( \vx) f_{ \vx} ( \vx| \boldsymbol{\theta}_{\hat{ \vz}}) \text{d} \vx, 
%     \label{appe:exp_mean1}
% \end{split}
% \end{equation}
\begin{equation}
\scalebox{0.80}{%
$\begin{aligned}
    \int^{\infty}_{-\infty} f_{ \vx} ( \vx | \boldsymbol{\theta}_{\hat{ \vz}}) \log f_{ \vx} ( \vx | \boldsymbol{\theta}_{\hat{ \vz}}) \mathsf{d}  \vx & = \int^{\infty}_{-\infty} f_{ \vx} ( \vx| \boldsymbol{\theta}_{\hat{ \vz}}) \cdot \\
    &~~~~~~~~~~~~~~[\boldsymbol{\theta}_{\hat{ \vz}}^\intercal \mathbf{T}( \vx) - A(\boldsymbol{\theta}_{\hat{ \vz}}) + B( \vx)] \text{d} \vx 
    % \\
    % & = -A(\boldsymbol{\theta}_{\hat{ \vz}}) \int^{\infty}_{-\infty} f_{ \vx} ( \vx | \boldsymbol{\theta}_{\hat{ \vz}}) \text{d} \vx \\ 
    % & + \int^{\infty}_{-\infty} f_{ \vx} ( \vx | \boldsymbol{\theta}_{\hat{ \vz}}) [\boldsymbol{\theta}_{\hat{ \vz}}^\intercal \mathbf{T}( \vx) + B( \vx)]\text{d} \vx\\
    % & = -A(\boldsymbol{\theta}_{\hat{ \vz}}) + \boldsymbol{\theta}_{\hat{ \vz}}^\intercal  \int^{\infty}_{-\infty} T( \vx) f_{ \vx} ( \vx| \boldsymbol{\theta}_{\hat{ \vz}})  \text{d} \vx \\
    % & + \int^{\infty}_{-\infty} B( \vx) f_{ \vx} ( \vx| \boldsymbol{\theta}_{\hat{ \vz}}) \text{d} \vx,
\end{aligned}$%
}
\label{appe:exp_mean1}
\end{equation}
Expanding the Eq.~\ref{appe:exp_mean1}, and $\int^{\infty}_{-\infty} f_{ \vx} ( \vx | \boldsymbol{\theta}_{\hat{ \vz}}) \text{d} \vx = 1$, then
\begin{equation}
\scalebox{0.80}{%
$\begin{aligned}
    \int^{\infty}_{-\infty} f_{ \vx} ( \vx | \boldsymbol{\theta}_{\hat{ \vz}}) \log f_{ \vx} ( \vx | \boldsymbol{\theta}_{\hat{ \vz}}) \mathsf{d}  \vx
    % & = -A(\boldsymbol{\theta}_{\hat{ \vz}}) \int^{\infty}_{-\infty} f_{ \vx} ( \vx | \boldsymbol{\theta}_{\hat{ \vz}}) \text{d} \vx \\ 
    % & + \int^{\infty}_{-\infty} f_{ \vx} ( \vx | \boldsymbol{\theta}_{\hat{ \vz}}) [\boldsymbol{\theta}_{\hat{ \vz}}^\intercal \mathbf{T}( \vx) + B( \vx)]\text{d} \vx\\
    & = -A(\boldsymbol{\theta}_{\hat{ \vz}}) + \boldsymbol{\theta}_{\hat{ \vz}}^\intercal  \int^{\infty}_{-\infty} T( \vx) f_{ \vx} ( \vx| \boldsymbol{\theta}_{\hat{ \vz}})  \text{d} \vx \\
    & + \int^{\infty}_{-\infty} B( \vx) f_{ \vx} ( \vx| \boldsymbol{\theta}_{\hat{ \vz}}) \text{d} \vx.
\end{aligned}$%
}
\label{appe:exp_mean1-1}
\end{equation}

% and
Expanding $\int^{\infty}_{-\infty} f_{ \vx} ( \vx | \boldsymbol{\theta}_{\hat{ \vz}}) \log f_{ \vx} ( \vx | \boldsymbol{\theta}_{\hat{\boldsymbol{\epsilon}}}) \mathsf{d}  \vx$ using Eq.~\ref{appe:exp_mean1} and~\ref{appe:exp_mean1-1} and plug into Eq.~\ref{appe: kl_divergence} with Eq.~\ref{appe:exp_mean1-1}, we have
% yields the following:
% \begin{equation}
% % \small
% \begin{split}
%     \int^{\infty}_{-\infty} f_{ \vx} ( \vx | \boldsymbol{\theta}_{\hat{ \vz}}) \log f_{ \vx} ( \vx | \boldsymbol{\theta}_{\hat{\boldsymbol{\epsilon}}}) \mathsf{d}  \vx & = -A(\boldsymbol{\theta}_{\hat{\boldsymbol{\epsilon}}}) \\ 
%     & + \boldsymbol{\theta}_{\hat{\boldsymbol{\epsilon}}}^\intercal  \int^{\infty}_{-\infty} T( \vx) f_{ \vx} ( \vx | \boldsymbol{\theta}_{\hat{\boldsymbol{\epsilon}}})  \text{d} \vx \\
%     & + \int^{\infty}_{-\infty} B( \vx) f_{ \vx} ( \vx | \boldsymbol{\theta}_{\hat{\boldsymbol{\epsilon}}}) \text{d} \vx.
% \label{appe:exp_mean2}
% \end{split}
% \end{equation}
% Plugging Eq.~\ref{appe:exp_mean1-1} and~\ref{appe:exp_mean2} into Eq.~\ref{appe: kl_divergence}, we have
%\resizebox{1.0\linewidth}{!}{
%\begin{minipage}{1.05\linewidth}
\begin{equation}
\begin{split}
    \KL (f_{ \vx}( \vx|\boldsymbol{\theta}_{\hat{ \vz}}) || f_{ \vx}( \vx|\boldsymbol{\theta}_{\hat{\boldsymbol{\epsilon}}})) & = A(\boldsymbol{\theta}_{\hat{\boldsymbol{\epsilon}}}) - A(\boldsymbol{\theta}_{\hat{ \vz}}) \\ 
    & + \boldsymbol{\theta}_{\hat{ \vz}} \int^{\infty}_{-\infty} T( \vx) f_{ \vx} ( \vx | \boldsymbol{\theta}_{\hat{ \vz}})  \text{d} \vx \\
    & - \boldsymbol{\theta}_{\hat{\boldsymbol{\epsilon}}} \int^{\infty}_{-\infty} T( \vx) f_{ \vx} ( \vx | \boldsymbol{\theta}_{\hat{\boldsymbol{\epsilon}}})  \text{d} \vx.
    \label{appendix: eq: 28}
\end{split}
\end{equation}
%\end{minipage}
%}

%Then, $\int^{\infty}_{-\infty} \mathbf{T}(\mathbf{x}) f_{\mathbf{x}} (\mathbf{x} | \boldsymbol{\theta}_{\hat{\mathbf{z}}})  \text{d}\mathbf{x} =  \int^{\infty}_{-\infty} \mathbf{T}(\mathbf{x}) f_{\mathbf{x}} (\mathbf{x} | \boldsymbol{\theta}_{\hat{\boldsymbol{\epsilon}}}) \text{d}\mathbf{x} =  \mathbf{T} \mathbb{E}_{\mathbf{x}}[\mathbf{x}]$ because $\mathbf{T}(\mathbf{x}) = \mathbf{x}^\intercal\mathbf{T}$.
%To estimate the mean of distribution ($\mathbb{E}_{\mathbf{x}}[\mathbf{x}]$), we approximate it as like mini-batch sampling, which is introduced in~\cite{beta-tcvae}.
%Our approximation is followed as:
The gradient of the log-partition function with respect to the natural parameters yields the expected value of the sufficient statistics: $\nabla_{\boldsymbol{\theta}}A(\boldsymbol{\theta}) = \mathbb{E}_{p(x;\boldsymbol{\theta})}[T(x)]$~\citep{prml}.
Then the mean of the sufficient statistic is followed as:
% \begin{equation}
% % \small
%      \int^{\infty}_{-\infty} T( \vx) f_{ \vx} ( \vx | \boldsymbol{\theta})  \text{d} \vx = \frac{\partial A^{\ast} (\boldsymbol{\theta})}{\partial\boldsymbol{\theta}} \approx \frac{\partial A(\boldsymbol{\theta})}{\partial\boldsymbol{\theta}} ~~~~~\because A^{\ast} (\boldsymbol{\theta}) = \boldsymbol{\theta}^\intercal A^{\ast},
% \end{equation}
\begin{equation}
\scalebox{0.85}{%
$\displaystyle
     \int^{\infty}_{-\infty} T( \vx) f_{ \vx} ( \vx | \boldsymbol{\theta})  \text{d} \vx = \frac{\partial A^{\ast} (\boldsymbol{\theta})}{\partial\boldsymbol{\theta}} \approx \frac{\partial A(\boldsymbol{\theta})}{\partial\boldsymbol{\theta}} ~~~~~\because A^{\ast} (\boldsymbol{\theta}) = \boldsymbol{\theta}^\intercal A^{\ast},
$%
}
\label{eq:dual_grad_approx}
\end{equation}
where $A^{\ast}(\cdot)$ is a true log-partition function of the exponential family (ideal case of $A(\cdot)$).
However, estimating $A^\ast$ is difficult, and there is no direct method without random samplings, such as mini-batch weighted sampling or mini-batch stratified sampling~\citep{beta-tcvae}.
Then, we approximate $A^\ast$ to $A$, and train $A$ to be close to $A^\ast$.
Plugging Eq.~\ref{eq:dual_grad_approx} into Eq.~\ref{appe:exp_mean1-1} then we have
% and Eq.~\ref{appe:exp_mean2}, we obtain KL divergence of the exponential family as:
%Integration at second term in Eq.~\ref{appe:exp_mean}, is a mean of distribution, which is equal to derivation of $\mathbf{A}(\boldsymbol{\theta}_{\hat{\mathbf{z}}})$ over $\boldsymbol{\theta}_{\hat{\mathbf{z}}}$.
\begin{equation}
\begin{split}
    \int^{\infty}_{-\infty} f_{ \vx} (\mathbf{x} | \boldsymbol{\theta}_{\hat{ \vz}}) \log f_{ \vx} ( \vx | \boldsymbol{\theta}_{\hat{ \vz}}) & = -A(\boldsymbol{\theta}_{\hat{ \vz}}) + \boldsymbol{\theta}_{\hat{ \vz}}^{\intercal} \frac{\partial A(\boldsymbol{\theta}_{\hat{ \vz}})}{\partial \boldsymbol{\theta}_{\hat{ \vz}}} \\ 
    & + \int^{\infty}_{-\infty} f_{ \vx} ( \vx | \boldsymbol{\theta}_{\hat{ \vz}}) B( \vx) \text{d} \vx,
\end{split}
\label{appe: kld_sub1}
\end{equation}
Applying same work to $\int^{\infty}_{-\infty} f_{ \vx} ( \vx | \boldsymbol{\theta}_{\hat{ \vz}}) \log f_{ \vx} ( \vx | \boldsymbol{\theta}_{\hat{\boldsymbol{\epsilon}}}) \mathsf{d}  \vx$ and plug into Eq.~\ref{appe: kl_divergence} with Eq.~\ref{appe: kld_sub1},
% \begin{equation}
% \begin{split}
%     \int^{\infty}_{-\infty} f_{ \vx} ( \vx | \boldsymbol{\theta}_{\hat{ \vz}}) \log f_{ \vx}(\mathbf{x}|\boldsymbol{\theta}_{\hat{\boldsymbol{\epsilon}}}) & = -Z(\boldsymbol{\theta}_{\hat{\boldsymbol{\epsilon}}}) + \boldsymbol{\theta}_{\hat{\boldsymbol{\epsilon}}}^{\intercal} \frac{\partial A(\boldsymbol{\theta}_{\hat{\boldsymbol{\epsilon}}})}{\partial \boldsymbol{\theta}_{\hat{\boldsymbol{\epsilon}}}} \\ 
%     & + \int^{\infty}_{-\infty} f_{ \vx} ( \vx | \boldsymbol{\theta}_{\hat{ \vz}}) B( \vx) \text{d} \vx.
% \end{split}
% \label{appe: kld_sub2}
% \end{equation}
% Plugging Eq.~\ref{appe: kld_sub1} and~\ref{appe: kld_sub2} into Eq.~\ref{appe: kl_divergence}, 
then the final Kullback-Leibler divergence of exponential family is followed as:
% \begin{equation}
% % \scriptsize
%     \begin{split}
%      \KL(f_{ \vx}( \vx|\boldsymbol{\theta}_{\hat{ \vz}}) || f_{ \vx}( \vx|\boldsymbol{\theta}_{\hat{\boldsymbol{\epsilon}}})) & = A(\boldsymbol{\theta}_{\hat{\boldsymbol{\epsilon}}}) - A (\boldsymbol{\theta}_{\hat{ \vz}}) + \boldsymbol{\theta}_{\hat{ \vz}}^{\intercal} \frac{\partial A(\boldsymbol{\theta}_{\hat{ \vz}})}{\partial \boldsymbol{\theta}_{\hat{ \vz}}} - \boldsymbol{\theta}_{\hat{\boldsymbol{\epsilon}}}^{\intercal} \frac{\partial A(\boldsymbol{\theta}_{\hat{\boldsymbol{\epsilon}}})}{\partial \boldsymbol{\theta}_{\hat{\boldsymbol{\epsilon}}}}.    
%     \end{split}
% \end{equation}
\begin{equation}
\scalebox{0.80}{%
$\begin{aligned}
     \KL(f_{ \vx}( \vx|\boldsymbol{\theta}_{\hat{ \vz}}) \,\|\, f_{ \vx}( \vx|\boldsymbol{\theta}_{\hat{\boldsymbol{\epsilon}}}))
     & = A(\boldsymbol{\theta}_{\hat{\boldsymbol{\epsilon}}}) 
     - A (\boldsymbol{\theta}_{\hat{ \vz}}) 
     + \boldsymbol{\theta}_{\hat{ \vz}}^{\intercal} \frac{\partial A(\boldsymbol{\theta}_{\hat{ \vz}})}{\partial \boldsymbol{\theta}_{\hat{ \vz}}} 
     - \boldsymbol{\theta}_{\hat{\boldsymbol{\epsilon}}}^{\intercal} \frac{\partial A(\boldsymbol{\theta}_{\hat{\boldsymbol{\epsilon}}})}{\partial \boldsymbol{\theta}_{\hat{\boldsymbol{\epsilon}}}}.
\end{aligned}$%
}
\label{eq:kl_two_param}
\vspace{-10pt}
\end{equation}

% \subsection{Integration for Multiple IPE-Transformation and EF-Conversion}
\section{Objective Function}
\label{appendix: multiple unit}
%IE-transformation is a process to restrict the distribution of latent vectors for disentanglement, which possibly reduce reconstruction error~\cite{}.
%This trade-off is possibly reduced by increasing the transformation unit as the report in~\cite{}. 
We mathematically extend IPE-transformation to MIPE-transformation, which is the equivalent process of $\beta$-VAE to enhance disentanglement. 
%The equivalence is proven in Appendix~\ref{appendix: elbo}.
Each IPE-transformation function operates independently, then the reconstruction error for objective function is defined as:
%$\log p_{\theta}( \vx|\hat{ \vz}_1, \hat{ \vz}_2, \cdots, \hat{ \vz}_k) =$ $-(k-1) \log p_\theta( \vx) + \Pi_{i=1}^k p_\theta( \vx | \hat{ \vz}_i )$. 
%more details are in Appendix~\ref{appendix: elbo}.
%Therefore, the reconstruction error term ($\mathcal{L}_{rec}$) for given all $\hat{\mathbf{z}}_i$ in Eq.~\ref{eq:objective_function} is 
%supplementary material.
% \begin{equation}
% % \scriptsize
%     \begin{split}
%        \mathcal{L}_{rec} &\coloneqq \frac{1}{k} \sum_{i=1}^k \Bigg [ \int q_i(\hat{ \vz}_i| \vx) \log p_\theta( \vx|\hat{ \vz}_i) \mathsf{d}\hat{ \vz}_i \prod_{j=1, j\neq i}^k \int q_j(\hat{ \vz}_j| \vx) %q_3(\mathbf{z}_3|\mathbf{x}) \cdots q_k(\mathbf{z}_k|\mathbf{x})
%        \text{d}\hat{ \vz}_j \Bigg ] \\
%        %\text{d}\mathbf{z}_3 \cdots \text{d}\mathbf{z}_k\\
%     &= \frac{1}{k} \sum_{i=1}^k  E_{q_{\phi,\psi_i} ( \vz| \vx)} \log p_\theta( \vx|\psi_i( \vz)),
%     \end{split}
%     \label{appendix eq: reconstruction error}
% \end{equation}
\begin{equation}
\scalebox{0.85}{%
$\begin{aligned}
       \mathcal{L}_{rec} &\coloneqq \frac{1}{k} \sum_{i=1}^k \Bigg [ \int q_i(\hat{ \vz}_i| \vx) \log p_\theta( \vx|\hat{ \vz}_i) \mathsf{d}\hat{ \vz}_i \prod_{j=1,\, j\neq i}^k \int q_j(\hat{ \vz}_j| \vx) \text{d}\hat{ \vz}_j \Bigg ] \\
    &= \frac{1}{k} \sum_{i=1}^k  E_{q_{\phi,\psi_i} ( \vz| \vx)} \log p_\theta( \vx|\psi_i( \vz)),
\end{aligned}$%
}
\label{appendix eq: reconstruction error}
\end{equation}

where $\hat{ \vz}_i = \psi_i( \vz)$.
Becuase the log likelihood of $p( \vx)$ can be derived as follows:
% \scriptsize
% \begin{align}
%     \log p_\theta( \vx) & = \int \prod_i^k q_1(\hat{ \vz}_i |  \vx) \log p_\theta( \vx) \mathsf{d} \hat{ \vz}^\prime \\
%     & = \int \prod_i^k q_1(\hat{ \vz}_i |  \vx) \log \frac{p_\theta( \vx, \hat{ \vz}_1, \hat{ \vz}_2, \cdots ,\hat{ \vz}_k)}{p_\theta(\hat{ \vz}_1, \hat{ \vz}_2, \cdots ,\hat{ \vz}_k| \vx)} \mathsf{d}\hat{ \vz}^\prime\\
%     \begin{split}
%     & = \int \prod_i^k q_1(\hat{ \vz}_i |  \vx) \cdot \\
%     & ~~~~\Big [ \log \frac{p_\theta( \vx,\hat{ \vz}_1, \hat{ \vz}_2, \cdots ,\hat{ \vz}_k)}{q(\hat{ \vz}_1, \hat{ \vz}_2, \cdots ,\hat{ \vz}_k| \vx)} - \log \frac{p_\theta(\hat{ \vz}_1, \hat{ \vz}_2, \cdots ,\hat{ \vz}_k | \vx)}{q(\hat{ \vz}_1, \hat{ \vz}_2, \cdots ,\hat{ \vz}_k | \vx)} \Big ] \mathsf{d} \hat{ \vz}^\prime
%     \end{split} \\
%     & \geqq \int \prod_i^k q_1(\hat{ \vz}_i |  \vx) \log \frac{p_\theta( \vx, \hat{ \vz}_1, \hat{ \vz}_2, \cdots ,\hat{ \vz}_k)}{q(\hat{ \vz}_1, \hat{ \vz}_2, \cdots ,\hat{ \vz}_k| \vx)} \mathsf{d} \hat{ \vz}^\prime\\
%     \begin{split}
%         & = \int \prod_i^k q_1(\hat{ \vz}_i |  \vx) \cdot \\
%     & ~~~~\Big [ \log p_\theta( \vx|\hat{ \vz}_1, \hat{ \vz}_2, \cdots ,\hat{ \vz}_k) + \log \frac{p(\hat{ \vz}_1, \hat{ \vz}_2, \cdots ,\hat{ \vz}_k)}{q(\hat{ \vz}_1, \hat{ \vz}_2, \cdots ,\hat{ \vz}_k | \vx)} \Big ] \mathsf{d} \hat{ \vz}^\prime,
%     \end{split}
% \end{align}
\begin{equation}
    \log p_\theta( \vx) = \int \prod_{i=1}^k q_i(\hat{ \vz}_i |  \vx) \log p_\theta( \vx) \mathsf{d} \hat{ \vz}^\prime,
\label{eq:elbo_multi}
\end{equation}
where d$\hat{ \vz}^\prime = \text{d}\hat{ \vz}_1 \text{d}\hat{ \vz}_2 \cdots \text{d}\hat{ \vz}_k$.
Expanding Eq.~\ref{eq:elbo_multi} as $\log p_\theta( \vx) 
= \log \frac{p_\theta( \vx, \hat{ \vz}_1, \hat{ \vz}_2, \cdots ,\hat{ \vz}_k)}{p_\theta(\hat{ \vz}_1, \hat{ \vz}_2, \cdots ,\hat{ \vz}_k| \vx)}
= \log \frac{p_\theta( \vx,\hat{ \vz}_1, \hat{ \vz}_2, \cdots ,\hat{ \vz}_k)}{q(\hat{ \vz}_1, \hat{ \vz}_2, \cdots ,\hat{ \vz}_k| \vx)} - \log \frac{p_\theta(\hat{ \vz}_1, \hat{ \vz}_2, \cdots ,\hat{ \vz}_k | \vx)}{q(\hat{ \vz}_1, \hat{ \vz}_2, \cdots ,\hat{ \vz}_k | \vx)}$. 
Then
\begin{equation}
\scalebox{1.0}{%
$\begin{aligned}
    \log p_\theta( \vx) 
    % & = \int \prod_{i=1}^k q_1(\hat{ \vz}_i |  \vx) \log p_\theta( \vx) \mathsf{d} \hat{ \vz}^\prime \\
    % & = \int \prod_{i=1}^k q_1(\hat{ \vz}_i |  \vx) \log \frac{p_\theta( \vx, \hat{ \vz}_1, \hat{ \vz}_2, \cdots ,\hat{ \vz}_k)}{p_\theta(\hat{ \vz}_1, \hat{ \vz}_2, \cdots ,\hat{ \vz}_k| \vx)} \mathsf{d}\hat{ \vz}^\prime \\
    % & = \int \prod_{i=1}^k q_1(\hat{ \vz}_i |  \vx) 
    % \Big[ \log \frac{p_\theta( \vx,\hat{ \vz}_1, \hat{ \vz}_2, \cdots ,\hat{ \vz}_k)}{q(\hat{ \vz}_1, \hat{ \vz}_2, \cdots ,\hat{ \vz}_k| \vx)} 
    % - \log \frac{p_\theta(\hat{ \vz}_1, \hat{ \vz}_2, \cdots ,\hat{ \vz}_k | \vx)}{q(\hat{ \vz}_1, \hat{ \vz}_2, \cdots ,\hat{ \vz}_k | \vx)} \Big] 
    % \mathsf{d} \hat{ \vz}^\prime \\
    \geqq \int \prod_{i=1}^k q_1(\hat{ \vz}_i |  \vx) \log \frac{p_\theta( \vx, \hat{ \vz}_1, \hat{ \vz}_2, \cdots ,\hat{ \vz}_k)}{q(\hat{ \vz}_1, \hat{ \vz}_2, \cdots ,\hat{ \vz}_k| \vx)} \mathsf{d} \hat{ \vz}^\prime.
    % & = \int \prod_{i=1}^k q_1(\hat{ \vz}_i |  \vx) 
    % \Big[ \log p_\theta( \vx|\hat{ \vz}_1, \hat{ \vz}_2, \cdots ,\hat{ \vz}_k) 
    % + \log \frac{p(\hat{ \vz}_1, \hat{ \vz}_2, \cdots ,\hat{ \vz}_k)}{q(\hat{ \vz}_1, \hat{ \vz}_2, \cdots ,\hat{ \vz}_k | \vx)} \Big] 
    % \mathsf{d} \hat{ \vz}^\prime,
\end{aligned}$%
}
\label{eq:elbo_multi_1}
\end{equation}

Expanding Eq.~\ref{eq:elbo_multi_1} then
\begin{equation}
\scalebox{1.0}{%
$\begin{aligned}
    \log p_\theta( \vx) 
    & \geqq \int \prod_{i=1}^k q_i(\hat{ \vz}_i |  \vx) 
    \Big[ \log p_\theta( \vx|\hat{ \vz}_1, \hat{ \vz}_2, \cdots ,\hat{ \vz}_k) \\
    &~~~ + \log \frac{p(\hat{ \vz}_1, \hat{ \vz}_2, \cdots ,\hat{ \vz}_k)}{q(\hat{ \vz}_1, \hat{ \vz}_2, \cdots ,\hat{ \vz}_k | \vx)} \Big] 
    \mathsf{d} \hat{ \vz}^\prime.
\end{aligned}$%
}
\label{eq:elbo_multi_2}
\end{equation}
% \begin{equation}
% \scalebox{0.70}{%
% $\begin{aligned}
%     \log p_\theta( \vx) & = \int \prod_{i=1}^k q_1(\hat{ \vz}_i |  \vx) \log p_\theta( \vx) \mathsf{d} \hat{ \vz}^\prime \\
%     & = \int \prod_{i=1}^k q_1(\hat{ \vz}_i |  \vx) \log \frac{p_\theta( \vx, \hat{ \vz}_1, \hat{ \vz}_2, \cdots ,\hat{ \vz}_k)}{p_\theta(\hat{ \vz}_1, \hat{ \vz}_2, \cdots ,\hat{ \vz}_k| \vx)} \mathsf{d}\hat{ \vz}^\prime \\
%     & = \int \prod_{i=1}^k q_1(\hat{ \vz}_i |  \vx) 
%     \Big[ \log \frac{p_\theta( \vx,\hat{ \vz}_1, \hat{ \vz}_2, \cdots ,\hat{ \vz}_k)}{q(\hat{ \vz}_1, \hat{ \vz}_2, \cdots ,\hat{ \vz}_k| \vx)} 
%     - \log \frac{p_\theta(\hat{ \vz}_1, \hat{ \vz}_2, \cdots ,\hat{ \vz}_k | \vx)}{q(\hat{ \vz}_1, \hat{ \vz}_2, \cdots ,\hat{ \vz}_k | \vx)} \Big] 
%     \mathsf{d} \hat{ \vz}^\prime \\
%     & \geqq \int \prod_{i=1}^k q_1(\hat{ \vz}_i |  \vx) \log \frac{p_\theta( \vx, \hat{ \vz}_1, \hat{ \vz}_2, \cdots ,\hat{ \vz}_k)}{q(\hat{ \vz}_1, \hat{ \vz}_2, \cdots ,\hat{ \vz}_k| \vx)} \mathsf{d} \hat{ \vz}^\prime \\
%     & = \int \prod_{i=1}^k q_1(\hat{ \vz}_i |  \vx) 
%     \Big[ \log p_\theta( \vx|\hat{ \vz}_1, \hat{ \vz}_2, \cdots ,\hat{ \vz}_k) 
%     + \log \frac{p(\hat{ \vz}_1, \hat{ \vz}_2, \cdots ,\hat{ \vz}_k)}{q(\hat{ \vz}_1, \hat{ \vz}_2, \cdots ,\hat{ \vz}_k | \vx)} \Big] 
%     \mathsf{d} \hat{ \vz}^\prime,
% \end{aligned}$%
% }
% \label{eq:elbo_multi}
% \end{equation}
The first term of Eq.~\ref{eq:elbo_multi_2} is expanded as:
\begin{equation}
    p_\theta ( \vx|\hat{ \vz}_1, \hat{ \vz}_2, \ldots , \hat{ \vz}_k) 
    = \frac{p_\theta (\hat{ \vz}_1, \hat{ \vz}_2, \ldots, \hat{ \vz}_k |  \vx) p_\theta( \vx)}{p_\theta (\hat{ \vz}_1, \hat{ \vz}_2, \ldots, \hat{ \vz}_k)}.
\label{appe: eq 30}
\end{equation}
Each IPE-transformation function operates independently, then all $\hat{\vz}_i$ are independent and conditionally independent given $\vx$.
Then we have 
\begin{equation}
    p_\theta ( \vx|\hat{ \vz}_1, \hat{ \vz}_2, \ldots , \hat{ \vz}_k) 
    = \frac{p_\theta(\vx) \prod_{i=1}^k p_\theta(\hat{\vz}_i|\vx)}{\prod_{i=1}^k p_\theta(\hat{\vz}_i)},
\label{appe: eq 31}
    % \quad (\because~ \hat{\vz}_i \indep \hat{\vz}_j~|~\vx)
\end{equation}
where $\prod_{i=1}^k p_\theta(\hat{\vz}_i|\vx) = p_\theta (\hat{ \vz}_1, \hat{ \vz}_2, \ldots, \hat{ \vz}_k |  \vx)$ and $\prod_{i=1}^k p_\theta(\hat{\vz}_i) = p_\theta (\hat{ \vz}_1, \hat{ \vz}_2, \ldots, \hat{ \vz}_k)$.
Let replace $p_\theta(\vx)$ as $p_\theta(\vx)^{-(k-1)} \prod_{i=1}^k p_\theta(\vx)$.
Then we have
\begin{equation}
    p_\theta ( \vx|\hat{ \vz}_1, \hat{ \vz}_2, \ldots , \hat{ \vz}_k) 
    = p_\theta( \vx)^{-(k-1)} \prod_{i=1}^k p_{\theta} ( \vx | \hat{ \vz}_i),
\label{eq:product_form}
\end{equation}
Plugging Eq.~\ref{eq:product_form} into the first term of Eq.~\ref{eq:elbo_multi_2}, then we have
\begin{equation}
\scalebox{0.90}{%
$\begin{aligned}
     & \int \prod_{i=1}^k q_i(\hat{ \vz}_i| \vx) \log p_{\theta}( \vx|\hat{ \vz}_1, \hat{ \vz}_2, \cdots, \hat{ \vz}_k) \mathsf{d} \hat{ \vz}^\prime \\
    % &= \int \prod_{i=1}^k q_i(\hat{ \vz}_i| \vx)  \Big [ -(k-1) \log p_\theta( \vx) + \prod_{i=1}^k p_\theta( \vx | \hat{ \vz}_i ) \Big ] \mathsf{d} \hat{ \vz}^\prime \\ 
    &= -(k-1) \log p_\theta( \vx) + \int \prod_{i=1}^k q_i(\hat{ \vz}_i| \vx) \Big [ \sum_{j=1}^k \log p_\theta( \vx | \hat{ \vz}_j )\Big ]  \mathsf{d} \hat{ \vz}^\prime .
\end{aligned}$%
}
\label{eq:product_form_1}
\end{equation}
Applying same process between the Eq.~\ref{appe: eq 30} and Eq.~\ref{appe: eq 31}, and plugging Eq.~\ref{eq:product_form_1} into Eq.~\ref{eq:elbo_multi_2}, then we have
\begin{equation}
\scalebox{1.0}{%
$\begin{aligned}
    k \log p_\theta( \vx) 
    & \geqq \int \prod_{i=1}^k q_i(\hat{ \vz}_i| \vx) \Big [ \sum_{j=1}^k \log p_\theta( \vx | \hat{ \vz}_j ) \Big ] \mathsf{d} \hat{ \vz}^\prime  \\
    &~~~ + \int \prod_{i=1}^k q_i(\hat{ \vz}_i |  \vx)  
     \log \frac{\prod_{i=1}^k p(\hat{ \vz}_i)}{\prod_{i=1}^k q_i(\hat{ \vz}_i| \vx)} \mathsf{d}\hat{ \vz}^\prime.
\end{aligned}$%
}
\label{appe: k_elbo}
\end{equation}
% \begin{equation}
% % \scriptsize
%     \begin{split}
%        \log p_{\theta}( \vx) & \geq \frac{1}{k}\sum_{i=1}^k \Bigg [ \int q_i(\hat{ \vz}_i| \vx) \log p_\theta( \vx|\hat{ \vz}_i) \mathsf{d}\hat{ \vz}_i \prod_{j=1, j\neq i}^k \int q_j(\hat{ \vz}_j| \vx) %q_3(\mathbf{z}_3|\mathbf{x}) \cdots q_k(\mathbf{z}_k|\mathbf{x})
%        \text{d}\hat{ \vz}_j \Bigg ] \\
%        %\text{d}\mathbf{z}_3 \cdots \text{d}\mathbf{z}_k\\
%     & - \int \prod_i^k q_1(\hat{ \vz}_i |  \vx) \log \frac{\prod_{i=1}^k q_i(\hat{ \vz}_i| \vx)}{\prod_{i=1}^k p(\hat{ \vz}_i)} \mathsf{d}\hat{ \vz}^\prime \\
%     %\end{split}\\
%     %\begin{split}
%     & = \frac{1}{k} \sum_{i=1}^k  \E_{q(\hat{ \vz}_i| \vx )} \log p_\theta( \vx|\hat{ \vz}_i) \\
%     & - \sum_{i=1}^k \bigg [ \KL (q_{\phi}(\hat{ \vz}_i| \vx) || p(\hat{ \vz}_i)) \prod_{j=1, j\neq i}^k \int q_j(\hat{ \vz}_j| \vx) 
%     \text{d}\hat{ \vz}_j \bigg ] \\ %\end{split} \\
%     & = \frac{1}{k} \sum_{i=1}^k  \E_{q_{\phi} (\hat{ \vz}_i| \vx)} \log p_\theta( \vx|\hat{ \vz}_i) - \sum_{i=1}^k %q_{\phi, \psi_j}(z|x) modify this part.
%      \KL (q_{\phi}(\hat{ \vz}_i| \vx) || p( \vz_i))\\
%     & = \frac{1}{k} \Big [ \sum_{i=1}^k \E_{q_{\phi} (\hat{ \vz}_i| \vx)} \log p_\theta( \vx|\hat{ \vz}_i) - k  \KL (q_{\phi}(\hat{ \vz}_i| \vx) || p(\hat{ \vz}_i)) \Big ].
% \end{split}
% \end{equation}
Expanding the first term and inverse the second term of Eq.~\ref{appe: k_elbo}, then we have
\begin{equation}
\scalebox{0.85}{%
$\begin{aligned}
       \log p_{\theta}( \vx) & \geq \frac{1}{k}\sum_{i=1}^k \Bigg [ \int q_i(\hat{ \vz}_i| \vx) \log p_\theta( \vx|\hat{ \vz}_i) \mathsf{d}\hat{ \vz}_i \prod_{j=1,\, j\neq i}^k \int q_j(\hat{ \vz}_j| \vx) \text{d}\hat{ \vz}_j \Bigg ] \\
    &\quad - \sum_{i=1}^k \KL \left(q_{\phi}(\hat{ \vz}_i| \vx) \,\|\, p( \vz_i)\right)
    % &\quad - \int \prod_{i=1}^k q_i(\hat{ \vz}_i |  \vx) \log \frac{\prod_{i=1}^k q_i(\hat{ \vz}_i| \vx)}{\prod_{i=1}^k p(\hat{ \vz}_i)} \mathsf{d}\hat{ \vz}^\prime \\
    % & = \frac{1}{k} \sum_{i=1}^k  \E_{q(\hat{ \vz}_i| \vx )} \log p_\theta( \vx|\hat{ \vz}_i) \\
    % &\quad - \sum_{i=1}^k \bigg [ \KL \left(q_{\phi}(\hat{ \vz}_i| \vx) \,\|\, p(\hat{ \vz}_i)\right) \prod_{j=1,\, j\neq i}^k \int q_j(\hat{ \vz}_j| \vx) \text{d}\hat{ \vz}_j \bigg ] \\
    % & = \frac{1}{k} \sum_{i=1}^k  \E_{q_{\phi} (\hat{ \vz}_i| \vx)} \log p_\theta( \vx|\hat{ \vz}_i) - \sum_{i=1}^k \KL \left(q_{\phi}(\hat{ \vz}_i| \vx) \,\|\, p( \vz_i)\right) \\
    % & = \frac{1}{k} \Bigg[ \sum_{i=1}^k \E_{q_{\phi} (\hat{ \vz}_i| \vx)} \log p_\theta( \vx|\hat{ \vz}_i) - k\, \KL \left(q_{\phi}(\hat{ \vz}_i| \vx) \,\|\, p(\hat{ \vz}_i)\right) \Bigg]
\end{aligned}$%
}
\label{eq:scaled_elbo_1}
\end{equation}
Replacing $\int q_j(\hat{ \vz}_j| \vx) \text{d}\hat{ \vz}_j$ as $1$, then we have:
\begin{equation}
\scalebox{0.85}{%
$\begin{aligned}
       \log p_{\theta}( \vx) \geq 
       % \frac{1}{k}\sum_{i=1}^k \Bigg [ \int q_i(\hat{ \vz}_i| \vx) \log p_\theta( \vx|\hat{ \vz}_i) \mathsf{d}\hat{ \vz}_i \prod_{j=1,\, j\neq i}^k \int q_j(\hat{ \vz}_j| \vx) \text{d}\hat{ \vz}_j \Bigg ] \\
    % &\quad - \int \prod_{i=1}^k q_i(\hat{ \vz}_i |  \vx) \log \frac{\prod_{i=1}^k q_i(\hat{ \vz}_i| \vx)}{\prod_{i=1}^k p(\hat{ \vz}_i)} \mathsf{d}\hat{ \vz}^\prime \\
    % & = \frac{1}{k} \sum_{i=1}^k  \E_{q(\hat{ \vz}_i| \vx )} \log p_\theta( \vx|\hat{ \vz}_i) \\
    % &\quad - \sum_{i=1}^k \bigg [ \KL \left(q_{\phi}(\hat{ \vz}_i| \vx) \,\|\, p(\hat{ \vz}_i)\right) \prod_{j=1,\, j\neq i}^k \int q_j(\hat{ \vz}_j| \vx) \text{d}\hat{ \vz}_j \bigg ] \\
    % & = \frac{1}{k} \sum_{i=1}^k  \E_{q_{\phi} (\hat{ \vz}_i| \vx)} \log p_\theta( \vx|\hat{ \vz}_i) - \sum_{i=1}^k \KL \left(q_{\phi}(\hat{ \vz}_i| \vx) \,\|\, p( \vz_i)\right) \\
    \frac{1}{k} \Bigg[ \sum_{i=1}^k \E_{q_{\phi} (\hat{ \vz}_i| \vx)} \log p_\theta( \vx|\hat{ \vz}_i) - k\, \KL \left(q_{\phi}(\hat{ \vz}_i| \vx) \,\|\, p(\hat{ \vz}_i)\right) \Bigg]
\end{aligned}$%
}
\label{eq:scaled_elbo}
\end{equation}
%However, our model can not evaluate $q(z_1, z_2|x)$ because, latent $z$s are sampled from independent distributions and different deterministic encoders.
%Therefore, $p(z_1, z_2) = p(z_1)p(z_2)$ and we assume $q(z_1, z_2|x) = q_1(z_1|x)q_2(z_2|x)$, finally our proposed method objective function following:
Therefore, we define ELBO as:
\begin{align}
% \small
    \begin{split}
        \mathcal{L}^\prime(\phi, \theta, \psi_{i \in [1, k]};  \vx) & = \frac{1}{k} \sum_{i=1}^k  \E_{q_{\phi,\psi_i} ( \vz_i| \vx)}  \log p_\theta( \vx|\psi_i( \vz))- \\ 
        & \sum_{i=1}^k  \KL (q_{\phi, \psi_i}( \vz| \vx) || p_{\psi_i}( \vz)).
        %& = \underbrace{\frac{1}{k} \sum_{i=1}^k \mathbb{E}_{q_{\phi,\psi_i} (\mathbf{z}_i|\mathbf{x})}  \log p_\theta(x|\psi_1(\mathbf{z}), \psi_2(\mathbf{z}) \cdots, \psi_k(\mathbf{z}))}_{\circled{1}~ \big{\text{reconstruction loss}}} - \underbrace{\sum_{i=1}^k \mathcal{D}_{KL} (q_{\phi, \psi_i}(\mathbf{z}|\mathbf{x}) || p_{\psi_i}(\mathbf{z}))}_{\circled{2}~ \big{\text{KL divergence}}}.
    \end{split}
    \label{appendix: eq: elbo}
\end{align}
However, following Eq.~\ref{appendix: eq: elbo}, k samples are generated, and each sample is disentangled for different factors.
We implement the output as the average of the sum of the k samples to obtain a single sample with a superposition effect from k samples.
%Because we assume $q_{\phi}(\mathbf{z}_1, \mathbf{z}_2, \cdots, \mathbf{z}_k|\mathbf{x})$ and $p_{\theta}(\mathbf{z}_1, \mathbf{z}_2, \cdots, \mathbf{z}_k|\mathbf{x})$ are conditional independence, where $q_{\phi}(\mathbf{x}|\mathbf{z}_1, \mathbf{z}_2, \cdots, \mathbf{z}_k) = q_{\phi}(\mathbf{x}|\mathbf{z}_i)$ and $p_{\theta}(\mathbf{x}|\mathbf{z}_1, \mathbf{z}_2, \cdots, \mathbf{z}_k) = p_{\theta}(\mathbf{x}|\mathbf{z}_i)$ then, $p_{\theta}(\mathbf{z}_1, \mathbf{z}_2, \cdots, \mathbf{z}_k|\mathbf{x}) = \prod_{i=i}^k p_{\theta}(\mathbf{z}_k|\mathbf{x})$.
Moreover, the KL divergence term in Eq.~\ref{appendix: eq: elbo} represents that increasing number of MIPE-transformation is equal to increasing $\beta$ hyper-parameter in $\beta$-VAE~\citep{betaVAE}.

%, and $\text{d} \hat{ \vz} = \text{d}\hat{ \vz}_1 \text{d}\hat{ \vz}_2 \cdots \text{d}\hat{ \vz}_k$.
%However, according to the following Eq.~\ref{eq: reconstrunction error}, $k$ samples are generated, and each sample is disentangled for different factors.
%We implement the output as the average of the sum of the $k$ samples to obtain a single sample with a superposition effect of disentanglement from $k$ samples, as shown in Fig.~\ref{fig:overview}.
%More derivation details of Eq.~\ref{eq: reconstrunction error} are in Appendix~\ref{appendix: elbo}.

The VAEs equipped with MIPE-transformation (MIPET-VAEs) can be trained with the following loss:
\begin{equation}
%\footnotesize
\begin{split}
    %\scriptsize
    &\mathcal{L}(\phi, \theta, \psi_{i \in [1, k ]};  \vx) = \mathcal{L}_{rec} - \mathcal{L}_{kl}
    - \mathcal{L}_{el} -\mathcal{L}_{cali}.
     %\underbrace{\frac{1}{k} \sum_{i=1}^k   \E_{q_{\phi,\psi_i} ( \vz| \vx)}  \log p_\theta( \vx|\psi_i( \vz))}_{\circled{1}~ \big{\text{reconstruction loss}}} \\
     %&~~~~~~~~~~~~~~~~~~~~~~~~~~~~~~~~~~~~~~~~~~~~~~~~~~~~~~~ 
    %-\underbrace{\sum_{i=1}^k  \KL (q_{\phi, \psi_i}( \vz| \vx) || p(\psi_i( \vz)))}_{\circled{2}~ \big{\text{KL divergence}}} - \underbrace{\sum_{i=1}^k \mathcal{L}_{c}(\hat{ \vz}_i, \hat{\boldsymbol{\epsilon}}_i) }_{ \circled{3} ~\big{\text{conversion loss}}},~\text{where} \\
    %&\mathcal{L}_{c} (\hat{ \vz}_m, \hat{\boldsymbol{\epsilon}}_m) = \underbrace{|| \nabla_{\hat{ \vz}_m, \hat{\boldsymbol{\epsilon}}_m, \boldsymbol{\lambda}_m} \mathcal{L}_{s} ||_2^2}_{\tiny \circled{3.1} ~\BBig{\text{EF similarity loss}}} + \underbrace{\text{MSE}(  \KL(q_\phi( \vz| \vx)||p_\theta( \vz)),    \KL(f_{ \vx}( \vx|\vtheta_{\hat{ \vz}_m}) || f_{ \vx}( \vx|\boldsymbol{\theta}_{\hat{\boldsymbol{\epsilon}}_m})))}_{\tiny \circled{3.2} ~\BBig{\text{KL divergence calibration loss}}}.
\end{split}
\label{appendix eq:objective_function}
% \vspace{-30pt}
\end{equation}

\section{Details of Experimental Environment}
\label{appendix: details of experimentla environment}
\subsection{Model Architecture and Hyper-Parameter Tuning}
The details of model architecture is demonstrated in Table~\ref{tab:dsprites architecure}and~\ref{tab:others architecure}.
The details of hyper-parameter settings are in Table~\ref{tab:hyper-parameter setting}.
\begin{table}[ht]
    \caption{VAE architecture for dSprites dataset.}
    \centering
    \scriptsize
    \begin{tabular}{c|c}
    \hline
         Encoder & Decoder  \\
         \hline
         Input $64\times64$ binary image & input $\in \mathbb{R}^{10}$ \\
         \hline
         $4 \times 4$ conv. 32 ReLU. stride 2 & FC. 128 ReLU. \\
         \hline
         $4 \times 4$ conv. 32 ReLU. stride 2 & FC. $4 \times 4 \times 64$ ReLU. 
         \\
         \hline
         $4 \times 4$ conv. 64 ReLU. stride 2 & $4 \times 4$ upconv. 64 ReLU. stride 2. \\
         \hline
         $4 \times 4$ conv. 64 ReLU. stride 2 & $4 \times 4$ upconv. 32 ReLU. stride 2. \\
         \hline
         FC. 128. FC. $2 \times 10$ & $4 \times 4$ upconv. 32 ReLU. stride 2. \\
         \hline
         & $4 \times 4$ upconv. 1. stride 2 \\
         \hline
    \end{tabular}
    \label{tab:dsprites architecure}
    \vspace{3mm}
    \caption{VAE architecture for 3D Shapes, and 3D Cars datasets. For exceptional case, CLG-VAE, we use ten dimension size on 3D Shapes dataset~\citep{commutative-vae}.}
    \centering
    \scriptsize
    \begin{tabular}{c|c}
    \hline
         Encoder & Decoder  \\
         \hline
         Input $64\times64 \times 3$ RGB image & input $\in \mathbb{R}^{6}$ (3D Shapes), $\mathbb{R}^{10}$ (3D Cars) \\
         \hline
         $4 \times 4$ conv. 32 ReLU. stride 2 & FC. 256 ReLU. \\
         \hline
         $4 \times 4$ conv. 32 ReLU. stride 2 & FC. $4 \times 4 \times 64$ ReLU. 
         \\
         \hline
         $4 \times 4$ conv. 64 ReLU. stride 2 & $4 \times 4$ upconv. 64 ReLU. stride 2. \\
         \hline
         $4 \times 4$ conv. 64 ReLU. stride 2 & $4 \times 4$ upconv. 32 ReLU. stride 2. \\
         \hline
         FC. 256. FC. $2 \times 10$ & $4 \times 4$ upconv. 32 ReLU. stride 2. \\
         \hline
         & $4 \times 4$ upconv. 3. stride 2 \\
         \hline
    \end{tabular}
    
    \label{tab:others architecure}
    % \vspace{-20pt}
\end{table}

\begin{table}[ht]
\caption{Hyper-parameters for dSprites, 3D Cars, and 3D Shapes. The epochs for dSprites and 3D cars are 30 and 200, respectively. lr is learning rate, latent dim is latent vector size, group reconst is group reconstruction, and forward group is forward group pass.}
\centering
\scriptsize
\subfloat[Hyper-parameters for dSprites and 3D Cars]{
    \begin{tabular}{c|c|c}
        \hline
        models & hyper-parameters & values \\
        \hline \hline
        \multirow{7}{*}{common}& batch size & 256 \\
        & epoch & $\{30, 200\}$ \\
        & optim & Adam \\
        & lr & 4e-4 \\
        & lr for MIPET & 4e-4 \\
        & weight decay & 1e-4 \\
        & latent dim & 10 \\
        \hline
        $\beta$-VAE & $\#$ of IE and EF & $\{1, 2, 4, 10\}$ \\
        \hline
        \multirow{3}{*}{$\beta$-TCVAE} & $\beta$ & $\{4, 6\}$ \\
        & $\#$ of IE and EF & $\{1, 3\}$ \\
        & $\alpha$, $\gamma$ & 1.0 \\
        \hline
        \multirow{4}{*}{CLG-VAE}& $\lambda_{\text{decomp}}$ & 40 \\
        & $\lambda_{\text{hessian}}$ & 40 \\
        & forward group & 0.2 \\
        & group reconst & $\{0.2, 0.5, 0.7\}$ \\
        \hline
    \end{tabular}
    \label{tab:dsprites_3dcars}
}
\hfill
\subfloat[Hyper-parameters for 3D Shapes]{
    \begin{tabular}{c|c|c}
        \hline
        models & hyper-parameters & values \\
        \hline \hline
        \multirow{5}{*}{common}& batch size & 256 \\
        & epoch & 67 \\
        & optim & Adam \\
        & lr & 4e-4 \\
        & lr for MIPET & 4e-4 \\
        \hline
        \multirow{3}{*}{$\beta$-VAE} & $\#$ of IE and EF & $\{1, 2, 4, 10\}$ \\
        & weight decay & 0.0 \\
        & latent dim & 6 \\
        \hline
        \multirow{5}{*}{$\beta$-TCVAE} & $\beta$& $\{4, 6\}$ \\
        & $\#$ of IE and EF & $\{1, 3\}$ \\
        & $\alpha$, $\gamma$ & 1.0 \\
        & weight decay & 1e-4 \\
        & latent dim & 6 \\
        \hline
        \multirow{6}{*}{CLG-VAE}& $\lambda_{\text{decomp}}$ & 40 \\
        & $\lambda_{\text{hessian}}$ & 40 \\
        & forward group & 0.2 \\
        & group reconst & $\{0.2, 0.5, 0.7\}$ \\
        & weight decay & 0.0 \\
        & latent dim & 10 \\
        \hline
    \end{tabular}
    \label{tab:3dshapes}
}
\label{tab:hyper-parameter setting}
\end{table}

\subsection{Details of Datasets}
\label{appendix: datasets}
We benchmark MIPET-VAEs against several standard VAEs on the following datasets: 1) dSprites~\citep{dsprites17} which consists of 737,280 binary $64 \times 64$ images of dSprites with five independent ground truth factors(number of values), $i.e.$ shape(3), orientation(40), scale(6), x-position(32), and y-position(32). 
2) 3D Shapes~\citep{3dshapes18} which consists of 480,000 RGB $64 \times 64 \times 3$ images of 3D Shapes with six independent ground truth factors: shape(4) orientation(15), scale(8), wall color(10), floor color(10), and object color(10).
3) 3D Cars~\citep{3d-car-dataset} which consists of 17,568 RGB $64 \times 64 \times 3$ images of 3D Shapes with three independent ground truth factors: car models(183), azimuth directions(24), and elevations(4).

\subsection{Details of Training and Evaluation}
\label{appendix: hyper-tuning}
\paragraph{Training} We set 256 mini-batch size in the datasets (dSprites, 3D Shapes, and 3D Cars), Adam optimizer with learning rate $4 \times 10^{-4}$, $\beta_1=0.9$, $\beta_2=0.999$, and epochs from $\{30, 67, 200\}$ as a common setting for all the comparative methods.
For the comparison, we follow training and inference on the whole dataset.
We train each model for 30, 67, and 200 epochs on the dSprites, 3D Shapes, and 3D Cars, respectively, as introduced in~\citet{factor-vae, 3d-car-setting}. 
We tune $\beta$ from $\{1, 2, 4, 10\}$ and $\{4, 6 \}$ for $\beta$-VAE and $\beta$-TCVAE, respectively.
We apply the proposed method to $\beta$-TCVAE only with the EF similarity loss term because $\beta$-TCVAE penalizes the divided KL divergence terms.
We set the same encoder and decoder architecture in each model to exclude the overlapped effects.
The $\alpha=1$ and $\gamma=1$ of $\beta$-TCVAE as~\citet{beta-tcvae}.
We set the dimension size of the latent vectors from $\{6, 10\}$ for 10 on dSprites and 3D Cars datasets and 6 for 3D Shapes, but we set 10 for CLG-VAE because it sets 10 dimensions size on 3D Shapes in~\citet{commutative-vae}.
Regarding the CLG-VAE, we fix $\lambda_{\text{decomp}}$, $\lambda_{\text{hessian}}$, and forward group features as 40, 20, and 0.2, respectively.
Because the hyper-parameters showed the best result in~\citet{commutative-vae}.
We set group reconstruction from $ \{0.2, 0.5, 0.7\}$.
For Control-VAE, we set the maximum KL divergence value from $\{10, 12, \ldots, 20 \}$.
In addition, we set masking ratio $\lambda$ from $\{0.0, 0.5, \ldots 2.0, \infty \}$.
To check the impact of MIPE-transformation, we do not consider the Groupified VAE because the latter is implemented with an extended decoder (different capacity).

\paragraph{Evaluation} 
We conduct experiments on  NVIDIA A100, RTX 2080 Ti, and RTX 3090. We set 100 samples to evaluate global empirical variance in each dimension and run it a total of 800 times to estimate the FVM score introduced in~\citet{factor-vae}.
For the MIG~\citep{beta-tcvae}, SAP~\citep{sap}, and DCI~\citep{dci}, we follow default values introduced in~\citet{Michlo2021Disent}, training and evaluation 100 and 50 times with 100 mini-batches, respectively.
We evaluate four disentanglement metrics for a less biased understanding of the actual states of disentanglement.

\begin{table*}[t]
    \caption{Impact of the number of MIPE-transformation function on the $\beta$-TCVAE and $\beta$-VAE with dSprites, 3D Shapes, and 3D Cars datasets in terms of the four metrics.
    The blue and red box plots represent each model's single and multiple IPE-transformation cases, respectively. (A-$n$: MIPET-$\beta$-TCVAE (4), B-$n$: MIPET-$\beta$-TCVAE (6), C-$n$: MIPET-$\beta$-VAE, $n$: the number of MIPE-transformation).}
    \centering
    \begin{tabular}{c|>{\centering\arraybackslash}m{30mm}|>{\centering\arraybackslash}m{30mm}|>{\centering\arraybackslash}m{30mm}|>{\centering\arraybackslash}m{30mm}}
        \hline
         \multirow{2}{*}{dataset} & \multicolumn{4}{c}{Metrics} \\
         \cline{2-5}
         & FVM & MIG & SAP & DCI \\
         \hline
         % dSprites & 
         % \includegraphics[trim=0 0 0 -5, width=0.18\textwidth]{figure_dSprites_FVM.jpg} & \includegraphics[width=0.18\textwidth]{figure_dSprites_MIG.jpg} & \includegraphics[width=0.18\textwidth]{figure_dSprites_SAP.jpg} & \includegraphics[width=0.18\textwidth]{figure_dSprites_DCI.jpg}\\
         % \hline
         3D Shapes & 
         \includegraphics[trim=0 0 0 -5, width=0.18\textwidth]{figure_3Dshape_FVM.jpg} &
         \includegraphics[width=0.18\textwidth]{figure_3Dshape_MIG.jpg} &
         \includegraphics[width=0.18\textwidth]{figure_3Dshape_SAP.jpg} &
         \includegraphics[width=0.18\textwidth]{figure_3Dshape_DCI.jpg}\\
         \hline
         3D Cars & 
         \includegraphics[trim=0 0 0 -5, width=0.18\textwidth]{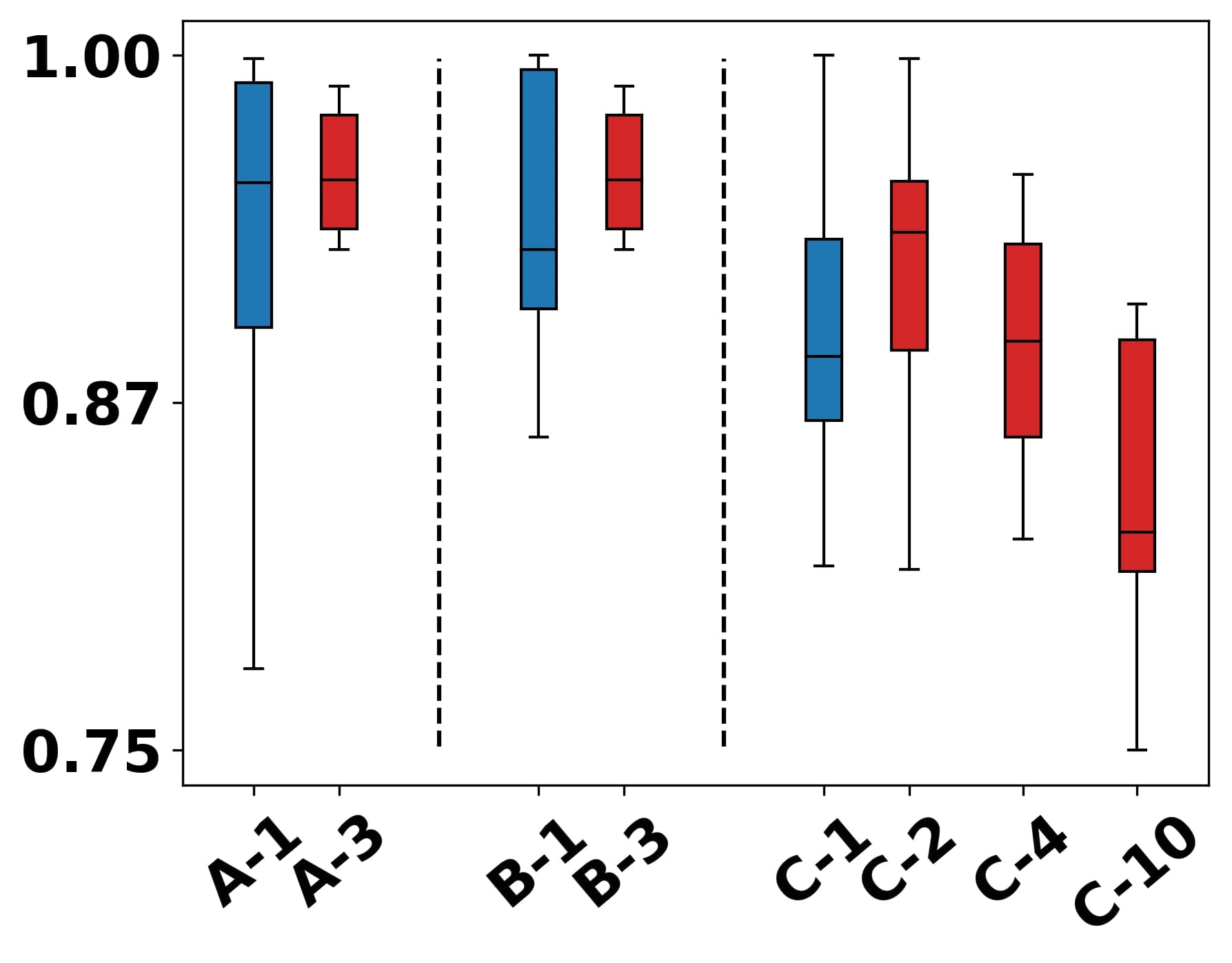} &
         \includegraphics[width=0.18\textwidth]{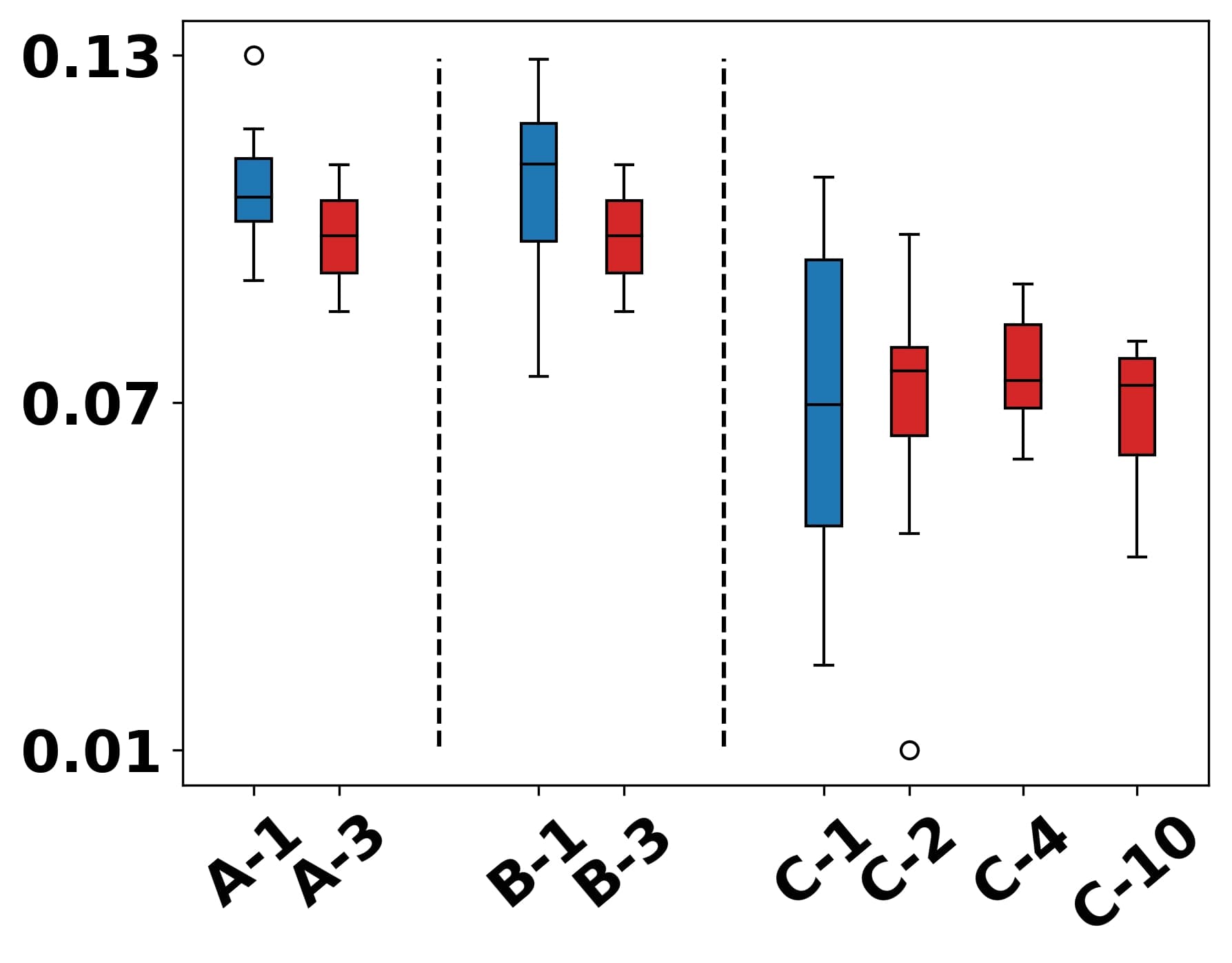} &
         \includegraphics[width=0.18\textwidth]{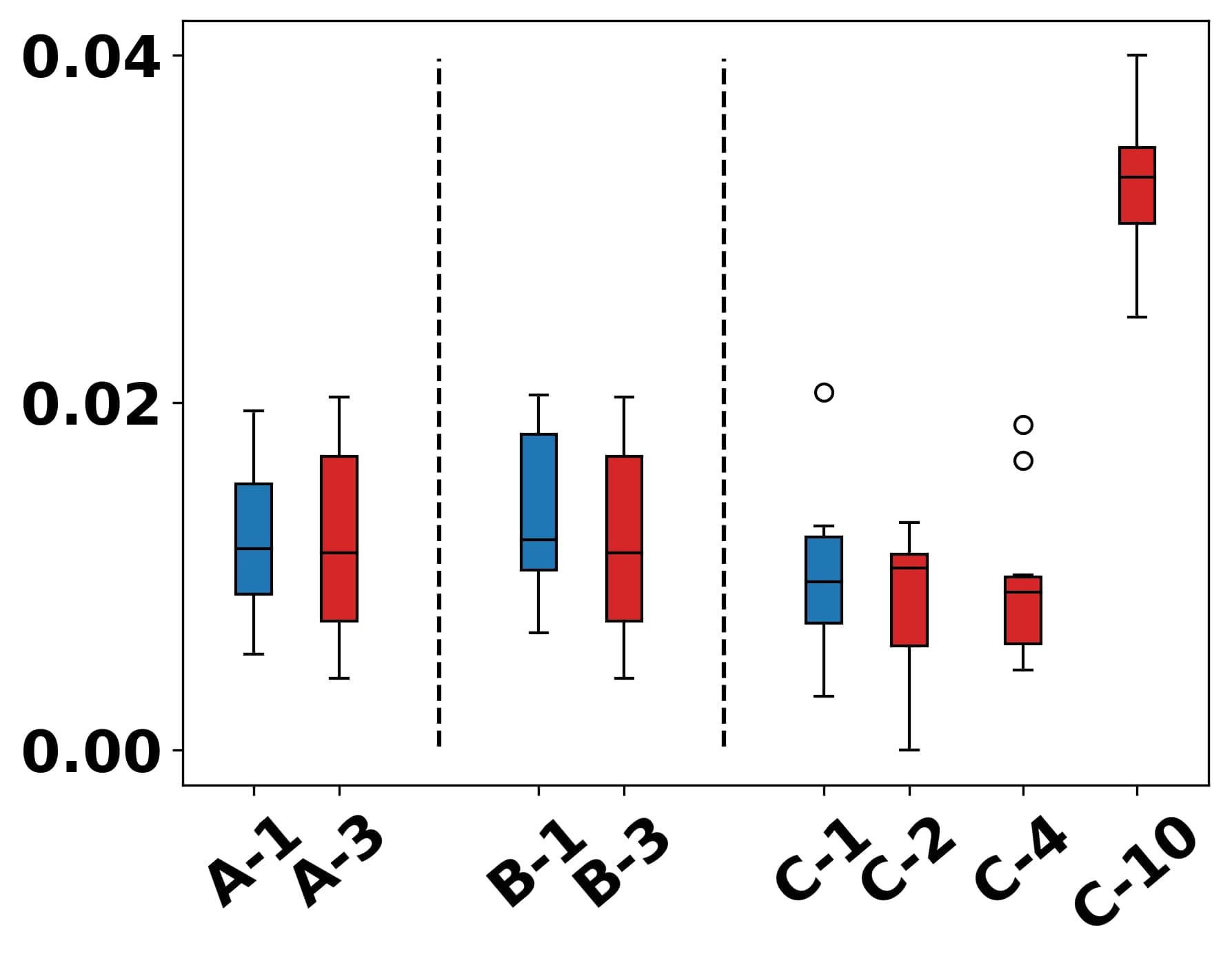} &
         \includegraphics[width=0.18\textwidth]{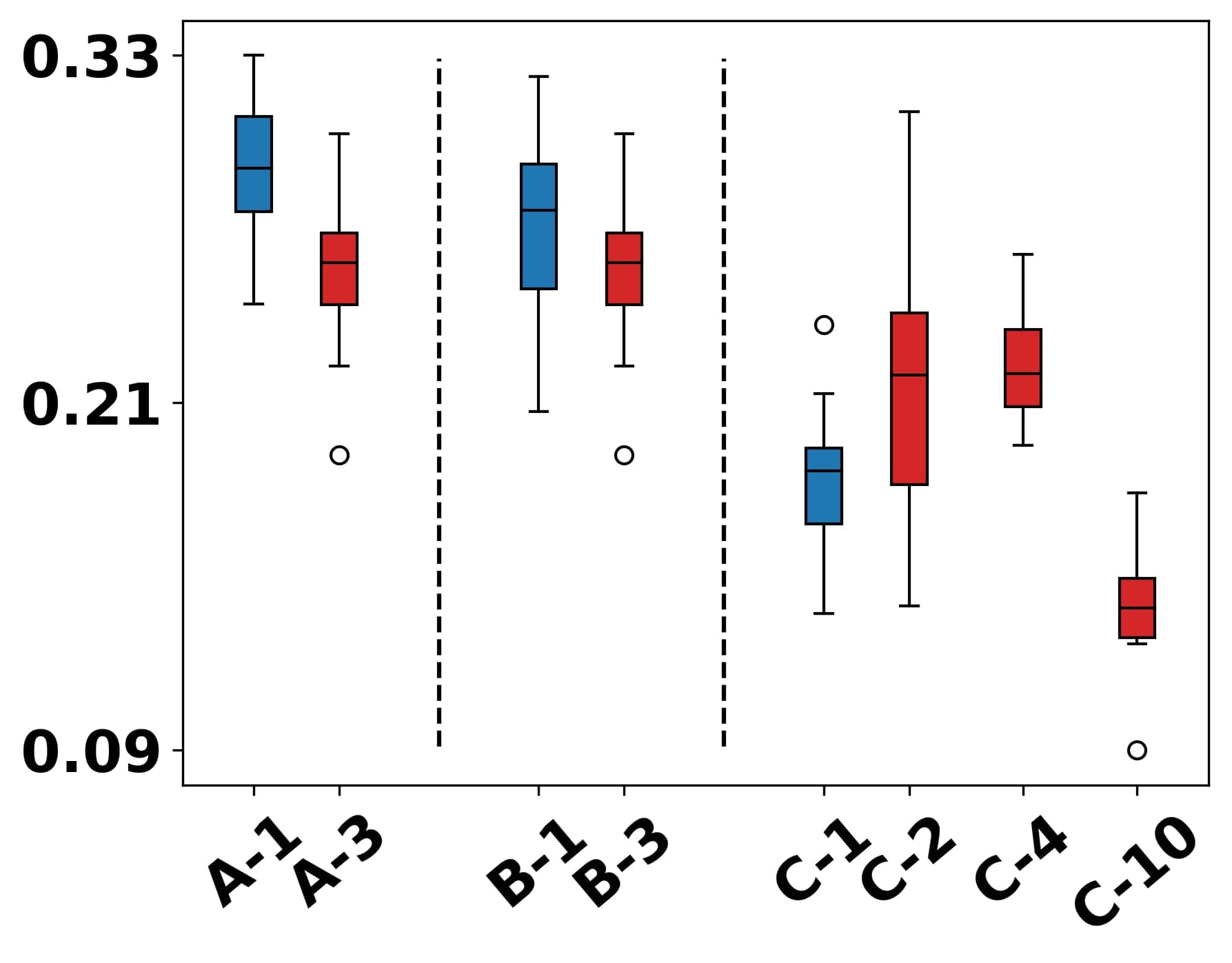}\\
         \hline 
    \end{tabular}
    \label{appendix table :main impact of IE}
\end{table*}

\subsection{Additional Result of Relation Between Factors and Latent Vector Dimensions}
\label{appendix: dci matrix}

% \begin{figure}[ht]
%     \centering
%     \includegraphics[width=0.48\textwidth]{figure_shapes3d_dci_matrix.jpg}
%     \caption{
%     Each square represents a value in the DCI matrix, which describes the relationship between the $i^{th}$ latent dimension and each factor.
%     The size of each square is relative to the values within each row.
%     The ideal case resembles a sparse matrix.
%     The y-axis represents the factors of each dataset, while the x-axis corresponds to the latent vector dimensions.
%     The number shown in each row of the matrix indicates the maximum value and standard deviation of that row.
%     Higher maximum and standard deviation values suggest greater sparsity, indicating closer alignment with the ideal case.
%     More details in Appendix~\ref{appendix: dci matrix}.}
%     \label{figure: dci_matrix}
% % \vspace{-20pt}
% \end{figure}

\begin{figure}[ht]
    \centering
    \includegraphics[width=0.48\textwidth]{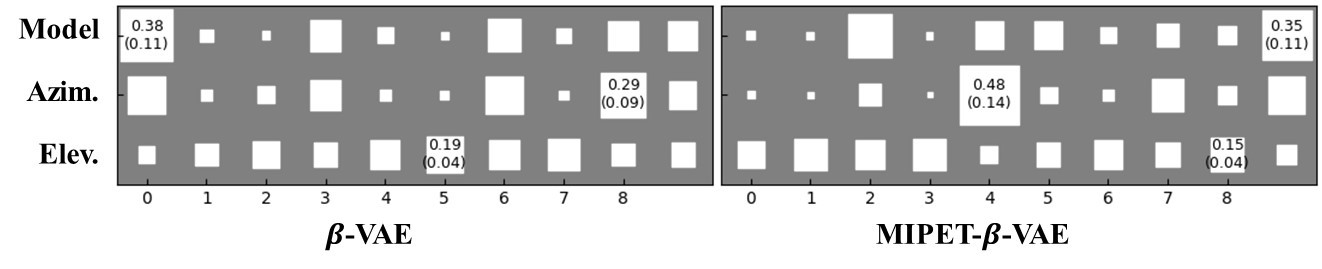}
    \caption{3D Car Dataset: Azim. refers to the azimuth factor, and Elev. refers to the elevation factor.}
    \label{fig:enter-label}
\end{figure}

% \paragraph{Relation Between Factors and Latent Vector Dimensions.}
% As shown in Fig.~\ref{figure: dci_matrix}, our model shows a close to sparse matrix compared to the $\beta$-VAE result in 3D Shapes, also the maximum value and standard deviation of each row are larger than $\beta$-VAE.
% This indicates that our model achieves better alignment of individual factors with specific latent dimensions compared to other methods.

In the 3D Car dataset, a trade-off was observed. While the maximum value across models and evaluation factors showed a slight decrease, the azimuth factor exhibited a notable improvement compared to the declines seen in other factors.

\subsection{Impact of Symmetric Matrix Exponential}
\label{subsec: discussion}

%\begin{table}[]
%    \caption{The ratio of seeds to show better performance with symmetric matrix
%    }
%    \vskip 0.15in
%    \centering
%    \small
%    \begin{tabular}{|c|c|c|}
%    \hline
%         dSprites & 3D Shapes & 3D Cars \\
%         \hline
%          0.58 & 0.56 & 0.67 \\
%         \hline
%    \end{tabular}

%    \label{tab: main empirical symmetric}
%    \vskip -0.15in
%\end{table}
%%%%%%%%%%%%%%%%%%%%%%%%%%%%%%%%%%%%%%%%%%%%%%%%%%%%%%%%%%
\begin{table}[h]
% \tiny
    \centering
    \caption{The ratio of seeds to show better performance with symmetric matrix}
    \begin{tabular}{|c|c|c|}
    \hline
         dSprites & 3D Shapes & 3D Cars \\
         \hline
          0.58 & 0.56 & 0.67 \\
         \hline
    \end{tabular}
    \label{tab: main empirical symmetric}
\end{table}
%%%%%%%%%%%%%%%%%%%%%%%%%%%%%%%%%%%%%%%%%%%%%%%%%%%%%%%%%%
%In this section, we show the empirical result of discussion on~\ref{discussion: why symmetric?}. 
We empirically show the benefit of using a symmetric matrix for $\psi$. Table~\ref{tab: main empirical symmetric} shows the ratio of runs with a symmetric matrix, which shows better performance than unrestricted matrices, to the total 240 (60 models $\times$ 4 metrics) runs for each dataset. 
%We estimate the ratio as $\eps = \frac{ \text{\# of cases that symmetric matrix shows better performance}}{ \text{\# of total case}}$ with 240 cases on each dataset.
%We count cases when the proposed method result is higher than the asymmetric then divide it by the total case 240 (60 models $\times$ 4 metrics).
%The results are presented in Table~\ref{tab: main empirical symmetric}.
All results are higher than 0.5, which implies that the constraint enhances I2L equivariance even with uncertain factors. 
%These results imply that our proposed method finds equivariant transformation function over latent space, in which group element is potentially expressed by subgroup $G_1^\prime$ better than the asymmetric cases.

\subsection{Additional Experiment of Computing Complexity}
\label{appendix: complexity}
\begin{table}[h]
    % \tiny
    \caption{Training complexity.}
    \centering
    \begin{tabular}{|c|c|}
    \hline
         $\#$ of IE & Complexity \\
         \hline
         0 & $\times$ 1.00 \\
         1 & $\times$ 0.75 \\
         3 & $\times$ 0.50 \\
         4 & $\times$ 0.33 \\
         \hline
    \end{tabular}

    \label{tab: complexity}
\end{table}

We additionally estimate the computing complexity depending on the number of IPE-transformation.
The results are in Table~\ref{tab: complexity} and represent the training time complexity compare to baselines (when the number of IE is equal to 0).

% \clearpage
\subsection{Semantic Mask: Mapping Gaussian Distribution to Diverse Exponential Family}
\begin{figure}[!t]
    \centering
    % \subfloat[2D Gamma Distribution]{\includegraphics[width=0.99\linewidth]{figure_2d_gamma.jpg}\label{fig:gamma}}\\
    % \subfloat[2D Dirichlet Distribution]
    \includegraphics[width=0.99\linewidth]{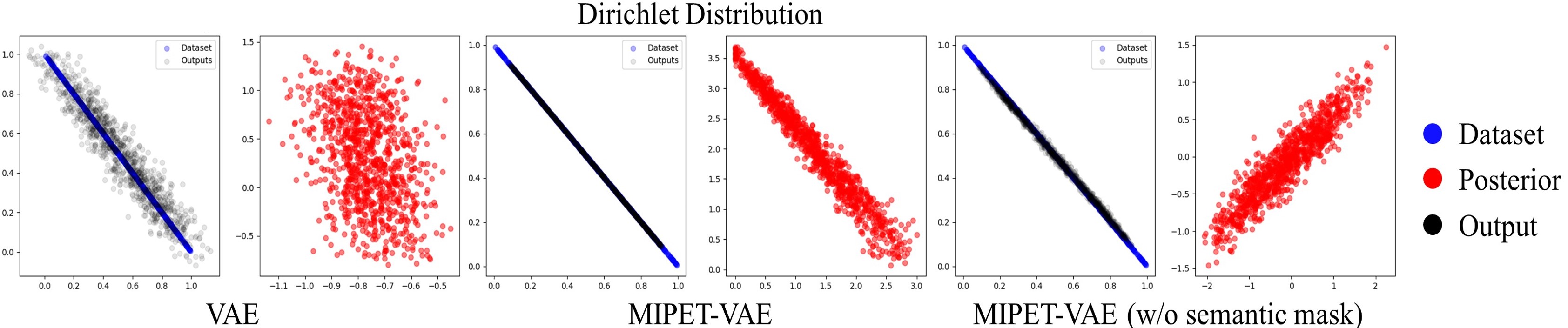}
    \caption{2D Exponential Family Distribution Datasets}
    \label{fig:2d_datasets}
\end{figure}

As shown in Fig.~\ref{fig:2d_datasets}, MIPET-VAE with semantic mask preserves the dataset structure compared to VAE. 
Also, EF conversion with semantic mask maps the Gaussian distribution to the exponential family compared to MIPET without the semantic mask and VAE cases.

\label{appendix: ef}
% \begin{figure}
%     \centering
%     \begin{subfigure}{0.99\textwidth}
%     \includegraphics[width=1.0\textwidth]{figure_2d_gamma.jpg}
%     \caption{2d gamma distribution dataset.}
%     \end{subfigure}
%     \vfill
%     \begin{subfigure}{0.99\textwidth}
%     \includegraphics[width=1.0\textwidth]{figure_2d_dirichlet.jpg}
%     \caption{2d Dirichlet distribution dataset.}
%     \end{subfigure}
%     \caption{2d exponential family distribution dataset.}
%     \label{appendix figure: 2d toy}
% \end{figure}

\end{document}

%% file: math_commands.tex
\usepackage{amsmath,amsfonts,bm}

\def\eqref#1{equation~\ref{#1}}
\def\1{\bm{1}}

\def\vx{{\bm{x}}}

\def\vz{{\bm{z}}}

\def\mM{{\bm{M}}}

\def\mS{{\bm{S}}}

\DeclareMathAlphabet{\mathsfit}{\encodingdefault}{\sfdefault}{m}{sl}
\SetMathAlphabet{\mathsfit}{bold}{\encodingdefault}{\sfdefault}{bx}{n}

\newcommand{\E}{\mathbb{E}}

\newcommand{\KL}{D_{\mathrm{KL}}}

%% file: main.bib
@inproceedings{
intel-vae,
title={On Incorporating Inductive Biases into {VAE}s},
author={Ning Miao and Emile Mathieu and Siddharth N and Yee Whye Teh and Tom Rainforth},
booktitle={International Conference on Learning Representations},
year={2022},
}

@InProceedings{commutative-vae,
  title = 	 {Commutative Lie Group VAE for Disentanglement Learning},
  author =       {Zhu, Xinqi and Xu, Chang and Tao, Dacheng},
  booktitle = 	 {Proceedings of the 38th International Conference on Machine Learning},
  pages = 	 {12924--12934},
  year = 	 {2021},
  editor = 	 {Meila, Marina and Zhang, Tong},
  volume = 	 {139},
  series = 	 {Proceedings of Machine Learning Research},
  month = 	 {18--24 Jul},
  publisher =    {PMLR}
}

@misc{vae,
  doi = {10.48550/ARXIV.1312.6114},
  
  author = {Kingma, Diederik P and Welling, Max},
  
  title = {Auto-Encoding Variational Bayes},
  
  publisher = {arXiv},
  
  year = {2013},
  
  copyright = {arXiv.org perpetual, non-exclusive license}
}

@inproceedings{
groupified-vae,
title={Towards Building A Group-based Unsupervised Representation Disentanglement Framework},
author={Tao Yang and Xuanchi Ren and Yuwang Wang and Wenjun Zeng and Nanning Zheng},
booktitle={International Conference on Learning Representations},
year={2022},
}

@article{cfasl,
title={{CFASL}: Composite Factor-Aligned Symmetry Learning for Disentanglement in Variational AutoEncoder},
author={Hee-Jun Jung and Jaehyoung Jeong and Kangil Kim},
journal={Transactions on Machine Learning Research},
issn={2835-8856},
year={2024},
url={https://openreview.net/forum?id=mDGvrH7lju},
note={}
}

@InProceedings{homomorphicVAE,
  title = 	 {Homomorphism {A}uto{E}ncoder -- Learning Group Structured Representations from Observed Transitions},
  author =       {Keurti, Hamza and Pan, Hsiao-Ru and Besserve, Michel and Grewe, Benjamin F and Sch\"{o}lkopf, Bernhard},
  booktitle = 	 {Proceedings of the 40th International Conference on Machine Learning},
  pages = 	 {16190--16215},
  year = 	 {2023},
  editor = 	 {Krause, Andreas and Brunskill, Emma and Cho, Kyunghyun and Engelhardt, Barbara and Sabato, Sivan and Scarlett, Jonathan},
  volume = 	 {202},
  series = 	 {Proceedings of Machine Learning Research},
  month = 	 {23--29 Jul},
  publisher =    {PMLR},
  url = 	 {https://proceedings.mlr.press/v202/keurti23a.html}
}

@inproceedings{Locatello2020WeaklySupervisedDW,
  title={Weakly-Supervised Disentanglement Without Compromises},
  author={Francesco Locatello and Ben Poole and Gunnar R{\"a}tsch and Bernhard Scholkopf and Olivier Bachem and Michael Tschannen},
  booktitle={International Conference on Machine Learning},
  year={2020},
  url={https://api.semanticscholar.org/CorpusID:211066424}
}

@inproceedings{
group-invariant-equivariant,
title={Unsupervised Learning of Group Invariant and Equivariant Representations},
author={Robin Winter and Marco Bertolini and Tuan Le and Frank Noe and Djork-Arn{\'e} Clevert},
booktitle={Advances in Neural Information Processing Systems},
editor={Alice H. Oh and Alekh Agarwal and Danielle Belgrave and Kyunghyun Cho},
year={2022},
url={https://openreview.net/forum?id=47lpv23LDPr}
}

@ARTICLE{concave,
  author={Togo, Ren and Nakagawa, Nao and Ogawa, Takahiro and Haseyama, Miki},
  journal={IEEE Transactions on Neural Networks and Learning Systems}, 
  title={ConcVAE: Conceptual Representation Learning}, 
  year={2025},
  volume={36},
  number={4},
  pages={7529-7541},
  doi={10.1109/TNNLS.2024.3404496}}

@inproceedings{
exp_family_bayesian,
title={Natural Posterior Network: Deep Bayesian Predictive Uncertainty for Exponential Family Distributions},
author={Bertrand Charpentier and Oliver Borchert and Daniel Z{\"u}gner and Simon Geisler and Stephan G{\"u}nnemann},
booktitle={International Conference on Learning Representations},
year={2022},
}

@book{prml,
author = {Bishop, Christopher M.},
title = {Pattern Recognition and Machine Learning (Information Science and Statistics)},
year = {2006},
isbn = {0387310738},
publisher = {Springer-Verlag},
address = {Berlin, Heidelberg}
}

@InProceedings{factor-vae,
  title = 	 {Disentangling by Factorising},
  author =       {Kim, Hyunjik and Mnih, Andriy},
  booktitle = 	 {Proceedings of the 35th International Conference on Machine Learning},
  pages = 	 {2649--2658},
  year = 	 {2018},
  editor = 	 {Dy, Jennifer and Krause, Andreas},
  volume = 	 {80},
  series = 	 {Proceedings of Machine Learning Research},
  month = 	 {10--15 Jul},
  publisher =    {PMLR}
}

@inproceedings{beta-tcvae,
 author = {Chen, Ricky T. Q. and Li, Xuechen and Grosse, Roger B and Duvenaud, David K},
 booktitle = {Advances in Neural Information Processing Systems},
 editor = {S. Bengio and H. Wallach and H. Larochelle and K. Grauman and N. Cesa-Bianchi and R. Garnett},
 pages = {},
 publisher = {Curran Associates, Inc.},
 title = {Isolating Sources of Disentanglement in Variational Autoencoders},
 volume = {31},
 year = {2018}
}

@inproceedings{
sap,
title={{VARIATIONAL} {INFERENCE} {OF} {DISENTANGLED} {LATENT} {CONCEPTS} {FROM} {UNLABELED} {OBSERVATIONS}},
author={Abhishek Kumar and Prasanna Sattigeri and Avinash Balakrishnan},
booktitle={International Conference on Learning Representations},
year={2018},
}

@inproceedings{
dci,
title={A framework for the quantitative evaluation of disentangled representations},
author={Cian Eastwood and Christopher K. I. Williams},
booktitle={International Conference on Learning Representations},
year={2018},
}

@misc{dsprites17,
author = {Loic Matthey and Irina Higgins and Demis Hassabis and Alexander Lerchner},
title = {dSprites: Disentanglement testing Sprites dataset},
howpublished= {https://github.com/deepmind/dsprites-dataset/},
year = "2017",
}

@misc{3dshapes18,
  title={3D Shapes Dataset},
  author={Burgess, Chris and Kim, Hyunjik},
  howpublished={https://github.com/deepmind/3dshapes-dataset/},
  year={2018}
}

@inproceedings{zero-shot,
author = {Tenenbaum, Josh},
title = {Building Machines That Learn and Think Like People},
year = {2018},
publisher = {International Foundation for Autonomous Agents and Multiagent Systems},
address = {Richland, SC},
booktitle = {Proceedings of the 17th International Conference on Autonomous Agents and MultiAgent Systems},
pages = {5},
numpages = {1},
location = {Stockholm, Sweden},
series = {AAMAS '18}
}

@InProceedings{domain-1,
author = {Li, Yu-Jhe and Lin, Ci-Siang and Lin, Yan-Bo and Wang, Yu-Chiang Frank},
title = {Cross-Dataset Person Re-Identification via Unsupervised Pose Disentanglement and Adaptation},
booktitle = {Proceedings of the IEEE/CVF International Conference on Computer Vision (ICCV)},
month = {October},
year = {2019}
}

@inproceedings{domain-2,
  title={Joint disentangling and adaptation for cross-domain person re-identification},
  author={Zou, Yang and Yang, Xiaodong and Yu, Zhiding and Vijayakumar, Bhagavatula and Kautz, Jan},
  booktitle={Proceedings of the European Conference on Computer Vision (ECCV)},
  year={2020}
}

@ARTICLE{disen_definitiaon_1,
  author={Bengio, Yoshua and Courville, Aaron and Vincent, Pascal},
  journal={IEEE Transactions on Pattern Analysis and Machine Intelligence}, 
  title={Representation Learning: A Review and New Perspectives}, 
  year={2013},
  volume={35},
  number={8},
  pages={1798-1828},
  doi={10.1109/TPAMI.2013.50}}

@article{disen_definitiaon_2,
  author    = {Irina Higgins and
               David Amos and
               David Pfau and
               S{\'{e}}bastien Racani{\`{e}}re and
               Lo{\"{\i}}c Matthey and
               Danilo J. Rezende and
               Alexander Lerchner},
  title     = {Towards a Definition of Disentangled Representations},
  journal   = {CoRR},
  volume    = {abs/1812.02230},
  year      = {2018},
  eprinttype = {arXiv},
  eprint    = {1812.02230},
  bibsource = {dblp computer science bibliography, https://dblp.org}
}

@inproceedings{betaVAE,
  title={beta-VAE: Learning Basic Visual Concepts with a Constrained Variational Framework},
  author={Irina Higgins and Lo{\"i}c Matthey and Arka Pal and Christopher P. Burgess and Xavier Glorot and Matthew M. Botvinick and Shakir Mohamed and Alexander Lerchner},
  booktitle={ICLR},
  year={2017}
}

@InProceedings{vae-inductive-bias,
  title = 	 {Challenging Common Assumptions in the Unsupervised Learning of Disentangled Representations},
  author =       {Locatello, Francesco and Bauer, Stefan and Lucic, Mario and Raetsch, Gunnar and Gelly, Sylvain and Sch{\"o}lkopf, Bernhard and Bachem, Olivier},
  booktitle = 	 {Proceedings of the 36th International Conference on Machine Learning},
  pages = 	 {4114--4124},
  year = 	 {2019},
  editor = 	 {Chaudhuri, Kamalika and Salakhutdinov, Ruslan},
  volume = 	 {97},
  series = 	 {Proceedings of Machine Learning Research},
  month = 	 {09--15 Jun},
  publisher =    {PMLR}
}

@inproceedings{classi_1,
  title     = {Understanding Failures of Deep Networks via Robust Feature Extraction},
  author    = {Sahil Singla and Besmira Nushi and Shital Shah and Ece Kamar and Eric Horvitz},
  booktitle = {{IEEE} Conference on Computer Vision and Pattern Recognition, {CVPR} 2021},
  publisher = {Computer Vision Foundation / {IEEE}},
  year      = {2021},}

@article{t-vae,
  author    = {T. Anderson Keller and
               Max Welling},
  title     = {Topographic VAEs learn Equivariant Capsules},
  journal   = {CoRR},
  volume    = {abs/2109.01394},
  year      = {2021},
  eprinttype = {arXiv},
  eprint    = {2109.01394},
  bibsource = {dblp computer science bibliography, https://dblp.org}
}

@article{broadcast-decoder,
  author    = {Nicholas Watters and
               Lo{\"{\i}}c Matthey and
               Christopher P. Burgess and
               Alexander Lerchner},
  title     = {Spatial Broadcast Decoder: {A} Simple Architecture for Learning Disentangled
               Representations in VAEs},
  journal   = {CoRR},
  volume    = {abs/1901.07017},
  year      = {2019},
  eprinttype = {arXiv},
  eprint    = {1901.07017},
  bibsource = {dblp computer science bibliography, https://dblp.org}
}

@inproceedings{3d-car-dataset,
 author = {Reed, Scott E and Zhang, Yi and Zhang, Yuting and Lee, Honglak},
 booktitle = {Advances in Neural Information Processing Systems},
 editor = {C. Cortes and N. Lawrence and D. Lee and M. Sugiyama and R. Garnett},
 pages = {},
 publisher = {Curran Associates, Inc.},
 title = {Deep Visual Analogy-Making},
 volume = {28},
 year = {2015}
}

@article{hyper-vae,
  author = {Davidson, Tim R. and Falorsi, Luca and De Cao, Nicola and Kipf, Thomas and Tomczak, Jakub M.},
  journal = {34th Conference on Uncertainty in Artificial Intelligence (UAI-18)},
  title = {Hyperspherical Variational Auto-Encoders},
  year = 2018
}

@inproceedings{3d-car-setting,
  title   = {Learning Disentangled Representation by Exploiting Pretrained Generative Models: A Contrastive Learning View},
  author  = {Ren, Xuanchi and Yang, Tao and Wang, Yuwang and Zeng, Wenjun},
  booktitle = {ICLR},
  year    = {2022}
}

@inproceedings{invertible-matrix-exponential,
  title = 	 {Generative Flows with Matrix Exponential},
  author =       {Xiao, Changyi and Liu, Ligang},
  booktitle = 	 {Proceedings of the 37th International Conference on Machine Learning},
  pages = 	 {10452--10461},
  year = 	 {2020},
  editor = 	 {III, Hal Daumé and Singh, Aarti},
  volume = 	 {119},
  series = 	 {Proceedings of Machine Learning Research},
  month = 	 {13--18 Jul},
  publisher =    {PMLR},
  url = 	 {https://proceedings.mlr.press/v119/xiao20a.html}
}

@InProceedings{cascadeVAE,
  title = 	 {Learning Discrete and Continuous Factors of Data via Alternating Disentanglement},
  author =       {Jeong, Yeonwoo and Song, Hyun Oh},
  booktitle = 	 {Proceedings of the 36th International Conference on Machine Learning},
  pages = 	 {3091--3099},
  year = 	 {2019},
  editor = 	 {Chaudhuri, Kamalika and Salakhutdinov, Ruslan},
  volume = 	 {97},
  series = 	 {Proceedings of Machine Learning Research},
  month = 	 {09--15 Jun},
  publisher =    {PMLR}
}

@inproceedings{
hierarchical-vae,
title={PROGRESSIVE LEARNING AND DISENTANGLEMENT OF HIERARCHICAL REPRESENTATIONS},
author={Zhiyuan Li and Jaideep Vitthal Murkute and Prashnna Kumar Gyawali and Linwei Wang},
booktitle={International Conference on Learning Representations},
year={2020},
}

@Misc{Michlo2021Disent,
  author =       {Nathan Juraj Michlo},
  title =        {Disent - A modular disentangled representation learning framework for pytorch},
  howpublished = {Github},
  year =         {2021},
}

@InProceedings{Yang_2020_CVPR,
author = {Yang, Yanchao and Chen, Yutong and Soatto, Stefano},
title = {Learning to Manipulate Individual Objects in an Image},
booktitle = {Proceedings of the IEEE/CVF Conference on Computer Vision and Pattern Recognition (CVPR)},
month = {June},
year = {2020}
}

@InProceedings{control-vae,
  title = 	 {{C}ontrol{VAE}: Controllable Variational Autoencoder},
  author =       {Shao, Huajie and Yao, Shuochao and Sun, Dachun and Zhang, Aston and Liu, Shengzhong and Liu, Dongxin and Wang, Jun and Abdelzaher, Tarek},
  booktitle = 	 {Proceedings of the 37th International Conference on Machine Learning},
  pages = 	 {8655--8664},
  year = 	 {2020},
  editor = 	 {III, Hal Daumé and Singh, Aarti},
  volume = 	 {119},
  series = 	 {Proceedings of Machine Learning Research},
  month = 	 {13--18 Jul},
  publisher =    {PMLR}
}

@article{gaussian-mixture,
  author    = {Nat Dilokthanakul and
               Pedro A. M. Mediano and
               Marta Garnelo and
               Matthew C. H. Lee and
               Hugh Salimbeni and
               Kai Arulkumaran and
               Murray Shanahan},
  title     = {Deep Unsupervised Clustering with Gaussian Mixture Variational Autoencoders},
  journal   = {CoRR},
  volume    = {abs/1611.02648},
  year      = {2016},
  eprinttype = {arXiv},
  eprint    = {1611.02648},
  bibsource = {dblp computer science bibliography, https://dblp.org}
}

@misc{exp,
      title={Deep Exponential Families}, 
      author={Rajesh Ranganath and Linpeng Tang and Laurent Charlin and David M. Blei},
      year={2014},
      eprint={1411.2581},
      archivePrefix={arXiv},
      primaryClass={stat.ML}
}

@Inproceedings{isotrophic_gaussian,
 author = {Emile Mathieu and Tom Rainforth and N. Siddharth and Y. Teh},
 booktitle = {International Conference on Machine Learning},
 title = {Disentangling Disentanglement in Variational Autoencoders},
 year = {2018}
}

@inproceedings{hierar_prior,
  author    = {Minyoung Kim and Yuting Wang and Pritish Sahu and V. Pavlovic},
  year      = {2019},
  title     = {Bayes-Factor-VAE: Hierarchical Bayesian Deep Auto-Encoder Models for Factor Disentanglement},
  booktitle = {IEEE International Conference on Computer Vision},
  doi       = {10.1109/ICCV.2019.00307},
}
